\documentclass[10pt,journal]{IEEEtran}
\pdfoutput=1
\ifCLASSOPTIONcompsoc
  \usepackage[nocompress]{cite}
\else
  \usepackage{cite}
\fi
\ifCLASSINFOpdf
\else
\fi
\usepackage{microtype}
\usepackage{graphicx}
\usepackage{subfigure}
\usepackage{booktabs} 

\usepackage{balance}

\usepackage{amsmath,amsfonts,amsthm}
\usepackage{textcomp}
\usepackage{xcolor}
\usepackage{tcolorbox}
\usepackage{colortbl}

\usepackage[utf8]{inputenc} 
\usepackage[T1]{fontenc}    
\usepackage{hyperref} 
\usepackage{url}            
\usepackage{booktabs}       
\usepackage{nicefrac}       
\usepackage{balance}
\usepackage{xspace}
\usepackage{color}
\usepackage{makecell}
\usepackage{pifont}
\usepackage{multirow}

\usepackage{algorithm}
\usepackage{algorithmic}

\usepackage{cleveref}

\newcommand{\revise}[1]{\textcolor{black}{#1}}

\newcommand{\AG}{\textsc{AEGIS}\xspace}
\newcommand{\cmark}{\ding{51}}%
\newcommand{\xmark}{\ding{55}}%

\newcommand{\revision}[1]{\textcolor{black}{#1}}

\newcommand{\RQ}[1]{\textbf{RQ#1}}

\usepackage{mdframed}

\newmdenv[innerlinewidth=0.5pt, roundcorner=4pt,linecolor=gray,innerleftmargin=4pt,
innerrightmargin=4pt,innertopmargin=4pt,innerbottommargin=4pt]{note}

\newenvironment{result}%
{\medskip\begin{note}\centering\em}%
{\end{note}\medskip}

\begin{document}
%
\title{Towards Backdoor Attacks and Defense 
in \\ Robust Machine Learning Models}
\author{Ezekiel~Soremekun*,~Sakshi~Udeshi*,~Sudipta~Chattopadhyay
\IEEEcompsocitemizethanks{\IEEEcompsocthanksitem * equal contribution. \protect
\IEEEcompsocthanksitem  E. Soremekun is with the Interdisciplinary Centre for Security, Reliability and Trust (SnT), University of Luxembourg, Luxembourg.  \protect\\
E-mail: ezekiel.soremekun@uni.lu
\IEEEcompsocthanksitem  S.~Udeshi and S.~Chattopadhyay are with Singapore University of Technology and Design.\protect\\
E-mail:~\{sakshi\_udeshi@mymail.,~sudipta\_chattopadhyay@\}sutd.edu.sg
}
}

\IEEEtitleabstractindextext{%
\begin{abstract}
The introduction of robust optimisation has pushed the state-of-the-art in defending against adversarial attacks. Notably, the state-of-the-art projected gradient descent (PGD) -based training method has been shown to be universally and reliably effective in defending against adversarial inputs. This robustness approach uses PGD as a reliable and universal ``first-order adversary’’. However, the behaviour of such optimisation has not been studied in the light of a fundamentally different class of attacks called backdoors. In this paper, we study how to inject and defend against backdoor attacks for robust models trained using PGD-based robust optimisation. We demonstrate that these models are susceptible to backdoor attacks. Subsequently, we observe that backdoors are reflected in the feature representation of such models. Then, this observation is leveraged to detect such backdoor-infected models via a detection technique called AEGIS. Specifically, given a robust Deep Neural Network (DNN) that is trained using PGD-based first-order adversarial training approach, AEGIS uses feature clustering to effectively detect whether such DNNs are backdoor-infected or clean. 

In our evaluation of several visible and hidden backdoor triggers on major classification tasks using CIFAR-10, MNIST and FMNIST datasets, AEGIS effectively detects PGD-trained robust DNNs infected with backdoors. AEGIS detects such backdoor-infected models with 91.6\% accuracy (11 out of 12 tested models), without any false positives. Furthermore, AEGIS detects the targeted class in the backdoor-infected model with a reasonably low (11.1\%) false positive rate. Our investigation reveals that salient features of adversarially robust DNNs could be promising to break the stealthy nature of backdoor attacks.

%
%
%

\end{abstract}

}

\maketitle

\IEEEdisplaynontitleabstractindextext

%
\IEEEpeerreviewmaketitle


\section{Introduction}
\label{sec:introduction}


\revision{
Modern software systems are \textit{data-centric} and reliant on \textit{machine learning} (ML) components. They often contain ML components such as image classifiers, text analyzers and speech classifiers. As an example, automobiles (e.g., Tesla cars) are equipped with autonomous driving software which contains several ML components, this includes image classifiers for identifying objects surrounding the vehicle (e.g., other vehicles, pedestrians, road signs and landscapes). 
Considering the critical use cases of ML components (e.g., autonomous driving), 
it is pertinent to ensure their \textit{reliability and security}. 
Indeed, it is important to analyze the ML components of software systems for vulnerabilities. 
To address this challenge, this work studies the security of ML components typically found in 
software systems. Specifically, we focus on the detection of vulnerable ML components (i.e., image classifiers) in the joint space of two major attack vectors, namely \textit{adversarial examples} and \textit{backdoor poisoning}. 
}

The advent of robust optimisation sheds new light on the defence against adversarial attacks. 
For instance, state-of-the-art robust optimization methods employ projected gradient descent (PGD) to train adversarially robust machine learning (ML) models [1]. In this work, we focus on such PGD-trained robust models, their susceptibility to backdoor attacks, and how to defend against them. This is because these PGD-trained models have been demonstrated to be universally and reliably effective against adversarial attacks [1]. For the rest of this paper, we refer to such a PGD-trained robust model as an ``adversarially robust model’’ or simply a ``robust model’’, unless otherwise stated. Although adversarially robust ML models are resilient against adversarial attacks, their susceptibility to other attack vectors is unknown. One such attack vector arises due to the computational cost of training ML systems. Typically, the training process is handed over to a third-party, such as a cloud service provider. Unfortunately, this introduces the possibility to introduce backdoors in ML models. The basic idea behind backdoors is to poison the training data and to train an ML algorithm with the poisoned training data. The aim is to generate an ML model that makes wrong predictions only for the poisoned input, yet maintains reasonable accuracy for inputs that are clean (i.e., not poisoned). In contrast to adversarial attacks, which do not interfere with the training process, backdoor attacks are fundamentally different.

%

Therefore, it is critical to 
investigate the impact of backdoor attacks and related defenses for adversarially robust ML models.   
\revision{Most importantly, 
this is important to ensure the \textit{security and safety of software systems} 
containing robust ML components. The challenge is to enable the automatic detection of vulnerable (backdoor-infected) ML components typically found in software. Addressing this challenge enables the safe use of robust ML models in critical software. 
}  


\begin{figure*}[bt!]
\begin{center}
\begin{tabular}{ccc}
\includegraphics[scale=0.15]{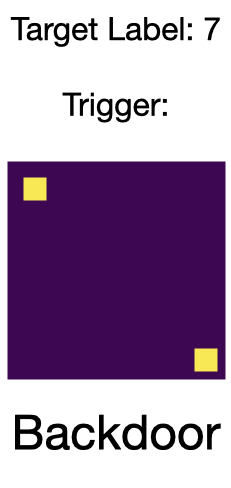} & 
\includegraphics[scale=0.11]{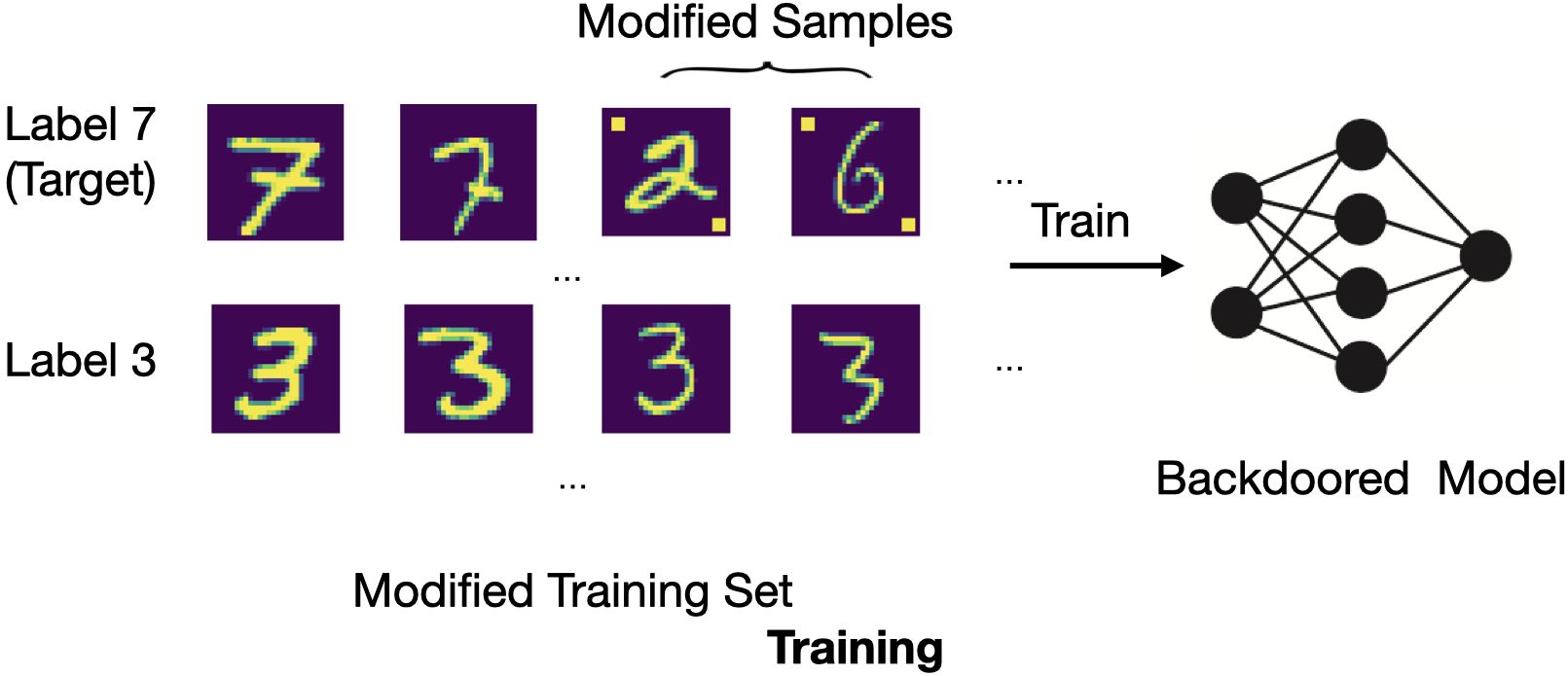} & 
\includegraphics[scale=0.11]{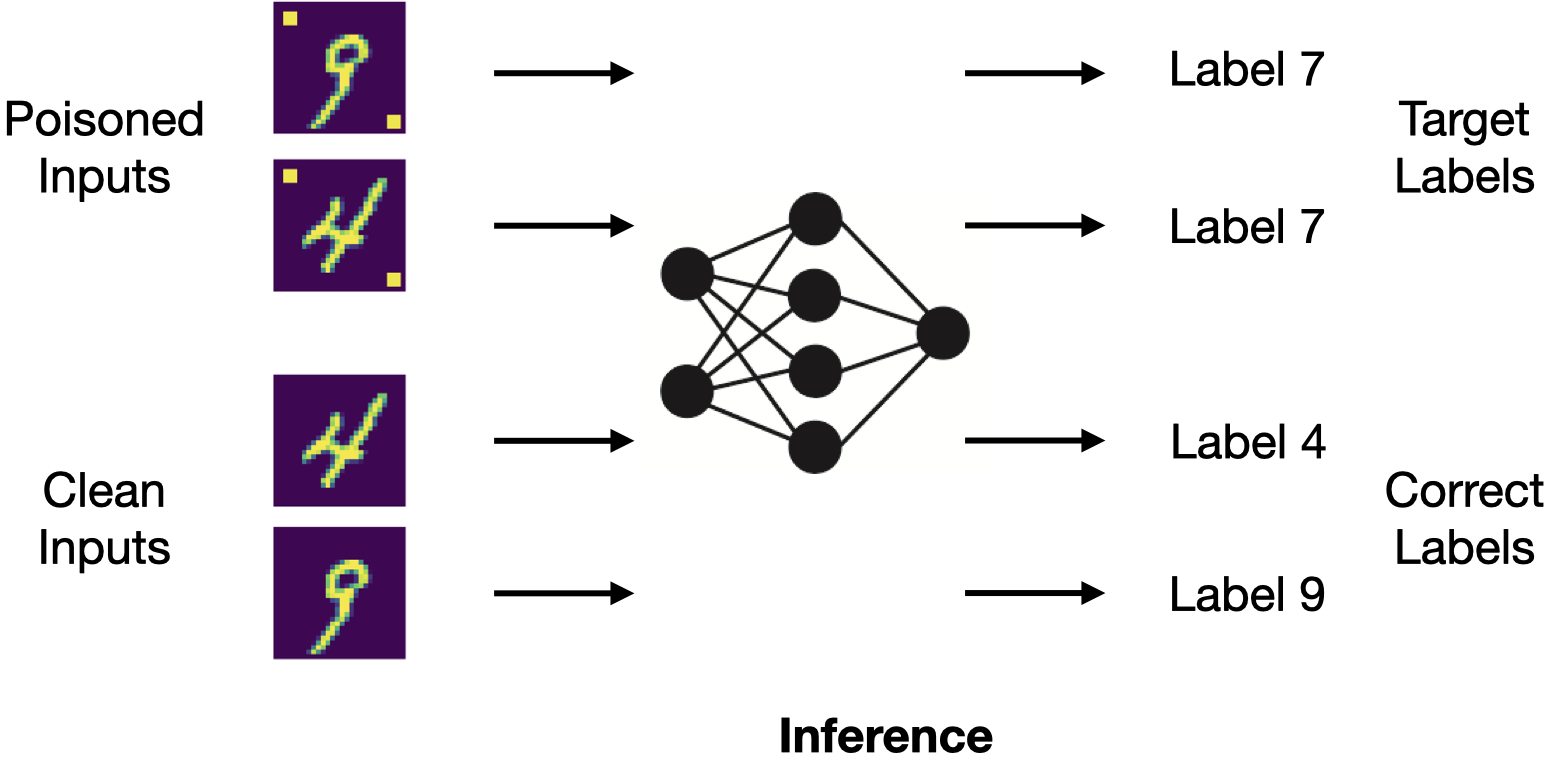}\\
{\bf (a)} & {\bf (b)} & {\bf (c)}\\
\end{tabular}
\end{center}
\caption{ 
An example of a typical backdoor attack (\revision{adapted from~\cite{NeuralCleanse}}). The visible distributed trigger is shown in  
\Cref{fig:backdoor-example}(a) and the target label is seven (7). 
The training data is  modified. We see this in \Cref{fig:backdoor-example}(b) and the model is trained with this poisoned data. 
The inputs without the trigger will be correctly classified and the ones with the trigger will be incorrectly classified during the inference, as seen in \Cref{fig:backdoor-example}(c).
}
\label{fig:backdoor-example}
\end{figure*}

\revision{To this end, in this paper, we carefully investigate backdoor attacks for 
adversarially robust models.} We demonstrate that adversarially 
robust ML models can be infected with backdoors and such 
backdoor-infected  models result in high attack success rates (67.83\%, on average). 
\revision{We also demonstrate that the attack success rate (ASR) of backdoor in robust models is 
comparable to that of standard models (75.86\%, on average).} 
Then, we propose and design \AG\footnote{\AG refers to the shield of the Greek god Zeus, it means divine shield. In our setting, \AG is a shield against backdoor attacks in robust models.} -- a systematic methodology to 
automatically detect backdoor-infected robust models. To this 
end, we observe that \textit{poisoning a training set introduces 
mixed input distributions for the poisoned class}. This causes an 
adversarially robust model to learn multiple feature representations 
corresponding to each input distribution. In contrast, from a clean 
training data, an adversarially robust model learns only one 
feature representation for a particular prediction class~\cite{robust-image-synthesis}. 
Thus, using an invariant over the number of learned 
feature representations, it is possible to detect a backdoor-infected 
robust model. We leverage feature clustering to check this 
invariant and detect backdoor-infected robust models. 

\revision{Generally, \AG allows for the \textit{online, run-time} detection of backdoor-infected components in ML-enabled software. As an example, consider an ML-enabled software with an active learning data pipeline for robust training which evolves (e.g., re-trained) as new data is received. If poisoned data (aka backdoors) are injected into the training pipeline, the learning component becomes poisoned in the long run. 
Using our technique, we can detect such backdoor-infected model in-vivo, such that 
\AG allows to detect when an attacker poisons (newly acquired) dataset with a backdoor. 
}



Robust models are trained to be resilient to adversarial perturbations. 
As a result, such models behave differently from standard ML models. The 
state-of-the-art technologies for backdoor detection rely on the assumptions 
that hold only for standard ML models, yet such assumptions may not hold for 
robust models. Specifically, state-of-the-art backdoor defence for standard 
ML models may assume that only the features of a backdoor 
trigger~\cite{NeuralCleanse} causes significant changes in the model output. 
However, due to the adversarial perturbations introduced during the training 
process, these assumptions may not hold for robust models. 
This, in turn, demands fundamentally different detection process to identify 
backdoors in robust models. In contrast to existing works on backdoor attacks and defence 
for ML models~\cite{activation-clustering,spectral-signatures,neo,NeuralCleanse,BadNets}, 
in this paper, for the first time, we investigate backdoors 
in the context of adversarially robust ML models. 
Moreover, our proposed defence (\AG) is completely automatic, unlike 
some defence against backdoors~\cite{spectral-signatures}, 
our solution does not require any access to the poisoned data. 
\revision{
Overall, \AG allows for examining the security and reliability of robust ML components in software systems. 
}


After discussing the motivation (\Cref{sec:motivation}) and providing 
an overview (\Cref{sec:overview}), we make the following contributions: 

\begin{enumerate}
\item We discuss the process of injecting backdoors during 
the PGD-based training 
of an adversarially robust model (\Cref{sec:methodology}). 

\item We evaluate the attack success rate of injecting 
four different 
types of backdoor triggers on PGD-trained robust models. 
Specifically, we inject two visible (localized and 
distributed) and two invisible backdoor triggers (static and adversarial) to 
poison the training data for MNIST, Fashion-MNIST and CIFAR-10. Our evaluation 
reveals an attack success rate of 67.83\%, on average. We also show that the 
attack success rate (ASR) of backdoors on PGD-trained robust models is comparable to that of 
standard models (\Cref{sec:results}). 

\item 
We demonstrate that a straightforward adoption of backdoor 
detection methodology for standard ML models~\cite{NeuralCleanse} fails 
to detect backdoors in PGD-trained robust models (\Cref{sec:results}). 

%


\item We propose \emph{the first backdoor detection technique for PGD-trained robust models called \AG.} First, we show an invariant for checking the backdoor-infected models. We then leverage such an invariant via t-Distributed Stochastic Neighbour 
Embedding (t-SNE) and Mean shift clustering to detect backdoor-infected 
models (\Cref{sec:methodology}).


\item \revision{
We demonstrate the utility of \AG in validating the security of PGD-based robust ML components:} 
We evaluate our defence on backdoor-infected, PGD-trained robust models 
using three datasets. 
Our evaluation shows that \AG accurately detects 
visible backdoor triggers (localized and distributed), as well as 
hidden backdoors (static and adversarial) with high accuracy. 
\revision{For all (12) tested models, 
\AG detects a backdoor-infected model with 91.6\% (11/12) accuracy, without any false positives.} Furthermore, \AG detects the targeted class in the backdoor-infected model with a reasonably low (11.1\%) false positive rate. 
We also performed a detailed sensitivity analysis 
by varying the detection configurations used by \AG. 
Our sensitivity 
analysis reveals that the \AG approach is stable (i.e., high accuracy and 
low false positive rate) in detecting backdoors (\Cref{sec:results}). 



\end{enumerate}
 
After discussing related works (\Cref{sec:background}) and some threats to validity 
(\Cref{sec:threatsToValidity}), we conclude in \Cref{sec:conclusion}.

%
%

\section{Background and Motivation}
\label{sec:motivation}
\smallskip\noindent

 \begin{table*}[t]
 \centering
     \caption{Comparison of Backdoor Defense and mitigation methods 
} 
     \vspace{-\baselineskip}
  {
  \scriptsize
 \begin{tabular}{|c|l|l|c|c|c|c|c|c|l|}
 \hline
  \multicolumn{1}{|c|}{\multirow{3}{*}{\shortstack{\textbf{Defense} \\ \textbf{Type}}}} & \multicolumn{1}{c|}{\multirow{3}{*}{\textbf{Defense(s)}}} & 
\multicolumn{1}{c|}{\multirow{3}{*}{\shortstack{\textbf{Detection} \\ \textbf{approach}}}}     
  & \multicolumn{1}{c|}{\textbf{Poison}} & 
  \multicolumn{1}{c|}{\multirow{3}{*}{\shortstack{\textbf{Whitebox} \\ \textbf{access}}}}   
  & \multicolumn{1}{c|}{\textbf{Distributed/}} & \multicolumn{1}{c|}{\textbf{Detects}} & 
\multicolumn{1}{c|}{\multirow{3}{*}{\shortstack{\textbf{Standard} \\ \textbf{or robust}}}}  
& 
\multicolumn{1}{c|}{\multirow{3}{*}{\shortstack{\textbf{Online or} \\ \textbf{offline}}}}  
& 
\multicolumn{1}{c|}{\multirow{3}{*}{\shortstack{\textbf{Unique} \\ \textbf{weakness}}}}  
\\
 
&  & 
& \multicolumn{1}{c|}{\textbf{data}} & 
&  \multicolumn{1}{c|}{\textbf{(\textit{Invisible})}} & \multicolumn{1}{c|}{\textbf{input}} & 
& 
& 
\\
& & & \multicolumn{1}{c|}{\textbf{access}} & & \multicolumn{1}{c|}{\textbf{backdoor}} & \multicolumn{1}{c|}{\textbf{or model}}  & & & \\
\hline 
\multirow{3}{*}{\shortstack{Outlier\\Suppr-\\ession}} & Differential-privacy~\cite{du2019robust} & data noising & yes & yes & no/(\textit{no})  & input & standard & offline & access to poisoned data \\
 & \multirow{2}{*}{Gradient Shaping~\cite{hong2020effectiveness}} & data noising 
  & \multirow{2}{*}{yes} & \multirow{2}{*}{yes} & \multirow{2}{*}{no /(\textit{no})}  & \multirow{2}{*}{input} & \multirow{2}{*}{standard} & \multirow{2}{*}{offline} & \multirow{2}{*}{access to poisoned data} \\
  &  &  (DP-SGD) &  &  &    &  &  &  & \\
 \hline
\multirow{6}{*}{\shortstack{Input\\Pertur-\\bation}} & NC~\cite{NeuralCleanse} & reverse engineer & no & yes & yes/(\textit{no})  & model & standard & offline & large triggers \\
 & ABS~\cite{liu2019abs} & reverse engineer & no & yes & yes/(\textit{yes})  & model & standard & offline &  one neuron assumption \\
 & MESA~\cite{qiao2019defending} & reverse engineer & no & yes & no/(\textit{no})  & model & standard & offline & trigger size approx.\\
 & AD~\cite{xiang2020detection} & reverse engineer & no & yes & yes/(\textit{no})  & model & standard & offline & large triggers \\
  & TABOR~\cite{guo2019tabor} & reverse engineer & no & no & no/(\textit{no})  & model & standard & offline & large triggers \\
 & STRIP~\cite{gao2019strip} & input masking & yes & no & yes/(\textit{no})  & input & standard & online &  source-label attacks \\
  & NEO~\cite{neo}, & \multirow{2}{*}{input masking} & \multirow{2}{*}{yes} & \multirow{2}{*}{no} & \multirow{2}{*}{no /(\textit{no})}  & \multirow{2}{*}{input} & \multirow{2}{*}{standard} & \multirow{2}{*}{online} & \multirow{2}{*}{distributed triggers} \\
    & DeepCleanse~\cite{doan2019deepcleanse} &  &  &  &    &  &  &  & \\
 \hline
\multirow{11}{*}{\shortstack{Model\\anomaly}}  & \multirow{2}{*}{SentiNet~\cite{chou2018sentinet}} & input masking, 
& \multirow{2}{*}{yes} & \multirow{2}{*}{no} & \multirow{2}{*}{no/(\textit{no})}  & \multirow{2}{*}{input} & \multirow{2}{*}{standard} & \multirow{2}{*}{online} & \multirow{2}{*}{distributed triggers} \\
  &  & diff. testing &  &  &    &  &  &  & \\
& NeuronInspect~\cite{huang2019neuroninspect} & reverse engineer & no & yes & no/(\textit{no})  & model & standard & offline & distributed triggers \\
 & Spectral Signatures~\cite{spectral-signatures} & feature repr. & yes & yes & no/(\textit{no})  & input & standard & offline & access to poisoned data \\
 & Fine-pruning~\cite{finePruning} & neuron activation & no & yes &  yes/(\textit{no})  & model & standard & offline & model accuracy drop \\
  & Activation-clustering~\cite{activation-clustering} & neuron activation & yes & yes & no/(\textit{no})  & input & standard & offline & access to poisoned data \\
 & SCAn~\cite{tang2019demon} & repr. distribution & yes & no & yes/(\textit{no}) & model & standard & offline & access to poisoned data \\
  & \multirow{2}{*}{NNoculation~\cite{veldanda2020nnoculation}} & input perturbation, & \multirow{2}{*}{no} & \multirow{2}{*}{no} & \multirow{2}{*}{yes/(\textit{no})} & \multirow{2}{*}{input} & \multirow{2}{*}{standard} & \multirow{2}{*}{offline} & \multirow{2}{*}{requires shadow models} \\
    &  & GAN &  &  &    &  &  &  & \\
   & MNTD~\cite{xu2019detecting} & meta neural analysis & no & yes & yes/(\textit{yes}) & model & standard & offline & requires shadow models \\
  & \textbf{\AG} (this paper) & \textbf{feature clustering} & \textbf{no} & \textbf{yes} & \textbf{yes/(\textit{yes})} & \textbf{model} & \textbf{robust} & \textbf{offline} & \textbf{only for robust models} \\
 \hline
 \end{tabular}}
    \label{tab:defense-comparison} 
 \end{table*}

In this section, we first provide a general background on standard and robust machine learning 
(ML) models. Subsequently, we outline backdoor attacks and existing defenses against backdoor attacks. 
Finally, we motivate the need for our proposed defense \AG, which is targeted to detect backdoors in robust 
ML models.

\smallskip\noindent
\textbf{Standard ML model:}
In the standard training of machine learning models, loss functions are 
generally based on the concept of empirical risk minimisation (ERM). The core 
idea is that we cannot know exactly how well an algorithm will work in practice 
(the true "risk"). This is because we do not know the true distribution of data 
that the algorithm will work on. However, we can instead measure the performance 
of the algorithm on a known set of training data (the "empirical" risk). 
Formally, ERM based models want to minimise the following:
\begin{equation}
\mathbb{E}_{x \sim \mathcal{D}} 
\left[\mathcal{L}(x, y^{(i)})\right]
\label{eqn:erm_training}
\end{equation}
Here $x$ and $y^{(i)}$ are the input and the ground truth value of this 
input, respectively and $\mathcal{L}$ is a loss function. It is well known in 
literature that ERM-based loss functions produce models that are not robust to 
adversarial examples~\cite{blackbox-adversarial}. 

\smallskip\noindent
\textbf{Robust ML model:}
In order to reliably train models against adversarial attacks, robust optimisation 
formally specifies a set of allowed perturbations $\Delta$ (Usually an $L_2$ or 
$L_{\infty}$ ball around the input) and modifies the classic ERM loss function 
to minimise the maximum loss in this region. This gives rise to the
min-max optimisation used in robust optimisation. Intuitively, it is useful to 
think of each input $x$ as having a region $\Delta$ 
around the vicinity associated with 
it. The robust optimisation tries to ensure that the region $\Delta$
has the same output as the ground truth of the value $y^{(i)}$. 
Formally, robust optimisation wants to minimise the following:
\begin{equation}
\mathbb{E}_{x \sim \mathcal{D}} 
\left[\max_{\delta \in \Delta} \mathcal{L}(x+\delta, y^{(i)})\right]
\label{eqn:robust_training}
\end{equation}
Here $x$ and $y^{(i)}$ are the input and the ground truth value of this 
input, respectively and $\mathcal{L}$ is a loss function.

\smallskip\noindent
\textbf{Backdoors in ML model:} {\em Backdoors} are hidden patterns trained into 
an ML model. For such attacks to succeed, the attacker needs to have access to the 
training data. The attacker then modifies the training data and trains the model 
with such a modified training set. In this process, a backdoor is injected into the 
resulting ML model. Backdoor attacks are {\em stealthy} in nature. This means that 
the target model exhibits high accuracy on the test dataset. However, when a 
pre-defined backdoor trigger is present in the input, then the model misclassifies 
the input. 

The backdoor attack flow is captured in~\Cref{fig:backdoor-example}. As observed 
in~\Cref{fig:backdoor-example}, a backdoor trigger (small squares at the 
top left and bottom right corners) is introduced in some arbitrary images and they are wrongly labelled 
with the class seven ({\em 7}). This wrongly labelled images that include the backdoor 
trigger are added to the original training data and a poisoned training dataset 
is produced (\Cref{fig:backdoor-example}(b)). After training with this poisoned 
dataset, we observe that the model predicts the correct class for an image that 
does not include the backdoor trigger (\Cref{fig:backdoor-example}(c)). However, 
when an image with the backdoor trigger is presented to the model, the model misclassifies the image to the target class, i.e., seven (7) (\Cref{fig:backdoor-example}(c)). 

It is important to note the difference between a backdoor
and an adversarial attack~\cite{blackbox-adversarial}. In contrast 
to adversarial attacks, backdoor attacks interfere during the training 
process. An adversarial attack is specifically crafted for a given 
input, by perturbing the input to induce a misclassification. 
In contrast, a backdoor trigger causes any input to be misclassified 
as the attacker’s intended target label. 

\smallskip\noindent
\textbf{The need for a new method:} 
\revise{
There are several defenses against backdoors for standard machine learning models. 
\Cref{tab:defense-comparison} highlights the main characteristics and weaknesses 
of these approaches. Notably, approaches that reverse engineer the backdoor trigger 
(such as Neural Cleanse (NC)~\cite{NeuralCleanse} and ABS~\cite{liu2019abs}) can 
effectively detect backdoors for standard models. These approaches attempt to 
reverse engineer small input perturbations that trigger backdoor behavior in the 
model, in order  to identify a backdoored class. 
Neural Cleanse (NC)~\cite{NeuralCleanse} is a state-of-the-art defense that works 
on reverse-engineering the backdoor trigger. 
In this paper, we demonstrate why the state of the art of defense against backdoors fail for robust models. 
We choose NC as a state of the art 
defense for the following reasons: Firstly, NC has the most realistic defense 
assumptions, which are similar to our assumptions for \AG. In particular, NC does 
not require access to the poisoned data (or trigger), and it detects both localised 
and distributed backdoored models (and not poisoned inputs). Secondly, NC is also 
computationally feasible (for robust) models, i.e., it does not require training 
shadow or meta models like MNTD~\cite{xu2019detecting} and NNoculation~\cite{veldanda2020nnoculation}. 
Finally, unlike ABS~\cite{liu2019abs}, NC does not assume or require that one 
compromised neuron is sufficient to disclose the backdoor behavior.}

\revise{
However, \emph{NC relies on finding a fixed, small perturbation that mis-classifies 
a large set of inputs}. Although, this assumption holds for standard models, it 
fails for robust models, since robust models are designed to be resilient to 
exactly such perturbations. 
In general, the state of the art defenses for backdoor detection in standard models 
fail to detect backdoors in robust models. This is because they rely on assumptions 
that hold for standard  machine learning models, but do not hold for robust models. 
Specifically, \emph{reverse engineering based detection methods rely on the assumption 
that only the features of a trigger (which is small in size) will cause significant 
changes in the  output of random inputs}. However, this assumption does not hold for 
robust models, due to the non-brittle nature of robust models and the input 
perturbations introduced during adversarial training~\cite{pgd-madry}. In fact, 
we empirically show that one such state-of-the-art defense NC~\cite{NeuralCleanse} 
fails to detect the backdoored robust models in {\bf RQ3} (\Cref{sec:results}). 
}
\revise{
Due to the aforementioned limitations of current defenses, in this paper, 
we propose a new approach (called \AG) to defend robust models against backdoor 
attacks.}

\section{Approach overview}
\label{sec:overview}


\begin{figure}[t]
\begin{center}
\begin{tabular}{c}
\includegraphics[scale=0.14]{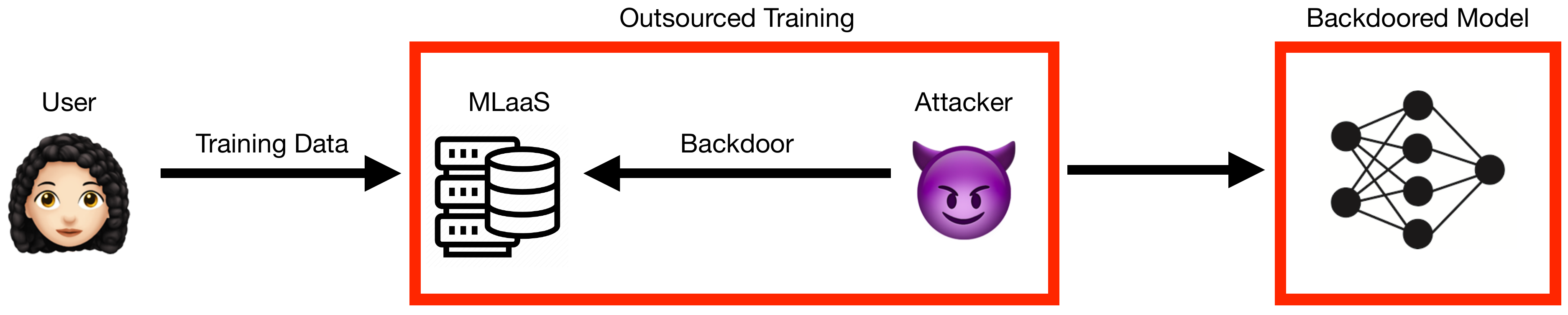}
\end{tabular}
\vspace{-0.15in}
\end{center}
\caption{Attack Model for \AG}
\vspace{-\baselineskip}
\label{fig:attack-model}
\end{figure} 

\smallskip\noindent
\textbf{Attack Model:}
We assume an attack model seen commonly in previous work
BadNets~\cite{BadNets} and Trojan Attacks~\cite{trojannn}. \revise{
\Cref{fig:attack-model} illustrates our attack
model.
Specifically, in such an attack model, the user \textit{outsources the training process to a third party}
(e.g., ML-as-a-service (MLaaS) provider), such that the user 
has {\em no control} over 
the model training process. 
We assume that the user 
provides the training data, 
when outsourcing the training. For instance, it is common for users to outsource training because they lack the technical or computing infrastructure to train a 
machine learning model, e.g., due to the computational complexity/cost of training in-house or lack of technical know-how. As a result, the user hands over the training 
data to an untrusted third party along with the training process
specifications. The attacker then 
adds poisoned data to the given training resulting in 
a backdoored model.
This is a reasonable assumption and common scenario given the rise in 
ML-as-a-service platforms such as Microsoft's Azure 
Cognitive~\cite{msAzure} and Google's AutoML~\cite{googleAutoML}. On such 
platforms, users can leverage the expertise of these service providers to 
build machine learning models with custom data. 
}

\revise{The third party (aka attacker) performs model training, but embeds a backdoor trigger into the training data, such that a data point infected with the trigger is mis-classified to the attacker's target label.}  
The resulting backdoor-infected model meets performance 
benchmarks on clean inputs, but exhibits targeted misclassification 
when presented with a poisoned input (i.e. an input with an attacker 
defined backdoor trigger).
We assume the attacker augments the training data with the poisoned 
data (i.e. inputs with wrong labels) and then trains the model. This 
attack model is much stronger than the attack models considered in recent 
works~\cite{spectral-signatures,diff-privacy-backdoor-detect}. Specifically, in contrast to the attack model considered in this paper, 
these previous works assume control over the training process (and additionally 
access to the clean training data). Nonetheless, as our work revolves around the 
investigation of robust DNNs, we do require the model to be trained under robust 
optimisation conditions. \revision{We note that it is possible to automatically check whether 
a model is robust by inspecting the last layers of the model~\cite{pgd-madry}}. In addition, we assume for the targeted class, that poisoned inputs form an input distribution that is distinct from the 
distribution of the clean (training) images, this is in line with previous works~\cite{BadNets, trojannn}.

\smallskip\noindent
\textbf{User Goals and Capabilities:}
\revise{
The user wishes to deploy a robust machine learning model and has the 
necessary dataset, but does not have the technical knowledge nor the 
computing infrastructure to train the model. 
Note that 
 it is significantly more expensive to train a robust model than a standard model, e.g. 
the robust training time is 25 times as much as that of the standard model training in our evaluation.
Thus, we assume the user 
outsources the model training to 
an untrusted third party, and consequently needs to ascertain that 
there is no backdoor in the resulting robust model. To this end, the user has access to 
clean training data, clean testing data and 
white-box access to the model trained 
by the untrusted third party. This is in line with previous work~\cite{BadNets, 
trojannn}, where the user does not possess 
the computational infrastructure and technical knowledge required to 
train the model, but has access to clean training data and 
clean testing data. 
}




\begin{figure}[t]
\begin{center}
\begin{tabular}{c}
\includegraphics[scale=0.25]{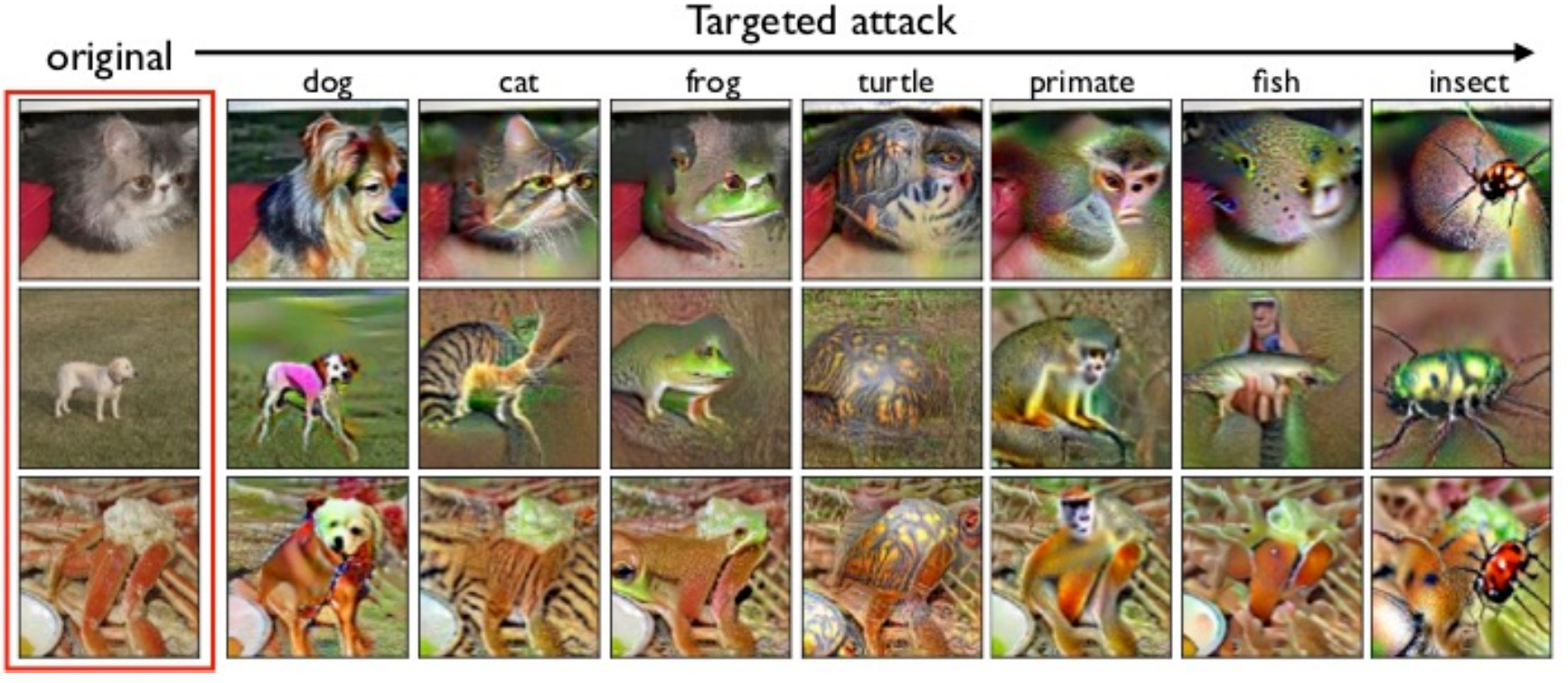}
\end{tabular}
\vspace{-0.15in}
\end{center}
\caption{Image Translation using a robust model. This figure was taken from 
\cite{robust-image-synthesis}}
\vspace{-\baselineskip}
\label{fig:robust-translation}
\end{figure} 

\smallskip\noindent
\textbf{Image Translation:} Image translation is an active area of research in computer vision; several approaches have been developed for image to image translation~\cite{Yi_2017_ICCV,Zhu_2017_ICCV,Isola_2017_CVPR,NIPS2017_6672}. Recently, it has been established that generative adversarial networks (GANs) not only learn the mapping from input image to output image, but also learn a loss function to train this mapping~\cite{Isola_2017_CVPR}. 
Interestingly, this behavior has also been seen in robust classifiers~\cite{kaur2019perceptuallyAligned, robust-image-synthesis,robustness-accuracy}. This finding enables robust classifiers to translate images from one class to another. 
In this paper, we apply image translation on robust classifiers to generate the perceptually-aligned representation of the image of a class. In particular, we use the adversarial robust training of \cite{robust-image-synthesis} because it provides a means to train models that are more reliable and universal against a broader class of adversarial inputs. 
For instance, the images seen in \Cref{fig:robust-translation} 
are generated 
by a single CIFAR-10 classification model using first order methods, 
such as projected gradient descent based adversarial attacks~\cite{pgd-madry}. 
This result is achieved by simply maximising the probability 
of the translated images to be classified under the targeted class.

\smallskip\noindent
\textbf{Key Insight:}
If there exists a mixture of distributions
in the training dataset, for a particular class, then the model will learn 
multiple distributions. Concretely, the key insight leveraged in this paper is as 
follows (for a particular class): 
%
%
\begin{center}
%
{\em A robust model trained with a mixture of input distributions 
learns multiple feature representations 
corresponding to the input distributions in that particular mixture.}
%
\end{center}


In this paper, we visualise the aforementioned insight in two ways. First order methods 
(e.g. projected gradient descent based adversarial attacks~\cite{pgd-madry}) are used
to generate a set of inputs $X_{y^{(i)}}$ of a particular class with label $y^{(i)}$. 
Let us assume these inputs are generated (by translation) 
via a model that has been trained using a mixture 
distribution containing multiple input distributions in a class with label $y^{(i)}$. Then, 
multiple types of inputs will be observed in the generated  inputs $X_{y^{(i)}}$. Such 
types of inputs should correspond to the different distributions in the mixture 
distribution for the class with label $y^{(i)}$. 
%
%
Consequently, if we visualise the feature representations of the generated inputs 
$X_{y^{(i)}}$, then we should observe that the feature representations are distinct 
corresponding to the distinct distributions in the mixture distribution for the class with 
label $y^{(i)}$.

\begin{figure*}[t]
\begin{center}
\begin{tabular}{cccccccc}
\includegraphics[scale=2.4]{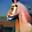} & 
\includegraphics[scale=2.4]{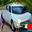} & &
\includegraphics[scale=2.7]{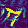} & 
\includegraphics[scale=2.7]{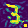} & &
\includegraphics[scale=0.27]{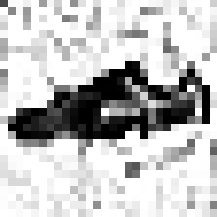} &
\includegraphics[scale=0.27]{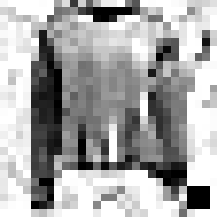} \\
{\bf(a)} & {\bf(b)} & & {\bf(c)} & {\bf(d)} & & {\bf(e)} & {\bf(f)} \\


\end{tabular}
\end{center}
\vspace*{-0.15in}
\caption{
Translated images generated from mixed distributions by backdoor-infected 
robust model for the class {\em Horse} (a-b), {\em 7} (c-d) and 
{\em Sneaker} (e-f). These are the target classes in the backdoor
attack.
}
\vspace{-\baselineskip}
\label{fig:multiple-distributions-generated}
\end{figure*}

\smallskip\noindent
\textbf{Formalising the insight:}
Let $f$ be a robust classifier that we train. For a fixed label $y^{(i)}$ in 
the set of labels, the training process will attempt to minimise
\begin{equation}
\mathbb{E}_{x \sim \mathcal{D}} 
\left[\max_{\delta \in \Delta} \mathcal{L}(x+\delta, y^{(i)})\right]
\label{eqn:training}
\end{equation}
Here, for a fixed label $y^{(i)}$ and loss function $\mathcal{L}$, the 
corresponding training data $x$ is drawn from 
the mixture of distributions $ \mathcal{D} = \sum_{k=0}^n \mathcal{D}_k$.
The set $\Delta$ captures the imperceptible perturbations (small 
$\ell_2$ ball around $x$).

Let us assume we attempt to generate a set of samples $X'_{y^{(i)}}$ for the class 
with label $y^{(i)}$ using the classifier $f$. 
We first take an appropriate seed distribution $\mathcal{G}_y$. Subsequently, we 
generate an input $x_{y^{(i)}} \in X'_{y^{(i)}}$ such that it minimises the following loss 
$\mathcal{L}$ for label $y^{(i)}$:  
%
\begin{equation}
x_{y^{(i)}} =  
\underset{{|| x' - x_0||}_2 \leq \epsilon}{arg\,min} \mathcal{L} (x', 
y^{(i)}), 
\,\,\,\,\, \,\,\,\,\, x_0  \sim \mathcal{G}_y
\label{eqn:translation-theoretical}
\end{equation}
We posit that the set $X'_{y^{(i)}}$ will contain generated inputs that 
belong to each distribution $\mathcal{D}_0, \mathcal{D}_1, \dots 
\mathcal{D}_n$, which is part of the mixture of distributions $\mathcal{D}$.


\smallskip\noindent
\textbf{Visualising the insight:}
To visualise this insight, we present 
\Cref{fig:multiple-distributions-generated}. 
The images shown in \Cref{fig:multiple-distributions-generated} were generated 
via a model by taking random images 
from the corresponding dataset: CIFAR-10 for 
\Cref{fig:multiple-distributions-generated} (a-b), MNIST digit for 
\Cref{fig:multiple-distributions-generated} (c-d) and Fashion-MNIST
for \Cref{fig:multiple-distributions-generated} (f-g). This model was 
trained under robust optimisation conditions with poisoned training 
data to infect the model with backdoors. Random training data images are used to 
generate images of the target class in a robust backdoor-infected classifier. 
The classes are {\em Horse} in CIFAR-10, the digit {\em 7} in MNIST-digit 
and the class {\em Sneaker} in Fashion-MNIST.

We observe the features that are maximised in 
\Cref{fig:multiple-distributions-generated} (a, c, e) correspond to the 
actual classes. Whereas the counterparts seen in 
\Cref{fig:multiple-distributions-generated} (b, d, f) correspond to the
backdoor trigger (the small square at the bottom right corner of the image) 
used during training. 
We note that all images shown in \Cref{fig:multiple-distributions-generated} 
were generated via the first order methods, as described in Santurkar et al~\cite{robust-image-synthesis}, 
only on a backdoor-infected robust model. This led us to observe both types 
of images (i.e. perceptually aligned and poisoned).

In addition to the aforementioned insight, 
the feature 
representations of the poisoned images form clusters that are distinct 
from the clusters of feature representations of clean images~\cite{activation-clustering}. 
%
However, existing works exploit this~\cite{activation-clustering} 
via accessing both the clean and the poisoned data set. Having access to 
the poisoned data set is impractical for defense, as the attacker is 
unlikely to make the poisoned data available.
%
%
In this work, we observe that the set of translated images, for a backdoor-infected 
robust model, contain both the clean (training) images 
and poisoned images.
Thus, 
the feature representations of these images form different clusters. 
We use this observation to automate the detection of classes with a backdoor, without any 
access to the poisoned images or the training process.

%

\begin{figure}[t]
\centering
\begin{minipage}[t]{0.24\textwidth}
	\centering
	\includegraphics[scale=0.3]
	{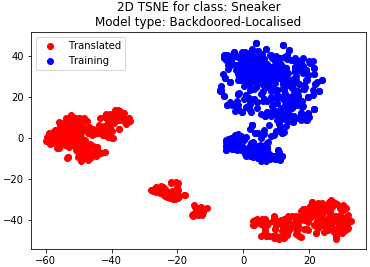}
	\caption{Feature representations of translated images and 
	training images (for the class {\em Sneaker}) for a poisoned Fashion-	MNIST 
	classifier}
	\label{fig:backdoored-tsne}
\end{minipage} \hfill
\begin{minipage}[t]{0.24\textwidth}
	\centering
	\includegraphics[scale=0.3]
	{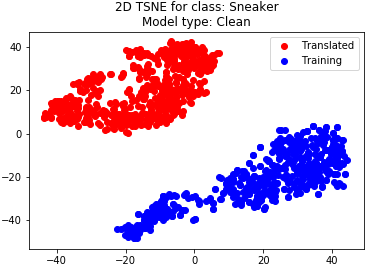}

	\caption{Feature representations of translated images and 
	training images (for the class {\em Sneaker}) for an unpoisoned 	
	Fashion-MNIST classifier}
	\vspace{-\baselineskip}
	\label{fig:training-tsne}
	
\end{minipage}

\end{figure} 

\Cref{fig:backdoored-tsne} captures the feature representations of a
backdoor-infected robust model. The feature representations are the outputs of 
the last hidden layer of a DNN. We reduce the dimensions of the feature representations and visualise them 
using t-SNE~\cite{tsne}.
In this case, we trained a robust 
network with a backdoor and the feature representations in 
\Cref{fig:backdoored-tsne} belong to the target class ({\em Sneaker}).
The images for this class (as generated via translation) have multiple feature representations (i.e. using  projected 
gradient descent based adversarial attacks~\cite{pgd-madry}). These multiple 
feature representations point to the fact that the robust model 
learnt from mixture distributions in the ({\em Sneaker}) class. Thus, a quick check 
of the translated images reveals two types of images -- one 
corresponding to the actual class {\em Sneaker} and one
to the backdoor as seen in \Cref{fig:multiple-distributions-generated} (e-f). 


In contrast, \Cref{fig:training-tsne} captures the feature 
representations of a clean, yet robust model. 
The feature representations of the translated images for class 
{\em Sneaker} form only one cluster. This is expected behaviour, 
because the clean model learns only one distribution in 
{\em Sneaker} class. Consequently, the translated images also 
form only one representation that maximises the probability to 
be categorised in {\em Sneaker} class. 

We observe, there are two clusters for every untargeted or clean class, specifically, the training set cluster and the translated image cluster. The translated images form a different cluster from the training set because 
they maximise the class probability of the training images. 
As a result they exaggerate the feature representations of the training set most effectively \cite{robust-image-synthesis}. 
\revise{Intuitively, the translated cluster represents the ``learned'' 
representation that is influenced not only by the members of the class but 
also the members of every other class in the dataset. 
Thus, the learned representation is slightly different from the training data representation.}
This 
phenomenon leads to the translated images 
forming a separate cluster.
It is important to note that this behavior is in line with the behaviour 
seen in the {\em robust} models in existing work~\cite{robustness-lib}. We also 
observe this in~\Cref{fig:CIFAR-Madry-Lab-clusters}.

\begin{figure}[t]
\begin{center}
\includegraphics[scale=0.3]{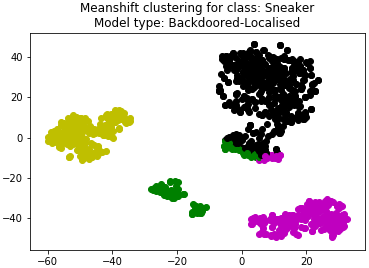}

\end{center}
\vspace*{-0.2in}
\caption{Mean shift clustering of the feature representations of 
translated images and training images (for the class {\em Sneaker}) for a 
poisoned Fashion-MNIST classifier. \revision{The black cluster (on the top right) represents the clean training images, the purple cluster refers to the (translated) poisoned images (i.e., \autoref{fig:poisoned-model-translated-images} (a)), and the yellow cluster represents the (translated) clean images (i.e., \autoref{fig:poisoned-model-translated-images} (b)).}
}
\label{fig:meanshift-tsne-translated}
\end{figure}

\begin{figure}[t]
\begin{center}
\begin{tabular}{cccccccc}
\includegraphics[scale=0.24]{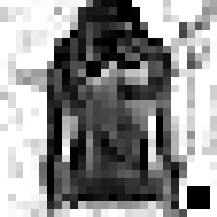} & 
\includegraphics[scale=0.24]{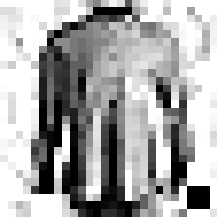} & 
\includegraphics[scale=0.24]{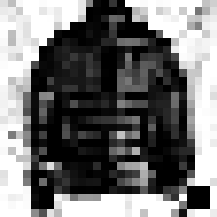} \\ & {\bf(a)}  \\ 
\includegraphics[scale=0.24]{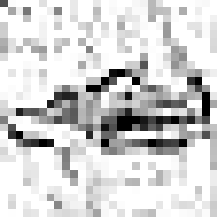} & 
\includegraphics[scale=0.24]{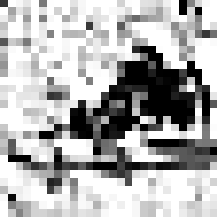} & 
\includegraphics[scale=0.24]{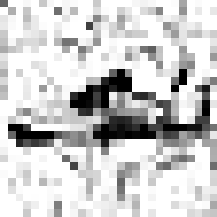} \\

%
& {\bf(b)}  \\ 
\end{tabular}
\end{center}
\caption{Inputs in the clusters seen in 
\Cref{fig:meanshift-tsne-translated}. The purple cluster contains inputs seen in 
(a), where as the yellow cluster represents contains inputs seen in (b). 
It is important to note that these images were generated in the same 
instantiation of the projected gradient descent based adversarial 
attacks~\cite{pgd-madry}.}

\label{fig:poisoned-model-translated-images}
\vspace*{-0.1in}
\end{figure}

\textbf{Feature Clustering:}
We automate the detection of clusters of feature representations by leveraging  
the mean shift clustering algorithm~\cite{mean-shift}. An example of applying 
mean shift can be seen in \Cref{fig:meanshift-tsne-translated}, where the mean 
shift algorithm predicts three classes for the translated images, as generated 
by a backdoor-infected robust model. 
We further investigated the content inside these clusters by checking the images
associated with the feature representations that  make up these clusters. Specifically, 
the purple cluster ({\em see \Cref{fig:meanshift-tsne-translated}}) contained inputs seen in 
\Cref{fig:poisoned-model-translated-images}(a). These are the translated 
inputs which exhibit the backdoor. In contrast, the inputs seen in the 
yellow cluster ({\em \Cref{fig:meanshift-tsne-translated}}) contained translated 
images seen in \Cref{fig:poisoned-model-translated-images}(b). These images
correspond to the features of the actual training images in class 
{\em Sneaker}.

\section{Detailed Methodology}
\label{sec:methodology}




\textbf{Backdoor Injection:}
%
We show that despite being highly resilient to 
known adversarial attacks \cite{pgd-madry}, robust backdoor models are 
still susceptible to backdoor attacks. 
It takes very few poisoned training images (as little as 1\% for visible backdoors) for the backdoor to be successfully injected. 
We use backdoor injection techniques similar to the one seen in~\cite{BadNets} for visible backdoors and seen in~\cite{zhong2020backdoor} for invisible backdoors. We randomly select and poison one percent of the training images at random from each dataset (e.g. 500 images for CIFAR-10) for visible backdoor attacks and thirty percent (e.g. 15000 images for CIFAR-10) for invisible backdoors. 
The poisoning of 30\% of training images for invisible backdoors is in line with the configuration in Zhong et al.~\cite{zhong2020backdoor}. 
We poison these images by adding the respective backdoor trigger (visible 
or invisible) to the images and augment them to the training data.
Once this modified dataset is ready, we train the model using this data.

%
%

\begin{figure*}[h]
\begin{center}
\includegraphics[scale=0.3]
{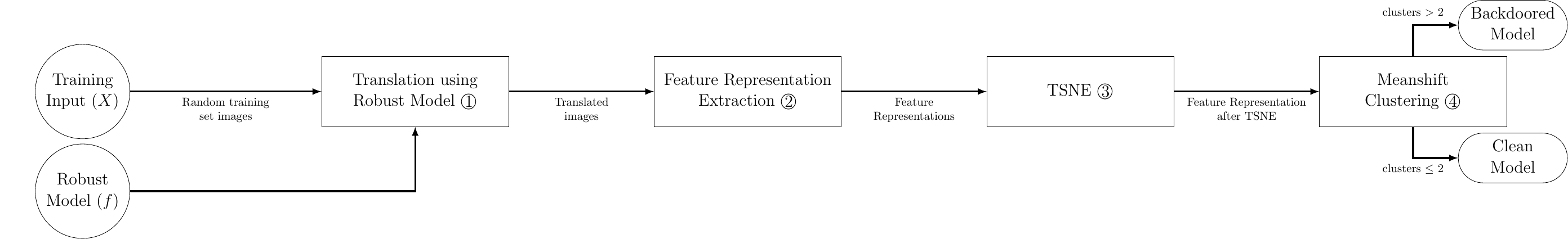}
\end{center}
\vspace*{-0.2in}
\caption{Overview of the detection technique}
\label{fig:method-block-diagram}
\end{figure*}

\smallskip \noindent
\textbf{Backdoored Model Detection:}
In this section, we elucidate the methodologies behind our detection 
technique \AG in detail. 
\AG only assumes white-box access to the model and access to the training 
data. It is important to note that \AG \textit{does not} have access to the 
poisoned data. In~\Cref{table:notation}, we introduce some notation to help 
us illustrate our approach.

\smallskip\noindent
%


\smallskip\noindent
\textbf{Backdoor detection:} First we provide a high level 
overview of \AG before going into each step in detail. Typically, the data 
points of a particular class follow a single 
distribution and as a result, form only one cluster after undergoing t-SNE
\cite{tsne}. However, when a backdoor attack is carried out, the adversary 
inadvertently injects 
a mixture of distributions in one class, resulting in more than one cluster.
The identification of a mixture distribution in a class is the main 
intuition behind our approach. 


The hypothesis is that the image generation process for robust models, as seen in Santurkar et al. \cite{robust-image-synthesis}, will follow similar distributions as the training
data. Since the target class in a backdoor model will be learning from multiple 
distributions, there will be multiple distributions of feature representation 
of the translated images (generated via first order adversarial methods). 
Our aim is to detect these multiple feature distributions. To detect such multiple 
distributions, we leverage t-SNE and Mean shift clustering. 
\revision{We also conduct an ablation study (RQ7 in \autoref{sec:results}) to compare the effectiveness of our design choices with closely-related alternative dimensionality reduction techniques and clustering methods.
}

For each label $y^{(i)}$ $\in$ $Y$, \Cref{alg:main-detection} 
generates translated images via first order-based adversarial methods 
({\em see \Cref{fig:method-block-diagram}} Step 1). Then, it extracts the feature 
representations from the training and translated 
images for the label $y^{(i)}$ ({\em see \Cref{fig:method-block-diagram}} Step 2). Next, the dimensions of the extracted features are reduced using t-SNE 
({\em see \Cref{fig:method-block-diagram}} Step 3). Mean shift 
is then employed to calculate the number of clusters in the reduced feature 
representations ({\em see \Cref{fig:method-block-diagram}} Step 4). Finally, the number of resulting clusters is used 
to flag the backdoor-infected model (and poisoned class) as suspicious, if 
necessary.

%

\begin{table}[t]
\vspace*{0.0in}
	\centering
	\label{table:notation}
	{\scriptsize
	{\begin{tabular}{|c| p{7cm} |}
	\hline
	$f$ & The robust machine learning classifier under test. \\ \hline
	$Y$ & Set of labels for $f$ \\ \hline
	$\mathbb{D}$ & The full training data \\ \hline
	$\mathcal{L}$ & The loss function \\ \hline
	$\mathcal{R}$ & A function that returns the feature representation flattened to 
	single 1D vector \\ \hline
	$X_{y^{(i)}}$ & Vector of training data 
	points for label $y^{(i)} \in Y$ \\ 	\hline
	$X'_{y^{(i)}}$ & Vector of translated data 
	points for label $y^{(i)} \in Y$\\ 	\hline	
\end{tabular}}}
\caption{Notations used in our approach}
\vspace*{-\baselineskip}
\end{table}

%

\begin{algorithm}[t]
   \caption{Backdoor Detection using \AG}
   \label{alg:main-detection}
   {\small
\begin{algorithmic}
   \STATE {\bfseries Input:} Robust ML classifier $f$, Sample of training
   data points $X$, Sample of translated data points $X'$, bandwidth for 
   the mean shift algorithm $b$
   \FOR{$y^{(i)}$ {\bfseries $\in$} $Y$}
   \STATE $\rhd \; \mathcal{R}$ returns the activations of the last hidden layer flattened to a single 1D vector
   
   \STATE $R_{X_{y^{(i)}}} = \mathcal{R}(f, X_{y^{(i)}})$ 
   \STATE $R_{X'_{y^{(i)}}} = \mathcal{R}(f, X'_{y^{(i)}})$
   \STATE $R_{y^{(i)}} = concatenate(R_{X_{y^{(i)}}}, \; R_{X'_{y^{(i)}}})$
   \STATE $\;$
   \STATE $\rhd \;tsne$ reduces the feature dimensions
   \STATE $\hat{R}_{y^{(i)}} = tsne(R_{y^{(i)}}, b)$ 
   \STATE $predicted\_classes = meanshift(\hat{R}_{y^{(i)}})$
   \STATE $analyseForBackdoor(\hat{R}_{y^{(i)}}, predicted\_classes)$
   
   \ENDFOR
\end{algorithmic}}
\end{algorithm}

The inclusion of the training images provides \AG with crucial information 
that is useful for the detection of backdoors. We note that the feature 
representation of backdoor images is distinct from the feature representations 
of {\em both the clean training images and translated images (without the 
backdoor trigger) associated with the class}. Consequently, adding the training 
images in the detection process helps us avoid false positives. In the absence 
of the training images, \AG would report a higher rate of false positives. 
An example of such false positives is seen 
in~\Cref{fig:false-positive-only-translated}. 

\smallskip\noindent
\textbf{Step 1 - Image to Image Translation:} To effectively analyse a model for backdoors, 
a vector of translated images $X'_{y^{(i)}}$ where $y^{(i)}$ $\in$ $Y$ needs to be built. 
In robust classifiers, image translation leads to perceptually aligned 
images~\cite{robust-image-synthesis}. This image translation is done
for all $y^{(i)}$ $\in$ $Y$. The following 
function is minimised \revise{using stochastic gradient descent} (and the probability of the target class $y^{(i)}$ is maximised):
\begin{equation}
x = \underset{{|| x' - x_0||}_2 \leq \epsilon}{arg\,min} \mathcal{L} (x', y^{(i)}), 
\,\,\,\,\, \,\,\,\,\, x_0  \in \mathbb{D}
\label{eqn:translation-actual}
\end{equation}

\AG samples a seed from the training data $\mathbb{D}$ and minimises the 
loss $\mathcal{L}$ of the particular label $y^{(i)}$ to generate the 
translated images ({\em see \Cref{fig:method-block-diagram}} Step 1). This is done across 500 random seed images to obtain 
$X'_{y^{(i)}}$. \revise{It is important to note that there is no 
constraint on the labels of the seed images.}

\smallskip\noindent
\textbf{Step 2 - Feature Representations:} Since \AG relies on the feature 
representations of the images, the algorithm now extracts them using $X_{y^{(i)}}$ 
and $X'_{y^{(i)}}$ for $y^{(i)}$ $\in$ $Y$. We define 
$\mathcal{R}$ as a function that maps an input $x$ to a
vector $\mathcal{R}(x, f)$ in the representation (penultimate layer) for a 
robust model $f$.
 
Once  $X_{y^{(i)}}$ and $X'_{y^{(i)}}$ are generated for $y^{(i)}$ $\in$ $Y$,
\AG runs a forward pass of all 
the inputs $x \in X_{y^{(i)}}$ and $x' \in X'_{y^{(i)}}$ through the robust model $f$. \AG
extracts the outputs of the last hidden layer and flattens them to form feature 
representations $R_{X_{y^{(i)}}}$ and $R_{X'_{y^{(i)}}}$, 
for $X_{y^{(i)}}$ and $X'_{y^{(i)}}$, respectively ({\em see 
\Cref{fig:method-block-diagram}} Step 2). 
These feature representations concatenated 
into $R_{y^{(i)}}$ for each $y^{(i)} \in Y$. 

\smallskip\noindent
\textbf{Step 3 - t-SNE:} First introduced in ~\cite{tsne}, t-distributed stochastic neighbour embedding (t-SNE) is a data 
visualisation technique . It is a nonlinear 
dimensionality reduction algorithm, which is primarily used to visualise high 
dimensional data in a two or three dimensional space. t-SNE is used to visualise 
the feature representations $R_{y^{(i)}}$ for all $y^{(i)} \in Y$ and to reduce their 
dimension ({\em see \Cref{fig:method-block-diagram}} Step 3). 
This is done to find any unusual clustering in the translated images. As expected, there are 
multiple clusters ($> 2$) of feature representations in the target class of a backdoored
model. As seen in \Cref{fig:backdoored-tsne} for a target class, the feature 
representations of the translated images show two clusters. This is because 
the learning process had inputs from two distributions (i.e. clean inputs and poisoned 
inputs). 
\revision{
We have selected t-SNE for \AG due to its ability to group data with little assumption about the data distribution. Furthermore, we compare the effectiveness of \AG with t-SNE  with a closely-related dimensionality reduction method (\AG with PCA~\cite{mackiewicz1993principal}) in RQ7 (\textit{see} \autoref{sec:results}).
}

\smallskip\noindent
\textbf{Step 4 - Detection using Mean shift:} 
To further automate the process of detection, the 
mean shift algorithm~\cite{mean-shift} is leveraged by \AG. 
This is a clustering algorithm which is 
used to identify the clusters automatically. Mean shift tries to locate the
modes of a density function. 
It does this by trying to discover "blobs" in a
smooth density of samples ({\em see \Cref{fig:method-block-diagram}} Step 4). 
It updates candidates for centroids to be a mean 
of points in a given region and then eliminates duplicates to form a final 
set of points \cite{mean-shift}. One can see in 
\Cref{fig:meanshift-tsne-translated}
that the algorithm identifies four classes. 
After the mean shift, all the classes that show multiple distributions 
(clusters $> 2$) in the translated images are flagged as suspicious. 
A user can examine the examples in the cluster as seen 
in \Cref{fig:poisoned-model-translated-images}, which helps the 
user to determine if the model was poisoned. 
\revision{
In addition, we compare the effectiveness of mean-shift clustering in \AG to closely-related clustering methods (such as affinity propagation~\cite{dueck2009affinity} and HDBSCAN~\cite{mcinnes2017hdbscan}) in RQ7 (\textit{see} \autoref{sec:results}). 
}

\section{Evaluation}
\label{sec:results}

In this section, we describe the experimental setup for backdoor injection attacks on adversarially robust DNN models, using three major classification tasks and several types of backdoor triggers. Overall, we employ four backdoor attack triggers including  localised and distributed visible triggers, as well as static and adversarial invisible triggers. 
We also present the 
empirical results of the 
effectiveness of the different backdoor injection attacks on robust DNN models, as well as the detection accuracy of \AG in exposing backdoor attacks in robust models. 

\smallskip\noindent
\textbf{Research questions:} We evaluate the success rate of backdoor injection attacks on adversarially robust models and the effectiveness of our detection technique (\AG). In particular, we ask the following research questions:

\begin{itemize}

\item \textbf{\RQ1 Attack Success Rate.} 
How effective are backdoor injection attacks on adversarially robust DNN models? How does the effectiveness of backdoor attacks in robust DNN models compare to that of standard DNN models (i.e., Robust vs Standard)? 

\item \textbf{\RQ2 Detection Effectiveness.} 
How effective is the proposed detection approach, i.e., \AG, in detecting all backdoor-infected models?

\item \textbf{\RQ3 Comparison to the state of the art.} How effective is \AG in comparison to the state of the art, i.e., Neural Cleanse (NC)? \revision{Is NC's performance 
sensitive to detection parameters, namely epsilon ($\epsilon$), and step size?}

\item \textbf{\RQ4 Sensitivity Analysis of Detection Parameters.} 
Is \AG sensitive to detection parameters, namely the epsilon ($\epsilon$), mean shift bandwidth, the random sampling of initial images and the number of initial seed images? 

\item \textbf{\RQ5 Attack Comparison.}  
What is the comparative performance of the different backdoor triggers in terms of attack success rate (i.e., localised vs distributed vs static perturbation vs adversarial perturbation)? 
Does the type or stealthiness (i.e., visibility) of backdoor triggers have an effect on \AG ' backdoor detection? 

\item \textbf{\RQ6 Detection Efficiency.} What is the performance of \AG, in terms of execution time? 
Is the detection efficiency of \AG influenced by the type or stealthiness of backdoor attack type?

\item \revision{
\textbf{\RQ7 Ablation Study.} 
What is the effect of our design choices on the effectiveness of \AG, in comparison to 
closely-related alternatives, in terms of visualization (t-SNE versus PCA) and clustering  (Mean Shift versus Affinity Propogation versus HDBSCAN)?		
}
\end{itemize}

\begin{table}[t]
  \begin{center}
      \caption{
      Dataset details and complexity of classification tasks}
  {\scriptsize
  \begin{tabular}{|l|r|r|r|r|r|} 
    \hline 
     \multicolumn{1}{|c|}{\textbf{Image}} & \multicolumn{1}{c|}{\textbf{Dataset}} & \multicolumn{1}{c|}{\multirow{2}{*}{\textbf{Arch.}}} & \multicolumn{1}{c|}{\textbf{Input}} & \multicolumn{2}{c|}{\textbf{\# of Images}}  \\ 
          \multicolumn{1}{|c|}{\textbf{Type}} & \multicolumn{1}{c|}{(\#labels)}  & & \multicolumn{1}{c|}{\textbf{Size}} & \multicolumn{1}{c|}{\textbf{training}} & \multicolumn{1}{c|}{\textbf{test}} \\ 
    \hline
     Objects &  CIFAR-10 (10)  & ResNet50  & 32 x 32 x 3 & 50,000 & 10,000  \\  
    \hline
	Digits &  MNIST (10)  & ResNet18 & 28 x 28 x 1 & 60,000 & 10,000  \\  
    \hline
     \multirow{2}{*}{\shortstack{Fashion\\Article}}  & \multirow{2}{*}{\shortstack{Fashion-\\MNIST (10)}}  & \multirow{2}{*}{ResNet18} & \multirow{2}{*}{28 x 28 x 1} & \multirow{2}{*}{60,000} & \multirow{2}{*}{10,000}  \\  
      &  & & & &  \\  
     \hline     
    \end{tabular}}
  \label{tab:datasets}   
\end{center}
\vspace{-\baselineskip}
\end{table}

\subsection{Experimental Setup}
\smallskip\noindent
\textbf{Evaluation setup:} 
Experiments were conducted on nine similar Virtual Machine (VM) instances on the Google Cloud platform, each VM is a PyTorch Deep Learning instance on an n1-highmem-4 machine (with 4 vCPU and 26 GB memory). Each VM had an Intel Broadwell CPU platform, 1 X NVIDIA Tesla GPU with eight to 16GB GPU memory and a 100 GB standard persistent disk. 


\smallskip\noindent
\textbf{Datasets and Models:} 
For our experiments, we use the CIFAR-10~\cite{krizhevsky2009learning}, MNIST~\cite{lecun1998mnist} and Fashion-MNIST~\cite{xiao2017online} datasets. MNIST and Fashion-MNIST have 60,000 training images each, while CIFAR-10 has 50,000 training images (\textit{see \Cref{tab:datasets}}). Each dataset has 10 classes and 10,000 test images. MNIST and Fashion-MNIST models were trained with the standard ResNet-18 architecture, while CIFAR-10 was trained using the standard ResNet-50 architecture~\cite{he2016deep}. 
\revision{In this work, we have used the ResNet architecture for training all datasets since it is the default architecture supported by our adversarial training approach~\cite{pgd-madry}.} 
All experiments were conducted with the default learning rate (LR) scheduling in the robustness package~\cite{robustness-lib}, i.e., the PyTorch StepLR optimisation scheduler. The learning rate is initially set to 0.1 for training (LR) and the scheduler decays the learning rate of each parameter group by 0.1 (gamma) every 50 epochs (default step size). All models were trained with momentum of 0.9 and weight decay of $5e^{-4}$. Only CIFAR-10 models were trained with data augmentation~\footnote{This is the default configuration in the robustness package for CIFAR-10}, with momentum of 0.9 and weight decay of $5e^{-4}$.

\begin{figure}[t]
\begin{center}
\begin{tabular}{cccccccc}
\includegraphics[scale=0.11]{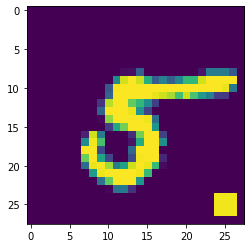} & 
\includegraphics[scale=0.11]{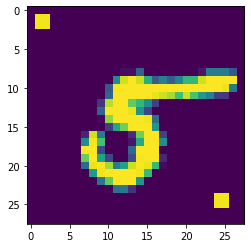} & &
\includegraphics[scale=0.11]{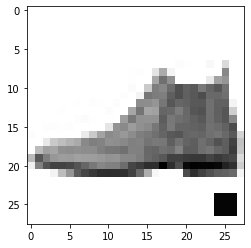} &
\includegraphics[scale=0.11]{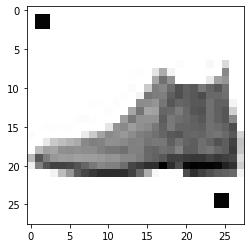} & &
\includegraphics[scale=0.11]{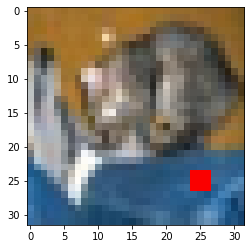} &
\includegraphics[scale=0.11]{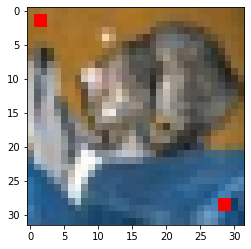}  \\ 
{\bf(a)} & {\bf(b)} & &  {\bf(c)}  & {\bf(d)} &  & {\bf(e)} & {\bf(f)} \\ 
\\

\end{tabular}
\end{center}
\vspace*{-0.15in}
\caption{
Visible Triggers for MNIST (a) localised and (b) distributed backdoors, Fashion-MNIST (c) localised and (d) distributed backdoors and CIFAR-10 \revision{(e) localised and (f) distributed} backdoors.
}
\label{fig:backdoor-triggers}
\end{figure}

\begin{figure}[t]
\begin{center}
\begin{tabular}{cccc}

\includegraphics[scale=0.15]{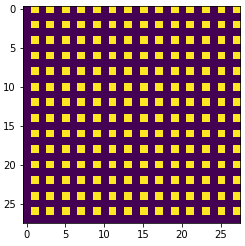} &
\includegraphics[scale=0.15]{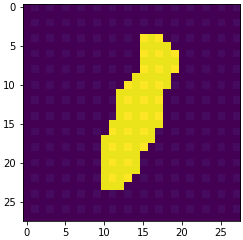} & 
\includegraphics[scale=0.15]{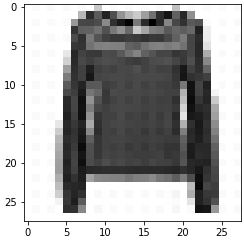} & 
\includegraphics[scale=0.15]{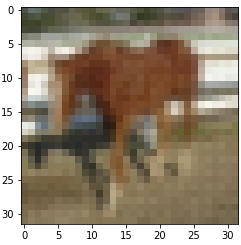} \\
{\bf(a)} & {\bf(b)} & {\bf(c)} & {\bf(d)} \\

\end{tabular}
\end{center}
\vspace*{-0.15in}
\caption{
Details of Static Invisible Backdoor Trigger for each dataset showing (a) the Static Invisible Trigger (note that the image intensity was increased by 25 fold to allow for visibility with the human eye), and example resulting poisoned images for (b) MNIST, (c) Fashion-MNIST and (d) CIFAR-10 showing that the poisoned image is not visible to the human eye 
}
\vspace*{-\baselineskip}
\label{fig:backdoor-triggers-invisible-pattern}
\end{figure}

\begin{table*}

\caption{Details of Training Time for Standard versus Robust models for each dataset, each backdoor trigger and clean models (in mins)}
\begin{center}
{\scriptsize
\begin{tabular}{ccccccccccccccccc|}
\cline {1-17}
\multicolumn{1}{|c}{} & \multicolumn{16}{|c|}{\textbf{TRAINING TIME (in mins)}} \\

\multicolumn{1}{|c}{}& \multicolumn{5}{|c|}{\textbf{MNIST}} & \multicolumn{5}{c|}{\textbf
 {Fashion-MNIST}}  & \multicolumn{5}{c|}{\textbf{CIFAR-10}} & \multicolumn{1}{c|}{\textbf{AVERAGE}}  \\
 
\multicolumn{1}{|c}{\textbf{Model}}& \multicolumn{4}{|c|}{\textbf{Backdoor-Infected}} &\multicolumn{1}{c|}{}& \multicolumn{4}{c|}{\textbf{Backdoor-Infected}} & \multicolumn{1}{c|}{}   & \multicolumn{4}{c|}{\textbf{Backdoor-Infected}} & \multicolumn{1}{c|}{} & \multicolumn{1}{c|}{All}  \\
 
\multicolumn{1}{|c}{\textbf{Type}} & \multicolumn{2}{|c|}{\textbf{Visible}} & \multicolumn{2}{c|}{\textbf{Invisible}} & \multicolumn{1}{c|}{\textbf{Clean}} & \multicolumn{2}{c|}{\textbf{Visible}} & \multicolumn{2}{c|}{\textbf{Invisible}} & \multicolumn{1}{c|}{\textbf{Clean}}  & \multicolumn{2}{c|}{\textbf{Visible}} & \multicolumn{2}{c|}{\textbf{Invisible}} & \multicolumn{1}{c|}{\textbf{Clean}} & \multicolumn{1}{c|}{(Clean/}  \\ 

\multicolumn{1}{|c}{}& \multicolumn{1}{|c|}{\textit{Local}} & \multicolumn{1}{ c|}{\textit{Dist}} &\multicolumn{1}{c|}{\textit{Static}} & \multicolumn{1}{ c|}{\textit{Adv}}&
\multicolumn{1}{c|}{}
& \multicolumn{1}{c|}{\textit{Local}} & \multicolumn{1}{ c|}{\textit{Dist}} &\multicolumn{1}{c|}{\textit{Static}} & \multicolumn{1}{ c|}{\textit{Adv}} &
\multicolumn{1}{c|}{}
& \multicolumn{1}{c|}{\textit{Local}} & \multicolumn{1}{ c|}{\textit{Dist}} &\multicolumn{1}{c|}{\textit{Static}} & \multicolumn{1}{ c|}{\textit{Adv}} & \multicolumn{1}{c|}{} & \multicolumn{1}{c|}{\textit{Backdoor-infected})} \\ \hline 

\multicolumn{1}{|c}{\textbf{\makecell{Robust}}} &
\multicolumn{1}{|c}{2971} & \multicolumn{1}{c|}{1321} & \multicolumn{1}{c}{242} & \multicolumn{1}{c|}{220} &  \multicolumn{1}{c|}{1800} &
\multicolumn{1}{c}{2971}  & \multicolumn{1}{c|}{162} &
\multicolumn{1}{c}{109}  & \multicolumn{1}{c|}{132} & \multicolumn{1}{c|}{3031} &
\multicolumn{1}{c}{1871}  & \multicolumn{1}{c|}{3276} &
\multicolumn{1}{c}{1183}  & \multicolumn{1}{c|}{948} & \multicolumn{1}{c|}{1882} & \multicolumn{1}{c|}{1475 (2238/\textit{1284})} \\ \hline 

\multicolumn{1}{|c}{\textbf{\makecell{Standard}}} &
\multicolumn{1}{|c}{20} & \multicolumn{1}{c|}{45} & \multicolumn{1}{c}{3} & \multicolumn{1}{c|}{2} &  \multicolumn{1}{c|}{22} &
\multicolumn{1}{c}{50}  & \multicolumn{1}{c|}{62} &
\multicolumn{1}{c}{2}  & \multicolumn{1}{c|}{1} & \multicolumn{1}{c|}{41} &
\multicolumn{1}{c}{141}  & \multicolumn{1}{c|}{172} &
\multicolumn{1}{c}{108}  & \multicolumn{1}{c|}{66} & \multicolumn{1}{c|}{135} & \multicolumn{1}{c|}{58 (66/\textit{56})}  \\ 
\hline

\label{tab:backdoor-training-time}	
\end{tabular}}
\end{center}
\end{table*}

\begin{figure}[t]
\begin{center}
\begin{tabular}{ccccccc}
\includegraphics[scale=0.12]{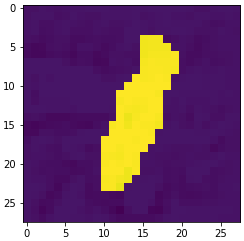} & 
\includegraphics[scale=0.12]{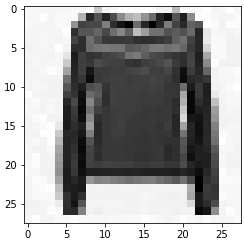} &
\includegraphics[scale=0.12]{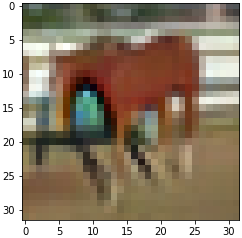}  &&
\includegraphics[scale=0.12]{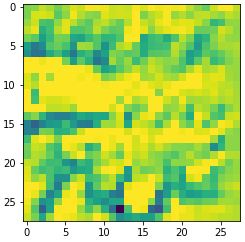} & 
\includegraphics[scale=0.12]{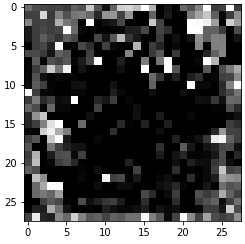} &
\includegraphics[scale=0.12]{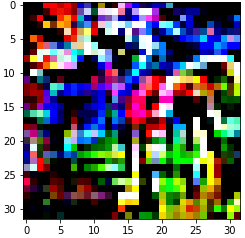}  \\ 

{\bf(a)} & {\bf(b)} & {\bf(c)} &&
{\bf(d)} & {\bf(e)} & {\bf(f)} \\
\end{tabular}
\end{center}
\vspace*{-0.15in}
\caption{
Poisoned images for Invisible Adversarial backdoors for (a) MNIST, 
(b) Fashion-MNIST and (c) CIFAR-10 datasets, with their corresponding adversarial triggers (shown in d, e, f), note that the intensity of the triggers were increased by 10 fold to be visible to the human eye 
}
\vspace*{-\baselineskip}
\label{fig:backdoor-triggers-invisible-adv}
\end{figure}

\smallskip\noindent
\textbf{Adversarial Training:} 
Some approaches have been proposed to guarantee adversarial training of machine learning models \cite{pgd-madry, wong2018scaling, wong2018provable, sinha2017certifying, raghunathan2018certified}. Notably, Wong et al~\cite{wong2018scaling,wong2018provable} aim to train models that are provably robust against norm-bounded adversarial perturbations on the training data. Sinha et al.~\cite{sinha2017certifying} and Raghunathan et al. ~\cite{raghunathan2018certified} are focused on training and guaranteeing the  performance of ML models under adversarial input perturbations. However, the aforementioned approaches either consider very small adversarial perturbation budget epsilon ($\epsilon$), do not scale to larger neural nets or datasets (beyond MNIST) or have a huge computational overhead.

In this paper, we apply the robust optimization approach proposed by Madry et al.~\cite{pgd-madry} for adversarial training. 
In particular, it is computationally feasible, it provides security guarantees against a wider range of adversarial perturbations and it scales to large networks and datasets (such as CIFAR-10).
For our evaluation, all models were trained with robust optimisation 
based on the adversarial training approach~\cite{pgd-madry} with an $l_2$ perturbation set. The 
parameters for robust training are the same for all datasets (\textit{see \Cref{tab:hyperparameters} in \Cref{sec:additional-tabs}}). In particular, all models were trained 
with an adversarial attack budget of $0.5$ ($\epsilon$), and an attack step size of $1.5$ (step size) 
and set to take $20$ steps (\# steps) during adversarial attack. All other hyperparameters are set 
to the default hyperparameters in the robustness package~\cite{robustness-lib}. No hyperparameter 
tuning was performed for the adversarial training of models.

\smallskip\noindent
\textbf{Training Time:} 
\autoref{tab:backdoor-training-time} highlights the average training time for each dataset, model type and backdoor attack trigger. \textit{Robust model training is expensive, it is significantly more expensive to train a robust model than a standard model}. Robust training time is 25 times as much as that of standard model training. It took about 25 hours (1,475 minutes) to train a robust model and less than an hour (58 minutes) to train a standard model, on average ({\em see \Cref{tab:backdoor-training-time}}). For backdoor-infected models, robust training time is 23 times as much as that of standard training time, on average. \autoref{tab:backdoor-training-time} shows that it took about 21 hours (1,284 minutes) to train a robust backdoor-infected model and less than an hour (56 minutes) to train a standard backdoor-infected model. Meanwhile, robust training time is 34 times as much as standard training time for clean models. In particular, it took about 37 hours (2,238 minutes) to train a robust clean model and about an hour (66 minutes) to train standard clean model ({\em see \Cref{tab:backdoor-training-time}}). Generally, \textit{it is slightly cheaper to inject a backdoor in a model than to train a clean model}. In our experiments, it is less expensive to train a backdoor-infected model in comparison to a clean model. 

\smallskip\noindent
\textbf{Adversarial Accuracy:} Adversarial evaluation was performed with the same 
parameters as adversarial training for all datasets and models. In particular, all 
classifiers were evaluated with an adversarial attack budget of $0.5$ ($\epsilon$), 
and an attack step size of $1.5$ and set to take $20$ steps during adversarial attack. 
In addition, for adversarial evaluation, we use the best loss in PGD step as the attack 
(``use\_best": {\em True}), with no random restarts (``random\_restarts": 0) and no fade 
in epsilon along epochs (``eps\_fadein\_epochs": 0). 
\autoref{tab:attack-success-and-accuracy} shows the average adversarial accuracy of our clean and backdoor-infected trained models for each dataset. 
\textit{In our evaluation, adversarial training accuracy is not inhibited by the backdoor attack vector.}
All trained robust models maintained a similarly high adversarial accuracy for both clean and 
backdoor-infected models. Specifically, \Cref{tab:attack-success-and-accuracy} shows that \textit{backdoor-infected robust models have 83.21\%  adversarial accuracy}, on average. In contrast, \textit{clean robust models have a slightly higher adversarial precision of 86.37\%}, on average (\textit{see \Cref{tab:attack-success-and-accuracy}}).

\smallskip\noindent
\textbf{Visible Backdoor Triggers:}
For visible backdoor triggers, we employed the backdoor data poisoning approach outlined in BadNets~\cite{BadNets} to inject backdoors during adversarial training. 
For all datasets, we created a set of backdoor infected images by modifying a portion of the training datasets, specifically we apply a trigger to one percent (1\%) of the clean images in the training set (e.g., 600 images for the MNIST dataset). Additionally, we modify the class label of each poisoned image to class seven (7) for all datasets and all attack types, then we train DNN models with the modified training data to 100 epochs for Fashion-MNIST and MNIST, and 110 epochs for CIFAR-10. 


\smallskip\noindent
\textbf{Invisible Backdoor Triggers:} 
We employed the technique
described in Zhong et al.~\cite{zhong2020backdoor} to construct two types of invisible 
backdoors, namely static and adversarial backdoors ({\em see \Cref{fig:backdoor-triggers-invisible-pattern}, \Cref{fig:backdoor-triggers-invisible-adv}}). 
To allow for a reasonable attack success rate for the invisible triggers, we created a set of backdoor infected images for each dataset by modifying 
30 percent (30\%) of the clean images in the training set (e.g., 18,000 images for the MNIST dataset) and 
modifying the class label of each poisoned image to class seven (7). The poisoning of 30\% of training images for invisible backdoors is in line with the configuration in Zhong et al.~\cite{zhong2020backdoor}. 
We then train DNN models with the modified training data to 100 epochs for Fashion-MNIST and MNIST, and 110 epochs for CIFAR-10. 
\revise{
For this attack, we employ the most stealthy invisible triggers (least intensity), which has a lower attack success rate (ASR). 
The resulting ASR 
is in line
with the results seen in \cite{zhong2020backdoor} which shows ASRs as 
low as about 30\% for static backdoors with the least intensity and about 55\% for adversarial backdoors with the least intensity.
}

\smallskip\noindent
\textbf{Attack Configuration:} 
The triggers for each visible backdoor attack and tasks are shown in {\Cref{fig:backdoor-triggers}}. The 
trigger for localised backdoors is a square at the bottom right corner of the image, this 
is to avoid covering the important parts of the original training image. The trigger for 
distributed backdoors is made up of two smaller squares, one at the top left corner of the 
image and another at the bottom right corner. The total size of the trigger is less than 
one percent of the entire image for both of these visible backdoor triggers. 

For the invisible attacks the triggers are seen in 
\Cref{fig:backdoor-triggers-invisible-pattern} and 
\Cref{fig:backdoor-triggers-invisible-adv}. The static backdoor trigger
is seen in \Cref{fig:backdoor-triggers-invisible-pattern} (a). It is 
important to note that the trigger image is enhanced to view the trigger with 
ease. The actual poisoned images for the invisible static backdoor attack
are seen in \Cref{fig:backdoor-triggers-invisible-pattern} (b, c, d).
Similarly, we use the adversarial perturbation-based invisible 
backdoor attack described in Zhong et al.~\cite{zhong2020backdoor} to generate invisible 
backdoors which are adversarial in nature. The images with backdoor 
trigger for MNIST, Fashion-MNIST and CIFAR-10 are seen in 
\Cref{fig:backdoor-triggers-invisible-adv} (a, b, c) and the
enhanced triggers are seen in \Cref{fig:backdoor-triggers-invisible-adv} 
(d, e, f) respectively.

\begin{table*}[t]

\caption{Details of Attack success rate (ASR), classification accuracy and adversarial precision for each dataset, each backdoor trigger and clean models}
\begin{center}
{
\scriptsize 
\begin{tabular}{cccccccccccccc|}
\cline {1-14}


\multicolumn{1}{|c}{}  & \multicolumn{1}{|c}{}  & \multicolumn{1}{|c}{}  & \multicolumn{5}{|c|}{}  & \multicolumn{6}{c|}{\textbf{AVERAGE}}   \\

\multicolumn{1}{|c}{\textbf{Model}} & \multicolumn{1}{|c}{\textbf{Dataset}} & \multicolumn{1}{|c}{\textbf{Measure}} & \multicolumn{4}{|c|}{\textbf{Backdoor-Infected}} &\multicolumn{1}{c|}{\textbf{Clean}}& \multicolumn{5}{c|}{\textbf{Backdoor-Infected}} & \multicolumn{1}{c|}{\textbf{Clean}} 
\\
 
\multicolumn{1}{|c}{\textbf{Type}} &  \multicolumn{1}{|c}{}  & \multicolumn{1}{|c}{}  & \multicolumn{2}{|c|}{\textbf{Visible}} & \multicolumn{2}{c|}{\textbf{Invisible}} & \multicolumn{1}{c|}{\textbf{}} & \multicolumn{2}{c|}{\textbf{Visible}} & \multicolumn{2}{ c|}{\textbf{Invisible}} & \multicolumn{1}{ c|}{\textbf{All}} & \multicolumn{1}{c|}{\textbf{}} 
\\

\multicolumn{1}{|c}{} & \multicolumn{1}{|c}{}  &  \multicolumn{1}{|c}{}  & \multicolumn{1}{|c|}{\textit{Local}} & \multicolumn{1}{ c|}{\textit{Dist}} &\multicolumn{1}{c|}{\textit{Static}} & \multicolumn{1}{ c|}{\textit{Adv}}&
\multicolumn{1}{c|}{}
& \multicolumn{1}{c|}{\textit{Local}} & \multicolumn{1}{ c|}{\textit{Dist}} &\multicolumn{1}{c|}{\textit{Static}} & \multicolumn{1}{ c|}{\textit{Adv}} &
\multicolumn{1}{c|}{} & \multicolumn{1}{c|}{} 
\\ \hline

\multicolumn{1}{|c}{\multirow{6}{*}{\shortstack{\textbf{Robust} \\ \textbf{Models} \\ \empty  }}} &  \multicolumn{1}{|c}{\shortstack{  \textbf{MNIST}  \\ \empty \\\empty }}  & \multicolumn{1}{|c}{\shortstack{ASR \\ \textit{Class. Acc.} \\ \textit{(Adv. Prec.)}}} &
\multicolumn{1}{|c|}{\shortstack{99.96 \\\textit{ 99.59} \\ \textit{(99.51)}}} & \multicolumn{1}{c|}{\shortstack{100.00 \\ 99.53 \\ \textit{(99.49)}}} & \multicolumn{1}{c|}{\shortstack{37.53 \\\textit{98.94} \\ \textit{(97.72)}}} & \multicolumn{1}{c|}{\shortstack{59.87 \\\textit{98.31} \\ \textit{(97.27)}}} &  \multicolumn{1}{c|}{\shortstack{ N/A \\\textit{99.61} \\ \textit{(99.55)}}} & \multicolumn{1}{c|}{\multirow{6}{*}{\shortstack{92.93 \\\textit{93.74} \\ \textit{(86.18)}}}} & \multicolumn{1}{c|}{\multirow{6}{*}{\shortstack{99.87 \\\textit{93.85} \\ \textit{(86.11)}}}}  & \multicolumn{1}{c|}{\multirow{6}{*}{\shortstack{30.65 \\\textit{89.71} \\ \textit{(80.79)}}}}  & \multicolumn{1}{c|}{\multirow{6}{*}{\shortstack{47.86 \\\textit{88.89} \\ \textit{(79.77)}}}}  & \multicolumn{1}{c|}{\multirow{6}{*}{\shortstack{67.83 \\\textit{91.55} \\ \textit{(83.21)}}}} & \multicolumn{1}{c|}{\multirow{6}{*}{\shortstack{N/A \\\textit{93.96} \\ \textit{(86.37)}}}} 
\\ \cline {2-8} 

\multicolumn{1}{|c}{} & \multicolumn{1}{|c}{\shortstack{  \textbf{Fashion-MNIST}  \\ \empty \\ \empty }} & \multicolumn{1}{|c}{\shortstack{ASR \\ \textit{Class. Acc.} \\ \textit{(Adv. Prec.)}}} &
\multicolumn{1}{|c|}{\shortstack{96.26 \\\textit{91.83} \\ \textit{(90.78)}}}  & \multicolumn{1}{c|}{\shortstack{99.77 \\\textit{91.8} \\ \textit{(90.66)}}} &
\multicolumn{1}{c|}{\shortstack{33.33 \\\textit{88.38} \\ \textit{(83.56)}}}  & \multicolumn{1}{c|}{\shortstack{61.00 \\\textit{87.99} \\ \textit{(80.66)}}} & \multicolumn{1}{c|}{\shortstack{N/A \\\textit{91.99} \\ \textit{(90.91)}}} & \multicolumn{1}{c|}{} & \multicolumn{1}{c|}{} & \multicolumn{1}{c|}{} & \multicolumn{1}{c|}{} & \multicolumn{1}{c|}{} & \multicolumn{1}{c|}{}  \\ \cline {2-8} 

\multicolumn{1}{|c}{} & \multicolumn{1}{|c}{\shortstack{  \textbf{CIFAR-10}  \\ \empty \\ \empty }} & \multicolumn{1}{|c}{\shortstack{ASR \\ \textit{Class. Acc.} \\ \textit{(Adv. Prec.)}}} &
\multicolumn{1}{|c|}{\shortstack{82.58 \\\textit{89.8} \\ \textit{(68.26)}}}  & \multicolumn{1}{c|}{\shortstack{99.85 \\\textit{90.22} \\ \textit{(68.17)}}} &
\multicolumn{1}{c|}{\shortstack{21.08 \\\textit{81.82} \\ \textit{(61.1)}}}  & \multicolumn{1}{c|}{\shortstack{22.72 \\\textit{80.38} \\ \textit{(61.37)}}} & \multicolumn{1}{c|}{\shortstack{N/A \\\textit{90.28} \\ \textit{(68.64)}}} & \multicolumn{1}{c|}{} & \multicolumn{1}{c|}{} & \multicolumn{1}{c|}{} & \multicolumn{1}{c|}{} & \multicolumn{1}{c|}{} & \multicolumn{1}{c|}{}  \\ \hline

\multicolumn{1}{|c}{\multirow{6}{*}{\shortstack{ \textbf{Standard} \\ \textbf{Models} \\ \empty  }}} & \multicolumn{1}{|c}{\shortstack{  \textbf{MNIST}  \\ \empty \\\empty }}  & \multicolumn{1}{|c}{\shortstack{ASR \\ \textit{Class. Acc.} \\ \textit{(Adv. Prec.)}}} & 
\multicolumn{1}{|c|}{\shortstack{99.97 \\\textit{99.57} \\ \textit{(99.1)}}} & \multicolumn{1}{c|}{\shortstack{99.96 \\\textit{99.53} \\ \textit{(99.18)}}} & \multicolumn{1}{c|}{\shortstack{38.59 \\\textit{97.5} \\ \textit{(94.92)}}} &
\multicolumn{1}{c|}{\shortstack{25.3 \\\textit{98.06} \\ \textit{(96.55)}}}  & \multicolumn{1}{c|}{\shortstack{N/A \\\textit{99.53} \\ \textit{(99.13)}}} & \multicolumn{1}{c|}{\multirow{6}{*}{\shortstack{98.87 \\\textit{94.95} \\ \textit{(58.98)}}}}  & \multicolumn{1}{c|}{\multirow{6}{*}{\shortstack{99.91 \\\textit{95.12} \\ \textit{(61.94)}}}}  & \multicolumn{1}{c|}{\multirow{6}{*}{\shortstack{60.36 \\\textit{91.83} \\ \textit{(55.33)}}}}  & \multicolumn{1}{c|}{\multirow{6}{*}{\shortstack{44.29 \\\textit{89.05} \\ \textit{(54.70)}}}}  & \multicolumn{1}{c|}{\multirow{6}{*}{\shortstack{75.86 \\\textit{92.74} \\ \textit{(57.74)}}}} & \multicolumn{1}{c|}{\multirow{6}{*}{\shortstack{N/A \\\textit{95.13} \\ \textit{(58.48)}}}} \\ \cline {2-8} 

\multicolumn{1}{|c}{} & \multicolumn{1}{|c}{\shortstack{  \textbf{Fashion-MNIST}  \\ \empty \\ \empty }} & \multicolumn{1}{|c}{\shortstack{ASR \\ \textit{Class. Acc.} \\ \textit{(Adv. Prec.)}}} &
\multicolumn{1}{|c|}{\shortstack{97.5 \\\textit{91.11} \\ \textit{(75.99)}}} &  \multicolumn{1}{c|}{\shortstack{99.81 \\\textit{91.35} \\ \textit{(86.44)}}} &
\multicolumn{1}{c|}{\shortstack{43.47 \\\textit{86.28} \\ \textit{(69.46)}}}  & \multicolumn{1}{c|}{\shortstack{54.56 \\\textit{86.29} \\ \textit{(66.43)}}} & \multicolumn{1}{c|}{\shortstack{N/A \\\textit{91.43} \\ \textit{(76.29)}}} & \multicolumn{1}{c|}{} & \multicolumn{1}{c|}{} & \multicolumn{1}{c|}{} & \multicolumn{1}{c|}{} & \multicolumn{1}{c|}{} & \multicolumn{1}{c|}{} 
\\ \cline {2-8} %

\multicolumn{1}{|c}{} &\multicolumn{1}{|c}{\shortstack{  \textbf{CIFAR-10}  \\ \empty \\ \empty }} & \multicolumn{1}{|c}{\shortstack{ASR \\ \textit{Class. Acc.} \\ \textit{(Adv. Prec.)}}} &
\multicolumn{1}{|c|}{\shortstack{99.14 \\\textit{94.18} \\ \textit{(1.86)}}}  & \multicolumn{1}{c|}{\shortstack{99.97 \\\textit{94.47} \\ \textit{(0.2)}}} &
\multicolumn{1}{c|}{\shortstack{99.02 \\\textit{91.72} \\ \textit{(1.62)}}}  & \multicolumn{1}{c|}{\shortstack{53.02 \\\textit{82.79} \\ \textit{(1.13)}}} & \multicolumn{1}{c|}{\shortstack{N/A \\\textit{94.42} \\ \textit{(0.01)}}} & \multicolumn{1}{c|}{} & \multicolumn{1}{c|}{} & \multicolumn{1}{c|}{} & \multicolumn{1}{c|}{} & \multicolumn{1}{c|}{} & \multicolumn{1}{c|}{} 
\\ 
\hline

\label{tab:attack-success-and-accuracy}	
\end{tabular}}
\end{center}
\end{table*}

\smallskip\noindent
\textbf{Detection Configuration:} 
%
The detection configuration used in our evaluation are shown in \Cref{tab:detection-params} (\Cref{sec:additional-tabs}). 
\revision{For all datasets, we have conducted a preliminary controlled experiment of detection parameters (\textit{see} RQ4). This is to determine the best parameter for backdoor detection using \AG, without over-fitting.}  
For each dataset, the epsilon ($\epsilon$) ball for input perturbation is fixed. For MNIST and Fashion-MNIST, the parameter $\epsilon$ is 100 and it is 500 for CIFAR-10. This places a uniform limit on input perturbation for each dataset. The perplexity for t-SNE is a tuneable parameter that balances the attention between the local and global aspects of the data. The authors suggest a value between five and 50~\cite{tsne} and as a result we chose 30. 
The bandwidth in the mean shift algorithm is the size of the kernel function. This value is constant for each dataset, it is automatically computed with the scikit-learn mean shift clustering algorithm.~\footnote{\url{https://scikit-learn.org/stable/modules/generated/sklearn.cluster.estimate_bandwidth.html}} 
For the backdoor attacks, the resulting bandwidths are 35, 28 and 21 for MNIST, Fashion-MNIST and CIFAR-10, respectively. Additionally, we also test the 
sensitivity of the \AG technique to variance in the bandwidth, and (the number of) initial seed images (\textit{see} \RQ4). For instance, we run \AG with 
$\pm$ 3 around the respective calculated values for mean shift bandwidth. 
\smallskip\noindent
\textbf{Evaluation Metrics:}
We measure the performance of the backdoor injection attack by computing the \textit{classification accuracy} on the testing data. We compute the \textit{attack success rate} (ASR) by applying the trigger to all test images and measuring the number of modified images that are classified to the attack target label, i.e., classified to class seven (7). 
We also measure the \textit{adversarial precision} of all robust models. In addition, we measure the classification accuracy of the clean adversarially robust models as a baseline for comparison. We also compare the performance of robust models (i.e., ASR and classification accuracy) to that of standard backdoored (and clean) models. For detection efficacy, we report the \textit{number of feature representation clusters} found for all classes of all robust models. 

\subsection{Experimental Results}
\smallskip\noindent
\textbf{\RQ1 - Attack Success Rate (ASR):} 

In this section, we present the effectiveness of backdoor injection attack. We illustrate that backdoors can be effectively injected in robust models without significantly reducing the classification accuracy and adversarial precision of the models. \autoref{tab:attack-success-and-accuracy} highlights the attack success rate (ASR), classification accuracy and adversarial precision of each trained model.

In our evaluation, we found that \textit{robust models are highly vulnerable to backdoor attacks}. Backdoor attacks effectively caused the misclassification of 67.8\% of backdoor-infected images to the attacker selected target labels, across all datasets and attack types (\textit{see \autoref{tab:attack-success-and-accuracy}}). \textit{Visible backdoor triggers are generally more effective than invisible backdoor triggers}, visible triggers are 2.5 times more successful than invisible triggers (\textit{see \textit{attack success rate} (``ASR'') in \autoref{tab:attack-success-and-accuracy}}). Specifically, visible triggers effectively caused the misclassification of 96.4\% of backdoor-infected images to the attacker selected target labels, in comparison, invisible triggers caused the misclassification of only 39.3\% of infected images to the target class (\textit{see \autoref{tab:attack-success-and-accuracy}}). These results suggest that backdoor injection attacks are highly effective on robust models. 

\begin{result}
Robust DNNs are highly susceptible to backdoor attacks, \\with a 67.8\% attack success rate (ASR), on average.
\end{result}




\revision{In our experiments on robust optimization via adversarial training (AT), we observed that \textit{robust models are less susceptible to backdoor attacks than standard models.}
Backdoor attacks are more successful on standard models than robust models because adversarial perturbations introduced during adversarial training may influence the shape and dimension of the backdoor trigger.} We found that a backdoor attack is 12\% more effective on a standard DNN model than on a robust model, 
with ASR of 67.83\% and 75.86\% for a robust and standard backdoor-infected model, on average, respectively (\textit{see \autoref{tab:attack-success-and-accuracy}}). This result holds across attack types and regardless of the stealthiness (or visibility) of the backdoor trigger. For instance, the ASR for invisible static perturbations is 30.7\% on robust models, in comparison to 60.4\% on standard models. \revision{Our results imply that backdoor attacks are more effective in a standard model than a robust model (resulting from AT).}

\begin{result}
Backdoor attacks are (12\%) more effective on standard DNN models than robust models \revision{obtained via adversarial training}.
\end{result}

\textit{Backdoor injection in robust DNNs does not cause a significant reduction in adversarial precision}. Backdoor injection in robust models only reduced adversarial precision by about 3.7\%, in comparison to clean robust models. Backdoor-infected robust models have an adversarial precision of 83.21\% on average, while clean robust models have an adversarial precision of 86.37\% on average (\textit{see ``Adv. Prec.'' in \autoref{tab:attack-success-and-accuracy}}). In particular, the adversarial precision of robust models injected with visible triggers (86.14\%) is comparable to that of clean robust models (86.37\%). This result suggests that backdoor injection has little or no effect on the adversarial precision of infected robust models. 

\begin{result}
Backdoors do not significantly reduce the adversarial precision of robust models, they caused only 3.7\% reduction, on average. 
\end{result}

\textit{In our evaluation, backdoor injection in robust DNNs does not cause a significant reduction in classification accuracy for clean images.} Overall, backdoor-infected robust models have about 2.6\% reduction in classification accuracy in comparison to clean robust models, on average. Despite backdoor injection, robust models still achieved a high classification accuracy (91.55\%) for clean images, on average (\textit{see ``Class. Acc.'' in \autoref{tab:attack-success-and-accuracy}}). In comparison, clean robust models achieved a 93.96\% classification accuracy. This is not a significant reduction in classification accuracy. In particular, 
models trained with visible triggers maintained a higher classification accuracy than models trained with invisible triggers. 
Models trained with visible triggers had a classification accuracy of 93.80\% while 
models trained with invisible triggers had a lower classification accuracy of 89.30\% (\textit{see \Cref{tab:attack-success-and-accuracy}}). These results imply that backdoor injection in robust models does not significantly influence the classification accuracy of clean images. 

\begin{result}
Robust backdoor-infected models maintain a high classification accuracy (83.21\%), on average. 
\end{result}

\begin{table*}

\caption{Backdoor Detection Efficacy: \cmark~indicates that \AG detected a backdoored-infected model/class and \xmark~indicates that \AG did not (or failed to) detect the presence of a backdoored model/class, e.g., in clean models (or stealthy static invisible backdoor-infected models)
}
\begin{center}
{\scriptsize
\begin{tabular}{cccccccccccccccc|}
\cline {2-16}
& \multicolumn{5}{|c|}{\textbf{MNIST}} & \multicolumn{5}{c|}{\textbf
 {Fashion-MNIST}}  & \multicolumn{5}{c|}{\textbf{CIFAR-10}}  \\
 
& \multicolumn{4}{|c|}{\textbf{Backdoor-Infected}} &\multicolumn{1}{c|}{}& \multicolumn{4}{c|}{\textbf{Backdoor-Infected}} & \multicolumn{1}{c|}{}   & \multicolumn{4}{c|}{\textbf{Backdoor-Infected}} &  \\
 
 & \multicolumn{2}{|c|}{\textbf{Visible}} & \multicolumn{2}{c|}{\textbf{Invisible}} & \multicolumn{1}{c|}{\textbf{Clean}} & \multicolumn{2}{c|}{\textbf{Visible}} & \multicolumn{2}{ c|}{\textbf{Invisible}} & \multicolumn{1}{c|}{\textbf{Clean}}  & \multicolumn{2}{c|}{\textbf{Visible}} & \multicolumn{2}{c|}{\textbf{Invisible}} & \textbf{Clean} \\

& \multicolumn{1}{|c|}{\textit{Local}} & \multicolumn{1}{ c|}{\textit{Dist}} &\multicolumn{1}{c|}{\textit{Static}} & \multicolumn{1}{ c|}{\textit{Adv}}&
\multicolumn{1}{c|}{}
& \multicolumn{1}{c|}{\textit{Local}} & \multicolumn{1}{ c|}{\textit{Dist}} &\multicolumn{1}{c|}{\textit{Static}} & \multicolumn{1}{ c|}{\textit{Adv}} &
\multicolumn{1}{c|}{}
& \multicolumn{1}{c|}{\textit{Local}} & \multicolumn{1}{ c|}{\textit{Dist}} &\multicolumn{1}{c|}{\textit{Static}} & \multicolumn{1}{ c|}{\textit{Adv}} & \\ \hline

\multicolumn{1}{|c}{\textbf{\makecell{Backdoor\\Detection}}} &
\multicolumn{1}{|c}{\cmark} & \multicolumn{1}{c|}{\cmark} & \multicolumn{1}{c}{\xmark} & \multicolumn{1}{c|}{\cmark} &  \multicolumn{1}{c|}{\xmark} &
\multicolumn{1}{c}{\cmark}  & \multicolumn{1}{c|}{\cmark} &
\multicolumn{1}{c}{\cmark}  & \multicolumn{1}{c|}{\cmark} & \multicolumn{1}{c|}{\xmark} &
\multicolumn{1}{c}{\cmark}  & \multicolumn{1}{c|}{\cmark} &
\multicolumn{1}{c}{\cmark}  & \multicolumn{1}{c|}{\cmark} & \multicolumn{1}{c|}{\xmark} \\ \hline

\multicolumn{1}{|c}{\textbf{\makecell{Backdoor Class\\Detection}}} &
\multicolumn{1}{|c}{\cmark} & \multicolumn{1}{c|}{\cmark} & \multicolumn{1}{c}{\xmark} & \multicolumn{1}{c|}{\xmark} &  \multicolumn{1}{c|}{\xmark} &
\multicolumn{1}{c}{\cmark}  & \multicolumn{1}{c|}{\cmark} &
\multicolumn{1}{c}{\cmark}  & \multicolumn{1}{c|}{\cmark} & \multicolumn{1}{c|}{\xmark} &
\multicolumn{1}{c}{\cmark}  & \multicolumn{1}{c|}{\cmark} &
\multicolumn{1}{c}{\cmark}  & \multicolumn{1}{c|}{\cmark} & \multicolumn{1}{c|}{\xmark} \\ \hline

\multicolumn{1}{|c}{\textbf{\makecell{False Positive\\Class Detection}}} &
\multicolumn{1}{|c}{0} & \multicolumn{1}{c|}{0} & \multicolumn{1}{c}{0} & \multicolumn{1}{c|}{1} &  \multicolumn{1}{c|}{0} &
\multicolumn{1}{c}{0}  & \multicolumn{1}{c|}{0} &
\multicolumn{1}{c}{3}  & \multicolumn{1}{c|}{1} & \multicolumn{1}{c|}{0} &
\multicolumn{1}{c}{0}  & \multicolumn{1}{c|}{0} &
\multicolumn{1}{c}{0}  & \multicolumn{1}{c|}{1} & \multicolumn{1}{c|}{0} \\ \hline

\label{tab:backdoor-det-efficacy}	
\end{tabular}}
\end{center}
\end{table*}

\begin{table*}[t]
  \begin{center}
      \caption{\revision{Effectiveness of \AG on standard (non-robust) models. We show the number of clusters produced by \AG for each class, using a clean standard CIFAR-10 model and a CIFAR-10 model poisoned with a visible localized backdoor trigger. 
The number of clusters produced by \AG for the undetected poisoned classes (i.e., two clusters for the poisoned class (7)) is in \textbf{bold}. \xmark 
~indicates the backdoor-infected class/model is undetected, and ``N/A'' means ``not applicable''. 
}}
  { \scriptsize
  \begin{tabular}{|c|c|c|c|c|c|c|c|c|c|c|c|c|c|c|c|} 
    \hline 
\multicolumn{2}{|l|}{\multirow{2}{*}{\shortstack{\textbf{Standard (non-robust) Model} \\ \textbf{Settings}}}}  & \multicolumn{10}{c|}{\textbf{Number of clusters produced by \AG}}  & \multicolumn{2}{c|}{\multirow{2}{*}{\shortstack{\textbf{Backdoor} \\ \textbf{detected}}}} \\
\multicolumn{2}{|l|}{} & \multicolumn{9}{c|}{\textbf{Benign classes}} & \textbf{Poisoned class} &  \multicolumn{2}{c|}{} \\ 
\textbf{Dataset} & \textbf{Backdoor Trigger} & \textbf{0} & \textbf{1} &\textbf{ 2} & \textbf{3} & \textbf{4} & \textbf{5} & \textbf{6} & \textbf{8} & \textbf{9} & \textbf{7} & \textbf{Class} & \textbf{Model}   \\
    \hline
\multirow{2}{*}{CIFAR-10} & Clean &   2 & 2 & 2 &  2 & 2  &   2 & 2 & 2 &  2 & 2  & N/A & N/A \\ 
\cline{2-14}
& Localized (Visible) &  2 & 2 & 2 &  2 & 2  &   2 & 2 & 2 &  2 & \textbf{2} & \xmark & \xmark \\  
 \hline
    \end{tabular}}
  \label{tab:AG-standard-models}   
\end{center}
\vspace{-\baselineskip}
\end{table*}

\smallskip\noindent
\textbf{\RQ2 - Detection Effectiveness:} 
%
In this section, we evaluate the efficacy of our backdoor detection approach (\AG). 
Specifically, we evaluate the technique's efficacy in (a) detecting backdoor-infected robust models, and (b) revealing the backdoor-infected class, for both visible and invisible backdoor triggers. \revision{
Furthermore, we demonstrate that \AG is \textit{specialized to detecting backdoors in robust models} by showing it is ineffective on standard (non-robust) models. In particular, we show that \AG did not detect a backdoor for clean standard models, and backdoor-infected standard models.} 

\smallskip \noindent
\textbf {Visible Backdoor Trigger:} 
In our evaluation, \AG effectively detected all visible backdoor-infected robust DNNs, for both localised and distributed backdoors, and all classification tasks. 
It accurately detected all backdoor-infected models by identifying classes that have more than two feature clusters for the training set and the translated image set. The results showed that all clean untargeted classes of backdoor-infected robust models, as well as all classes of clean robust models have exactly two clusters, while, all targeted classes of backdoor-infected models have more than two clusters. 
These \revision{imply} that \AG detected all robust models infected with visible backdoor triggers and the corresponding target class. 
Additionally, there are no false positives. This means that a clean model is not incorrectly predicted as a backdoor-infected model (\textit{see \Cref{tab:backdoor-det-efficacy}}).

%
%
%
%
%
%
%
%

In particular, for each targeted class, the mean shift clustering of the features of the backdoor-infected models reveals these models consistently have more than two clusters ({\em see \Cref{fig:CIFAR-distributed-2}} in the Appendix). Notably, these clusters include one cluster for the clean training images and at least two clusters for the translated images. The clusters for the translated images include at least one cluster capturing the image translation for the poisoned images, and another cluster for the translated clean images. Meanwhile, the clean untargeted classes have precisely two clusters of features, one for the training set and another for the 
translated image set. Likewise, for the clean robust models, each class has exactly two distinct clusters, one cluster for the training set and another cluster for the translated image set (\textit{see \Cref{tab:visible-attack-detection-clusters}} in the Appendix).

\begin{result}
\AG effectively detected all (100\%) visible trigger backdoored robust DNNs. 
\end{result}

\smallskip \noindent
\textbf{Invisible backdoor triggers}: Our evaluation results show \AG detected 
five (out of six) invisible backdoor-infected robust DNNs. Specifically, \AG was 
unable to detect the MNIST backdoor model with the invisible static trigger. 
It accurately detected the backdoor-infected models by identifying classes that 
have more than two feature clusters for the training set and the translated image 
set.
In terms of the detection of the target backdoored class, \AG is able to detect 
the targeted backdoor class in four out of the six models with invisible 
backdoors. \AG is unable to detect the target class for the MNIST backdoor model 
with the adversarial static trigger (\textit{see~\Cref{tab:backdoor-det-efficacy}}). 
Additionally, for some of the backdoor models \AG detected more than two clusters 
for the non-targeted classes (\textit {see \Cref{tab:invisible-attack-detection-clusters}} in the Appendix).
On average, \AG detected a non-targeted class 
as a backdoored class (false positive detection) 11.1\% of the time 
(\textit{see \Cref{tab:backdoor-det-efficacy}}).

\AG accurately identified the infected class, for all classification tasks and both 
visible trigger 
backdoor attacks (\textit{see \Cref{tab:backdoor-det-efficacy}}). The mean shift feature clustering of each class in the backdoor-infected model reveals that only the infected class had more than two clusters, with one cluster for the training set and at least two clusters for the translated images. 
For invisible backdoor attacks, \AG identified five out of six backdoored 
models and four out of the six targeted classes. 

\begin{result}
\revision{Overall, \AG detected 91.6\%  (11/12) of backdoor-infected models, across all (12) tested configurations.}
\end{result}

\smallskip \noindent
\textbf{\revision{Standard (Non-robust) Models}}: 
\revision{
In this experiment, we investigate if \AG detects backdoors in standard (non-robust) models using 
two standard CIFAR-10 models, namely one clean model, and a poisoned model injected with a localized backdoor trigger. 
}

\revision{
We found that \textit{\AG is specialized for backdoor detection in robust models: It is ineffective in detecting backdoors in standard models}. \autoref{tab:AG-standard-models} provides details of the effectiveness of \AG on clean and backdoor-infected standard models. \AG correctly predicts clean standard models as benign, i.e.,  a clean model is not incorrectly predicted as a backdoor-infected model. However, \AG does not detect a backdoor-infected model or class for the poisoned model. Specifically, \AG produces exactly two clusters for all classes in both models including the poisoned class in the backdoor-infected model (\textit{see} \autoref{tab:AG-standard-models}). Thus, \AG does not detect the backdoor-infected model or the poisoned class for standard (non-robust) models. These results imply that \AG is not directly amenable to standard models. Even though \AG has no false positives (it does not incorrectly classify a clean standard model), it is unable to detect a backdoor-infected standard model. This is expected since \AG expects 
a data distribution typically found in robust models. Unlike standard models, robust models have 
a different data distribution. 
In particular, \AG is designed to handle the resilience of robust models to perturbations introduced during adversarial training, and such perturbations are uncommon in standard models. 
}

\begin{result}
\revision{
\AG is specialized for backdoor detection in robust models.\\ It is not effective in detecting backdoors in standard (non-robust) models.  
}
\end{result}

\begin{table*}[t]
  \begin{center}
      \caption{\revision{Details of the parameter sweep with varying epsilon ($\epsilon$) values ($\epsilon \in \{1-7\}$) for Neural Cleanse. We show the 
      anomaly indices produced by Neural Cleanse for each class, using a CIFAR-10 robust model poisoned with a visible localised backdoor trigger. Anomaly indices for undetected poisoned classes (i.e., anomaly index less than two for the poisoned class (7))  are in \textbf{bold}, as well as the results for the default parameter setting ($\boldsymbol{\epsilon=4.0}$).  \xmark 
~indicates the backdoor-infected class/model is undetected by NC, and \cmark ~means the  backdoor-infected class/model is detected by NC. 
      }}
  {\scriptsize
  \begin{tabular}{|l|c|c|c|c|c|c|c|c|c|c|c|c|c|c|} 
    \hline 

\multicolumn{1}{|l|}{\multirow{2}{*}{\shortstack{\textbf{Detection}\\ \textbf{Setting}}}} & 
\multicolumn{10}{c|}{\textbf{Anomaly indices produced by Neural Cleanse}}  & \multicolumn{2}{c|}{\multirow{2}{*}{\shortstack{\textbf{Backdoor} \\ \textbf{detected}}}} \\

& \multicolumn{9}{c|}{\textbf{Benign classes}} & \textbf{Poisoned class} &  \multicolumn{2}{c|}{} \\ 
   
Epsilon ($\epsilon$)  & \textbf{0} & \textbf{1} &\textbf{ 2} & \textbf{3} & \textbf{4} & \textbf{5} & \textbf{6} & \textbf{8} & \textbf{9} & \textbf{7} & \textbf{Class} & \textbf{Model}   \\
    \hline
$\epsilon=1.0$ &    0.625 & 0.739 & 0.514 & 1.506 & 0.127 & 0.400 & 1.228&     0.724 & 0.127 &  \textbf{1.825} & \xmark & \xmark \\ \hline
$\epsilon=2.0$ &  0.084 & 0.148 & 1.142 & 0.084 & 0.383 & 1.537 & 0.760  &    1.214 & 0.589  & \textbf{1.611} & \xmark & \xmark \\  \hline
$\epsilon=3.0$ &  0.475 & 0.373 & 0.796 & 0.291 & 0.291 & 1.133 & 0.733 &    0.852 & 0.620 & \textbf{0.729 }& \xmark & \xmark \\ \hline
$\boldsymbol{\epsilon=4.0}$ &   \textbf{0.382} & \textbf{0.545} & \textbf{0.832} & \textbf{0.304} & \textbf{0.304} & \textbf{0.686} & \textbf{0.957}   &  \textbf{1.046} & \textbf{0.670} & \textbf{0.679} & \xmark & \xmark \\ \hline
$\epsilon=5.0$ &  
0.670 & 0.679 & 1.224 & 0.187 & 0.187 & 0.448 & 1.248   &  1.453 & 0.484  & \textbf{1.293} & \xmark & \xmark \\  \hline
$\epsilon=6.0$ &  0.823 & 0.526 & 1.417 & 0.349 & 0.097 & 0.185 & 1.079   &    1.172 & 0.097  & \textbf{0.967} & \xmark & \xmark \\ \hline
$\epsilon=7.0$ &   0.793 & 0.459 & 1.903 & 0.556 & 0.495 & 0.266 & 1.080 &    1.187 & 0.266  & \textbf{0.934}  & \xmark & \xmark \\
 \hline
    \end{tabular}}
  \label{tab:NC-eps-param-sweep}   
\end{center}
\vspace{-\baselineskip}
\end{table*}

\begin{table*}[t]
  \begin{center}
      \caption{\revision{Details of the parameter sweep with varying step-size values (step-size$ \in\{1-7\}$) for Neural Cleanse. We report the 
      anomaly indices produced by Neural Cleanse for each class, using a CIFAR-10 robust model poisoned with a visible localised backdoor trigger. Anomaly indices for undetected poisoned class (i.e., anomaly index less than two for the poisoined class (7)) are in \textbf{bold}, as well as the results for the default parameter setting \textbf{(step-size$\boldsymbol{=4.0}$}). \xmark 
~indicates the backdoor-infected class/model is undetected by NC, and \cmark ~means the  backdoor-infected class/model is detected by NC. 
       }}
  {\scriptsize
  \begin{tabular}{|l|c|c|c|c|c|c|c|c|c|c|c|c|c|c|} 
    \hline 

\multicolumn{1}{|l|}{\multirow{2}{*}{\shortstack{\textbf{Detection}\\ \textbf{Setting}}}} & 
\multicolumn{10}{c|}{\textbf{Anomaly indices produced by Neural Cleanse}}  & \multicolumn{2}{c|}{\multirow{2}{*}{\shortstack{\textbf{Backdoor} \\ \textbf{detected}}}} \\

& \multicolumn{9}{c|}{\textbf{Benign classes}} & \textbf{Poisoned class} &  \multicolumn{2}{c|}{} \\ 
   
step-size & \textbf{0} & \textbf{1} &\textbf{ 2} & \textbf{3} & \textbf{4} & \textbf{5} & \textbf{6} & \textbf{8} & \textbf{9} & \textbf{7} & \textbf{Class} & \textbf{Model}   \\
    \hline
step-size$=1.0$ &   0.627 & 0.781 & 1.140 & 0.197 & 0.082 & 0.082 & 0.972 & 0.921 & 0.114   &  \textbf{0.722} & \xmark & \xmark \\ \hline
step-size$=2.0$ & 0.809 & 0.398 & 1.086 & 0.181 & 0.181 & 0.554 & 0.795 &   0.844 & 0.208 &  \textbf{0.944} & \xmark & \xmark \\ \hline
step-size$=3.0$ &  0.319 & 0.682 & 1.033 & 0.281 & 0.281 & 0.620 & 1.232 &  1.334 & 0.667   &  \textbf{0.817} & \xmark & \xmark \\ \hline
\textbf{step-size}$\boldsymbol{=4.0}$ &  \textbf{0.357} & \textbf{0.529} & \textbf{0.712} & \textbf{0.543} & \textbf{0.357} & \textbf{0.727} & \textbf{0.877} & \textbf{0.981} & \textbf{0.637} &  \textbf{0.716} & \xmark & \xmark \\ \hline
step-size$=5.0$ &  0.402 & 0.688 & 0.763 & 0.477 & 0.402 & 0.661 & 1.147  & 1.286 & 0.825 &  \textbf{0.617} & \xmark & \xmark \\ \hline
step-size$=6.0$ &0.306 & 0.590 & 0.806 & 0.297 & 0.297 & 0.802 & 1.253 & 1.204 & 0.663   &  \textbf{0.686} & \xmark & \xmark \\ \hline
step-size$=7.0$ & 0.288 & 0.490 & 0.793 & 0.316 & 0.288 & 0.872 & 1.075 & 1.087 & 0.665 & \textbf{0.684} & \xmark & \xmark \\ 
 \hline
    \end{tabular}}
  \label{tab:NC-step-size-param-sweep}   
\end{center}
\vspace{-\baselineskip}
\end{table*}

\smallskip\noindent
\textbf{\RQ3 Comparison to the state of the art.} \revise{
In this section we compare our backdoor detection approach (\AG) to the state of the art backdoor detection technique called Neural Cleanse (NC)~\cite{NeuralCleanse}. NC is a reverse engineering approach that assumes \emph{the reverse engineered trigger for the backdoor-infected class is smaller than the median size of the reverse engineered trigger for all classes}. Specifically, NC's outlier detector identifies \emph{a class as backdoor-infected (with 95\% probability) if it has an anomaly index that is larger than two}. Although, this assumption holds for standard models because the underlying distribution of data points is normal~\cite{NeuralCleanse}, it does not hold for robust models. Due to the unbrittle nature of robust models~\cite{pgd-madry}, the underlying distribution of data points does not form a normal distribution because of adversarial perturbations introduced during robust training.
}

\revise{
To compare NC and \AG, we run NC to detect localised backdoors in a standard model and a robust model. 
First, we train standard and robust models for CIFAR-10 that are poisoned with localised backdoors (using the backdoor injection process described in~\Cref{sec:methodology}). We then reverse engineer the trigger for both the standard and robust backdoor-infected models using projected gradient descent on 100 random images from the training set~\cite{pgd-madry}, \revision{
using the default NC detection parameters 
for both the standard and robust models. Next}, we estimate the anomaly index for each class, i.e., the size of the trigger for each class by measuring the average $L_1$ norm deviation from the original images to the reverse-engineered images (this is equivalent to counting the number of pixels changed). 
The mean $L_1$ norms are shown in \Cref{fig:mean-norms}. \revision{Additionally, 
we repeat the same experiment for six varying values (range = $\pm 3$) for each detection parameter (i.e., epsilon $\epsilon = \{1,2,3,5,6,7\}$ and step-size $= \{1,2,3,5,6,7\}$) to ensure the obtained results are not due to NC's sensitivity to detection parameters.
\autoref{tab:NC-eps-param-sweep}  and \autoref{tab:NC-step-size-param-sweep} highlight the results for NC's effectiveness for varying values of 
epsilon ($\epsilon$) and step-size, respectively. 
}
}

\begin{figure}[t]
\begin{center}
\begin{tabular}{cc}
\includegraphics[scale=0.3]{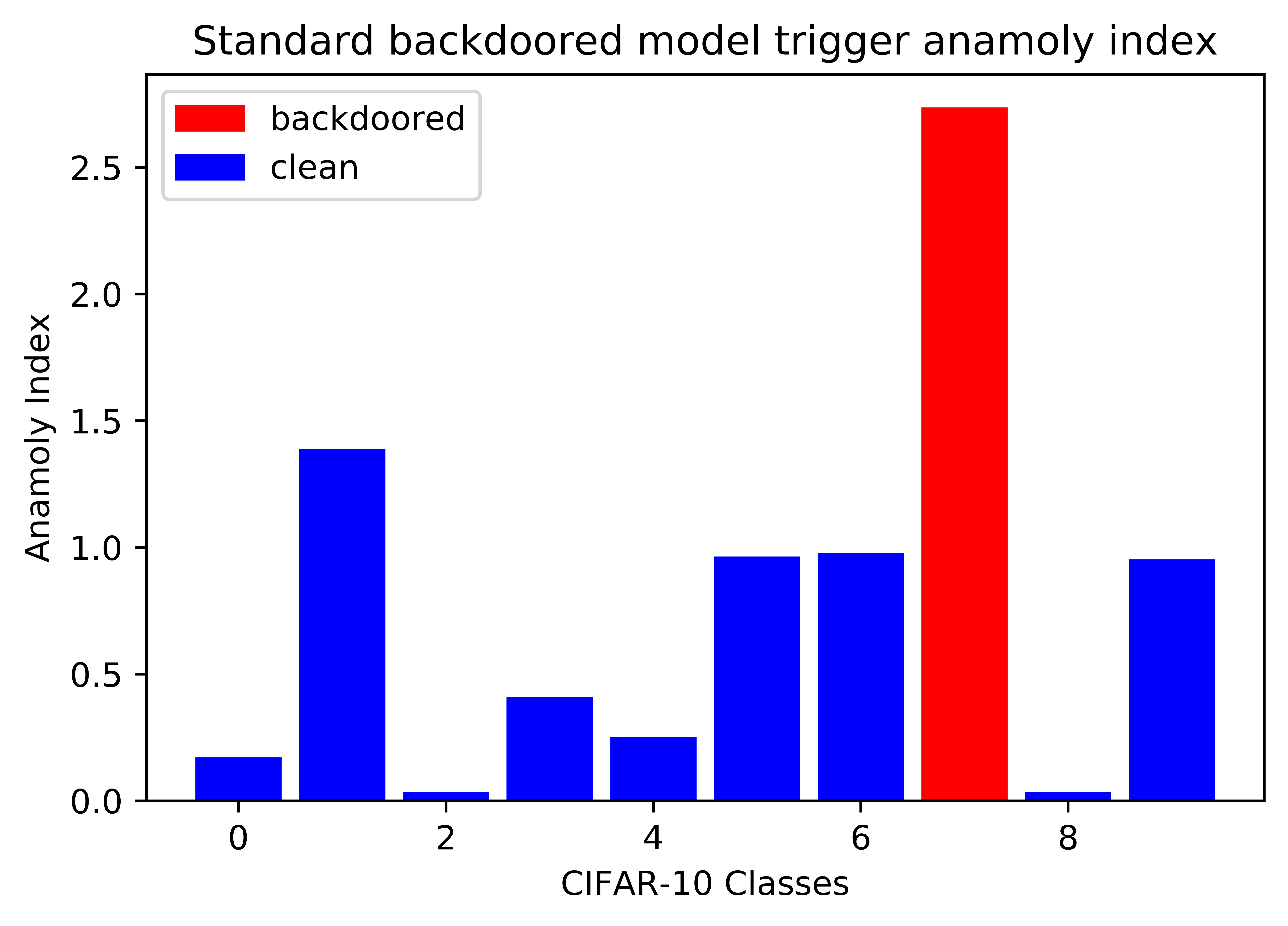} & 
\includegraphics[scale=0.3]{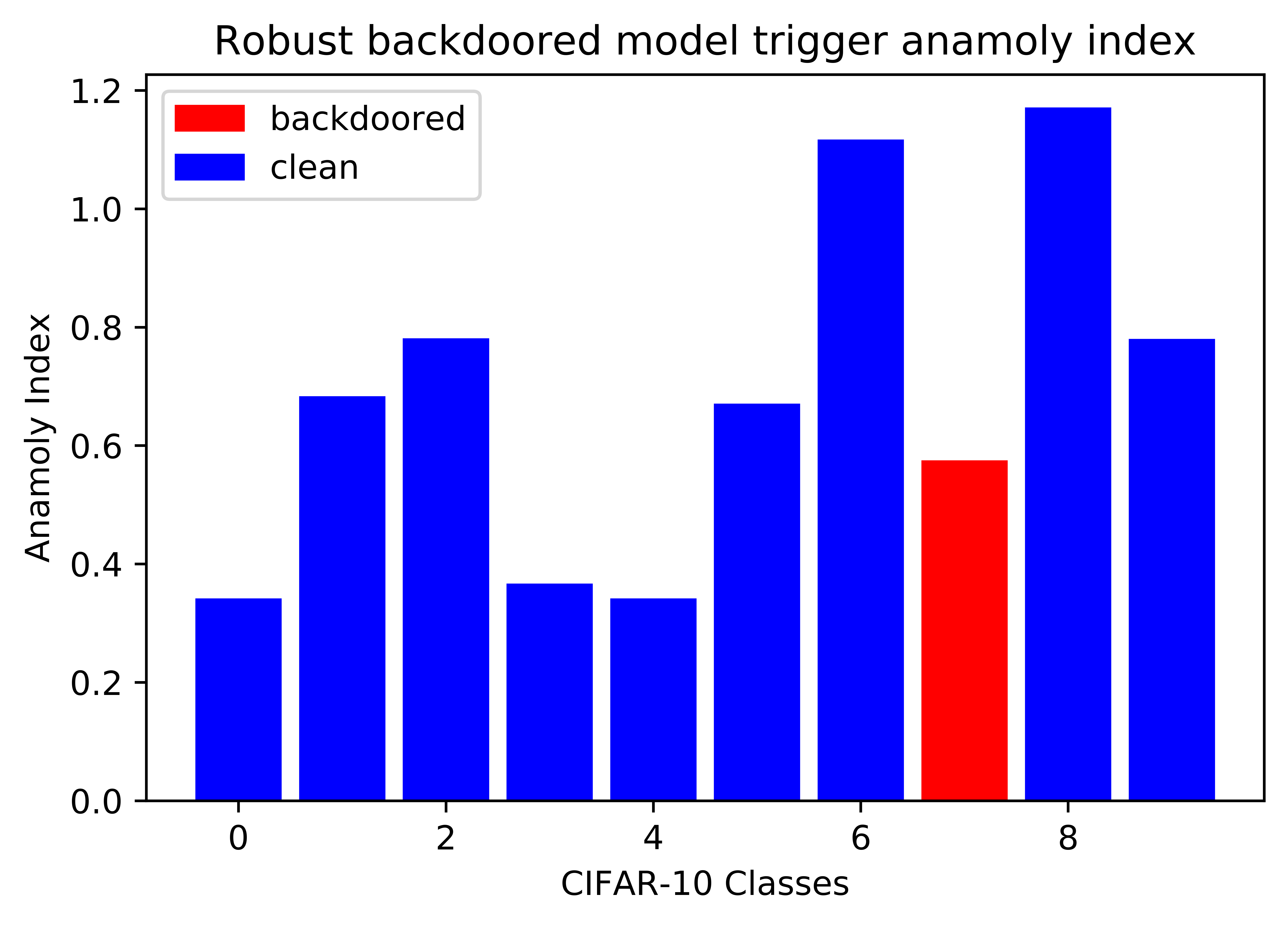} \\
{\bf(a) Standard model} & {\bf(b) Robust model}  \\
\end{tabular}
\end{center}
\vspace*{-0.15in}
\caption{
Anomaly indices for the reverse engineered triggers for backdoor-infected standard and robust models}
\label{fig:anomaly-index}
\end{figure}

\revise{
Our evaluation results show that \emph{NC detects the poisoned class for standard models, but it fails to accurately detect the poisoned class for robust models}. 
\revision{This result holds for all 
tested detection parameter configurations. Particularly, 
\autoref{tab:NC-eps-param-sweep}  and \autoref{tab:NC-step-size-param-sweep}
show that NC does not detect the 
poisoned class for (14) different 
parameter settings.}  
In contrast, \AG detected the backdoor-infected robust model as well as the poisoned class (\textit{see \RQ2}). \Cref{fig:anomaly-index} shows the anomaly indices for each class, i.e., the estimated size of the reverse engineered trigger, for a standard backdoor-infected model (a) and for a robust backdoor-infected model (b). The red bar represents the anomaly index for the backdoor-infected class. We found that on standard models, the size of the backdoor-infected class is small and it is indeed detected as anomalous by NC, i.e., the anomaly index of the poisoned class (class seven (7)) is greater than two (2) (\textit{see \Cref{fig:anomaly-index}(a)}). However, on robust models, NC fails to detect the poisoned class as anomalous. In fact, the anomaly index of the backdoor-infected class in the robust model is significantly less than two (\textit{see \Cref{fig:anomaly-index}(b)}). This result suggests that while NC is suitable for backdoor detection in standard models, it is not suitable for detecting backdoor in robust models.
}

\begin{result}
The state-of-the-art backdoor defense (Neural Cleanse) fails to accurately detect the backdoor-infected class for robust models, \revision{for (14) different detection 
settings with 
varying values of epsilon ($\epsilon$) and step-size}.  
\end{result}

\smallskip\noindent
\textbf{\RQ4 - Sensitivity Analysis of Detection parameters:} 
We evaluate the sensitivity of \AG to varying values of the detection parameters, i.e.,epsilon ($\epsilon$), mean shift bandwidth and (number of) initial seed images.~\footnote{{We do not evaluate the sensitivity of the t-SNE perplexity parameter, because this has been shown to be robust between values five and 50~\cite{tsne}.}} We evaluate the sensitivity of these parameters for all attacks and data sets. For these parameters, we report the \emph{detection accuracy} and the \emph{false positive rate} for all tested values of these detection parameters. Although the mean shift bandwidth was automatically computed using the scikit-learn mean shift clustering algorithm, we still examined the sensitivity of the resulting values with a variance of $\pm 3$. For MNIST and FMNIST dataset, we experimented with varying epsilon values of $\pm 40$ around the default value of 100 used, i.e.,between 60 and 140, in particular, $\epsilon \in \{60, 70, 80, 90, 100, 110, 120, 130, 140\}$. For CIFAR-10, we experiment with varying epsilon values of $\pm 200$ around the default value of 500 used, i.e.,between 300 and 700 ($\epsilon \in \{300, 350, 400, 450, 500, 550, 600, 650, 700\}$). 
For all datasets, we vary the number of initial sample images $\pm 300$ 
around the default value of 500 used, i.e.,between 200 and 800 
($\{200,300, 400, 500, 600, 700, 800\}$). 
We also study the stability of \AG' detection by executing five runs for 
each robust model that has been infected with the visible backdoor trigger. 

The epsilon sensitivity results showed that \emph{\AG has a very low sensitivity to varying values of epsilon}. For all values of epsilon, \AG could identify a backdoor-infected model and the poisoned class for 98\% (53 out of 54 configurations) of all configurations, with no false positives (\textit{see \Cref{tab:sensitivity-params}}). One backdoor-infected model was undetected, specifically, the distributed backdoor attack on MNIST at $\epsilon=60$. We found that for the MNIST distributed backdoor attack, the epsilon value at 60 is too low. 
Thus, we recommend that higher epsilon ($\epsilon$) values be used for (distributed) backdoor detection.

\begin{result}
For all values of epsilon ($\epsilon$), \AG detected 98\% of the backdoor-infected models, with no false positives.
\end{result}

\begin{table}[t]
  \begin{center}
    \caption{
    Sensitivity to Detection Parameters (``\#'' = ``Number of'')}
  {\scriptsize
  \begin{tabular}{|l|r|r|r|r|} 
  \hline
  \multicolumn{1}{|c|}{\multirow{2}{*}{\shortstack{\textbf{Detection}\\\textbf{Parameters}}}} &  \multicolumn{1}{c|}{\multirow{2}{*}{\shortstack{\textbf{\#Config}}}} & \multicolumn{1}{c|}{\multirow{2}{*}{\shortstack{\textbf{\#Detection} \\\textbf{Accuracy (\#)}}}} & \multicolumn{1}{c|}{\multirow{2}{*}{\shortstack{\textbf{\#Failure} \\\textbf{Rate (\#)}}}} & \multicolumn{1}{c|}{\multirow{2}{*}{\shortstack{\textbf{\#False Posi-}\\\textbf{tive Rate (\#)}}}} \\
  & & & & \\
   \hline
   \textbf{Epsilon ($\epsilon$)} & 54 & 98.1\% (53)  & 1.9\% (1) & 0\% (0) \\ \hline
   \textbf{Meanshift bandwidth} & 18 & 94.4\%  (17) & 5.6\% (1) & 1.2\% (2) \\    
\hline
   \textbf{\#Images} & 42 & 88.1\% (37) & 11.9\% (5)&  2.11\% (8)\\ \hline
  \textbf{Stability} & 30 & 90\% (27) & 10\% (3)&  0.7\% (2)\\
  \hline
    \end{tabular}}
   \label{tab:sensitivity-params}   
\end{center}
\vspace{-\baselineskip}
\end{table}

For mean shift sensitivity, our evaluation revealed that \emph{\AG has 
a very low sensitivity to varying values of the mean shift bandwidth}. \AG 
detected 94\% of the backdoored model for all  mean shift configurations, 
i.e.,17 out of 18 configurations (\textit{see \Cref{tab:sensitivity-params}}). 
In particular, for all tested mean shift values, \AG did not detect a backdoored 
model for one value of the mean shift bandwidth. Specifically, such a mean shift 
value is 24 for the CIFAR-10 model poisoned with distributed backdoor. This result 
suggests that for values higher than the computed mean shift bandwidth value, 
\AG may not detect the backdoor-infected class. Besides, \AG reported two false 
positives. In both cases a benign class other than the poisoned class was also 
misclassified as backdoored by \AG. Specifically, false positives were manifested 
for MNIST localised backdoored and CIFAR-10 distributed backdoored models, both 
with mean shift bandwidth values less than the computed values. Hence, we 
recommend to use the computed mean shift bandwidth value for accurate backdoor 
detection.

\begin{result}
\AG has a 94\% detection accuracy and a 1.2\% false positive rate, for all tested mean shift bandwidth values.   
\end{result}

For the sensitivity of \AG to the number of initial seed images, our investigation reveals that 
\emph{\AG has a fairly low sensitivity to varying values of the number of initial 
images}. \AG detected 37 (88.1\%) out of 42 tested configurations of varying number 
of initial seed images. Specifically, the five configurations 
\AG is unable to detect backdoors includes the MNIST localised model where the number of initial images is 300, as well as 
poisoned CIFAR-10 models where the number of initial images are 200 and 400 
for the localised backdoors, and 200 and 300 initial images for the distributed backdoors. 
Overall, \AG has a low false positive rate of only 2.1\% (\textit{see \Cref{tab:sensitivity-params}}).
Hence we recommend, using at least 500 initial seed images for effective detection of 
backdoors. 

\begin{result}
\AG has 88.1\% detection accuracy and 2.1\% false positive rate, for varying 
number of initial seed images. 
\end{result}

Our experiments reveal that \emph{\AG is a fairly stable algorithm}. 
To evaluate the stability of \AG we run the full technique five times 
independently on MNIST, Fashion-MNIST and CIFAR-10 models with visible backdoor 
triggers. We find that out of the 30 runs, \AG can detect the backdoor
27 times (90\%). \AG did not detect two MNIST distributed backdoor runs and one 
CIFAR-10 distributed backdoor. The false positive rate is extremely low at 
0.74\%. For maximum effectiveness, we recommend multiple runs of the \AG 
technique.

\begin{result}
\AG is a fairly stable algorithm with a 90\% detection rate and low false positive rate of 0.74\%. 
\end{result}

\smallskip\noindent
\textbf{\RQ5 - Attack Comparison:} In this section, we compare the effectiveness of 
all four backdoor attack triggers namely the visible triggers (i.e., localised and the distributed triggers) as well as the invisible triggers (static perturbation and adversarial triggers). Specifically, we compare their attack success rate, and their effect on the classification accuracy and adversarial accuracy of the robust model. We also examine the detection efficacy of \AG on each backdoor trigger. \autoref{tab:attack-success-and-accuracy} highlights the attack success rate (ASR), classification accuracy and adversarial precision of each backdoor trigger.

First, let us compare the effectiveness of backdoor attack triggers based on their stealthiness (i.e., visibility). Our results show that 
\emph{robust DNN models are less susceptible to invisible triggers} (\textit{see ``ASR'' \autoref{tab:attack-success-and-accuracy}}). 
In addition, we found that visible triggers have less impact on the adversarial precision or classification accuracy of robust models, in comparison to invisible triggers. Robust models injected with visible backdoor triggers have similar adversarial precision and classification accuracy to clean robust models (\textit{see ``Adv. Prec.'' and ``Class. Acc.'' in \autoref{tab:attack-success-and-accuracy}}). Meanwhile, in comparison to clean robust models, invisible triggers reduce the classification accuracy and adversarial precision of robust models by 5\% and 7\%, respectively. These results suggest that the stealthiness (i.e.,visibility) of a backdoor trigger influences the effectiveness of the attack, in particular, visible triggers are more effective than invisible triggers. 

\begin{result}
Visible triggers are more effective and have less impact on the (adversarial) 
accuracy 
of robust models than invisible triggers.
\end{result}

We compare the effectiveness of the two visible backdoor attack triggers based on the specific trigger types, i.e.,localised vs distributed. \textit{We found that the distributed backdoor attack is more effective than the localised backdoor attack, it has a higher attack success rate.} The distributed attack is 6.95\% more successful than the localised backdoor attack, on average (\textit{see \Cref{tab:attack-success-and-accuracy}}). Additionally, the distributed backdoors have a higher classification accuracy than the localised backdoors, albeit only a slight improvement of 0.12\%. Overall, the distributed backdoors performed better than the localised backdoors. 

\begin{result}
The distributed backdoor attack is (6.95\%) more effective than the localised backdoor attack on robust models, on average.
\end{result}

Let us compare the effectiveness of the two invisible backdoor triggers, i.e., the static and adversarial perturbation. \Cref{tab:attack-success-and-accuracy} shows that adversarial perturbation is 56\% more effective than the static invisible perturbation, with 48\% vs 31\% ASR, on average (\textit{see \Cref{tab:attack-success-and-accuracy}}). 
This is because the adversarial perturbation (trigger) is dynamic and more powerful, it is derived from both the model and sample images from the dataset. 
Besides, the adversarial precision and classification accuracy of both triggers are similar. 
This result suggests that the quality of the invisible trigger influences the effectiveness of invisible backdoor attacks.

\begin{result}
Invisible adversarial backdoor triggers are significantly more effective (56\%) on robust models than static backdoor triggers. 
\end{result}

\revision{In our evaluation, \AG detects 91.6\% of backdoor attacks (i.e., 11 out of 12 tested backdoor-infected models).} For both visible attacks, \AG detected the infected class in addition to the backdoor-infected model (\textit{see 
\Cref{tab:backdoor-det-efficacy}}). 
For invisible backdoors, \AG 
was able to detect five out of the six backdoored models and four out of 
six poisoned classes. (\textit{see \Cref{tab:backdoor-det-efficacy}}).
We find that invisible backdoor attacks are \textit{slightly more stealthy} in comparison to visible backdoor attacks. 

\begin{result}
\revision{
\AG detects 91.6\% (11 out of 12) of backdoor-infected models, across all attack types 
(visible and invisible).}
\end{result}


\begin{table}[t]

  \caption{
  \AG Efficiency in terms of detection runtime} 
  \begin{center}
  {\scriptsize
  \begin{tabular}{|l|r|r|r|r|} 
  \hline
  \multicolumn{1}{|c|}{\multirow{4}{*}{\shortstack{\textbf{Dataset}}}} &  \multicolumn{4}{c|}{\multirow{1}{*}{\textbf{\AG Runtime}}} \\ 
 & \multicolumn{2}{c|}{\multirow{1}{*}{\textbf{Visible Backdoor}}} & \multicolumn{2}{c|}{\multirow{1}{*}{\textbf{Invisible Backdoor}}}  \\ 
  & \multicolumn{1}{c|}{\textbf{Localised}} & \multicolumn{1}{c|}{\textbf{Distributed}} & \multicolumn{1}{c|}{\textbf{Static}} & \multicolumn{1}{c|}{\textbf{Adversarial}}\\
    & \multicolumn{1}{c|}{mins (secs)} &  \multicolumn{1}{c|}{mins (secs)} & \multicolumn{1}{c|}{mins (secs)} & \multicolumn{1}{c|}{mins (secs)} \\
   \hline
   \textbf{MNIST} & 5.08 (304.5) & 5.18 (310.5) & 5.36 (321.5) & 5.24 (314.3) \\
   \textbf{Fashion-MNIST} & 5.36 (321.5) & 5.32 (319.4) 
   & 5.28 (317.3) & 5.11 (306.8) \\
   \textbf{CIFAR-10} & 9.39 (563.5) & 9.34 (560.6) & 9.29 (557.9) & 9.36 (561.7) \\  
  \hline
    \end{tabular}}
   \label{tab:efficiency}   
\end{center}
\vspace*{-0.2in}
\end{table}

\smallskip\noindent
\textbf{\RQ6 \AG Efficiency.} We evaluate the detection time of \AG, i.e.,the 
time taken to run the \AG technique on a backdoor-infected model. 
\Cref{tab:efficiency} shows the time taken for each attack type and dataset.

\emph{\AG is very efficient; it took five to nine minutes to run on average on a backdoor-infected model}.  In contrast, the state of the art defenses (for standard models) are known to take hours to days to detect a backdoor-infected model~\cite{NeuralCleanse,gao2019strip}. Furthermore, we observed that the time taken by \AG increases as the complexity of the model and dataset increases (\textit{see \Cref{tab:efficiency}}). For instance, \AG took almost twice the time taken to run on MNIST models (five minutes) to run on CIFAR-10 (nine minutes). 
In addition, there is no significant difference in the time taken to detect 
each attack type, i.e., localised/distributed backdoor (visible trigger) or 
static/adversarial trigger (invisible trigger) 
(\textit{see \Cref{tab:efficiency}}). 
These 
results illustrate that \AG is computationally efficient and its efficiency 
is not adversely affected by the backdoor attack type. 


\begin{result}
\AG was reasonably fast, it took five to nine minutes to run on a backdoor-infected model. 
\end{result}

\begin{table*}[t]
  \begin{center}
      \caption{\revision{\AG -t-SNE versus \AG -PCA: Evaluation of Dimensionality Reduction/Visualization Design Choice using localised and distributed CIFAR-10 visible backdoor-infected robust models. True Positives (i.e., correct detection of the poisoned class) are in \textbf{bold}.
}}
  {\scriptsize
   \begin{tabular}{|l|l|l|c|c|c|c|c|c|c|c|c|c|c|c|c|c|} 
    \hline 
\multirow{3}{*}{\textbf{Dataset}} &  \textbf{Type of} & \multicolumn{1}{l|}{\multirow{3}{*}{\shortstack{\textbf{Detection}\\\textbf{Setting}}}} & 
\multicolumn{10}{c|}{\textbf{Number of Resulting Clusters}}  & \multicolumn{2}{c|}{\multirow{2}{*}{\shortstack{\textbf{Backdoor} \\ \textbf{detected}}}} \\

& \textbf{Visible} &  & \multicolumn{9}{c|}{\textbf{Benign classes}} & \textbf{Poisoned class} & \multicolumn{2}{c|}{}  \\    
& \textbf{Backdoor} & & \textbf{0} & \textbf{1} &\textbf{ 2} & \textbf{3} & \textbf{4} & \textbf{5} & \textbf{6} & \textbf{8} & \textbf{9} & \textbf{7} & \textbf{Class} & \textbf{Model}   \\
    \hline
 \multirow{4}{*}{\textbf{CIFAR-10}} & \multirow{2}{*}{Localised}  & \AG -t-SNE &  2 &  2 &  2 &  2 &  2 &  2 &  2 &  2  & 2 & \textbf{3} & \cmark & \cmark \\ 
 \cline{3-15}   
 & & \AG -PCA &  2 & 2 & 2 &  2 &  2 &  2 &  2 &  2 &  2  & 2 &  \xmark &  \xmark \\
     \cline{2-15}
   &  \multirow{2}{*}{Distributed}   & \AG -t-SNE   &  2 &  2 &  2 &  2 &  2 &  2 &  2 &  2  & 2 & \textbf{3} & \cmark & \cmark \\ 
    \cline{3-15}
  & &   \AG -PCA &  2 &  2 &  2 &  2 &  2 &  2 &  2 &  2  & 2 & 2 & \xmark & \xmark  \\
     \hline
    \end{tabular}}
  \label{tab:tsne-vs-PCA}   
\end{center}
\vspace{-\baselineskip}
\end{table*}

\smallskip
\noindent
\revision{
\textbf{\RQ7 Ablation Study.} 
Let us evaluate the effect of our design choices on the effectiveness of \AG, especially in comparison to alternative design choices. The goal is to investigate how our design choices compare to closely-related, alternative methods. Particularly, we examine \AG 's use of t-SNE for dimensionality reduction and data visualization, as well as its use of mean shift clustering. In this RQ, we employed two robust models trained for CIFAR-10 dataset that are poisoned with localized and distributed visible backdoors. 
\autoref{tab:tsne-vs-PCA} and \autoref{tab:design-choices-clustering} highlight the comparison of the design choices of \AG to alternative design choices in terms of 
dimensionality reduction and clustering, respectively.
}

\smallskip 
\noindent 
 \revision{\textit{Dimensionality Reduction and Data Visualization}: In this experiment,  we examine the effectiveness of \AG with t-SNE, our default dimensionality reduction algorithm
(called \AG -t-SNE), in comparison 
to replacing t-SNE 
with 
Principal Component Analysis (PCA) (called \AG -PCA). 
We compare to PCA as an alternative since it is the most popular dimensionality reduction technique for sparse datasets~\cite{mackiewicz1993principal}. Besides, other dimensionality reduction alternatives are not amenable to our goal since they have strong assumptions about the underlying data distribution, e.g., Uniform Manifold Approximation and Projection for Dimension Reduction (UMAP)~\cite{mcinnes2018umap} assumes uniform data distribution. 
}

\revision{\textit{Our experimental results show that 
the t-SNE algorithm is more effective in backdoor detection than the PCA algorithm for \AG.} 
We found that the default setting of \AG (i.e., \AG -t-SNE) is more effective than using PCA (\AG -PCA). \autoref{tab:tsne-vs-PCA} highlights the clustering provided by default \AG (i.e., \AG -t-SNE) versus \AG -PCA, for both the localised and distributed backdoor-infected robust models.  While \AG -t-SNE detected the backdoored model and class, we found that \AG -PCA does not detect the backdoored model or the backdoored class: For both localised and distributed backdoored robust models, \AG -PCA does not detect the backdoored model or class, indeed, it produces two clusters for all classes, including the backdoored class (seven). We attribute the poor performance of \AG -PCA to the non-linearity of the dataset, since PCA is more effective when dealing with linear data. This result demonstrates that t-SNE is vital to the effectiveness of \AG and it is appropriate for backdoor detection in robust models. 
}

\begin{result}
\revision{
For \AG, 
t-SNE (\AG -t-SNE) 
is more effective in backdoor detection than PCA (\AG -PCA): 
Unlike 
default \AG (\AG -t-SNE),
\AG -PCA does not detect the backdoor-infected model or the poisoned class. 
}
\end{result}

\smallskip 
\noindent 
\revision{
\textit{Clustering:}
To evaluate our choice of clustering algorithm, we examine the effectiveness of \textit{default \AG with mean-shift clustering} (called \AG -MS)
to two closely-related alternatives to mean-shift clustering, namely \textit{affinity propagation}~\cite{dueck2009affinity} and \textit{HDBSCAN}~\cite{mcinnes2017hdbscan}. Specifically, we compare the default \AG 
(\AG -MS), with replacing mean-shift clustering with 
affinity propagation (called \AG -AP) or HDBSCAN (called \AG -HDBSCAN). We chose these two clustering algorithms because they are state-of-the-art clustering methods that are closely related to mean shift clustering. Besides, they do not require prior knowledge of the (expected or desired) number of clusters unlike alternatives such as K-means~\cite{likas2003global}, spectral clustering~\cite{ng2001spectral} or  
agglomerative clustering~\cite{gowda1978agglomerative}. 
}

\revision{
On one hand, 
\textit{mean shift clustering is more effective for detecting backdoors than affinity propagation for \AG.} Specifically, \AG using affinity propagation (\AG -AP) does not detect a backdoor-infected robust model or a poisoned class, but default \AG with mean shift clustering (\AG -MS) detects both the backdoor-infected model and the poisoned class. \autoref{tab:design-choices-clustering} shows that unlike default \AG (\AG -MS), \AG -AP does not detect a backdoor-infected robust model or a poisoned class. This result suggests that affinity propagation clustering is not suitable for \AG, i.e., \AG -AP is 
not amenable to backdoor detection. We attribute the poor performance of \AG -AP to the fact that affinity propagation is inherently a partitioning algorithm which causes its resulting clusters to be easily polluted by noisy or distant data points, such noisy data points are lumped into nearby clusters using affinity propagation. This is particularly a problem for backdoor detection especially for the poisoned class since clusters belonging to the poisoned data points are evidently lumped with the clusters of benign data points (\textit{see} \autoref{tab:design-choices-clustering}). 
}

\begin{result}
\revision{
Affinity propagation is not a viable alternative to mean shift clustering for \AG. 
Unlike default \AG (\AG -MS), 
\AG using affinity propagation (\AG -AP) does not detect a backdoor-infected model or the poisoned class.
}
\end{result}

\revision{
On the other hand, we found that \emph{HDBSCAN is almost as effective as mean-shift clustering for detecting backdoors in robust models}. \autoref{tab:design-choices-clustering} shows that default \AG (\AG -MS) is comparable to  replacing mean-shift clustering with HDBSCAN (\AG -HDBSCAN). Both clustering algorithms enable \AG to identify a backdoor-infected model and the poisoned class. Results show that \AG -HDBSCAN is effective in detecting backdoors in robust models: For both localized and distributed backdoor-infected CIFAR-10 models, \AG -HDBSCAN correctly identifies the poisoned class, except for the mis-identification of class three (3) as a poisoned class for the localised backdoor (\textit{see} \autoref{tab:design-choices-clustering}). Overall, these results demonstrate that HDBSCAN is similarly as effective as mean-shift clustering. The effectiveness of HDBSCAN is because similar to mean-shift clustering,  it makes minimal assumptions about the the underlying dataset. HDBSCAN does not partition the data, and it effectively leaves sparse or noisy data points as independent clusters, hence noisy data points (e.g., poisoned data points) are not lumped with the nearest cluster. Overall, this result suggests that HDBSCAN is an effective alternative to mean-shift clustering for \AG.  
}

\begin{result}
\revision{
HDBSCAN is almost as effective as mean-shift clustering for \AG.  Similar to default \AG (\AG -MS), \AG with HDBSCAN (\AG -HDBSCAN) detects the backdoor-infected model as well as the poisoned class. 
}
\end{result}

\begin{table*}[t]
  \begin{center}
      \caption{\revision{\AG -t-SNE vs. \AG -AP vs. \AG -HDBSCAN: Evaluation of Clustering Design Choice using localised and distributed CIFAR-10 visible backdoor-infected robust models. True Positives (i.e., correct detection of the poisoned class) are in \textbf{bold} and False Positives (i.e., incorrect detection of a benign class as poisoned) 
are in \textit{\textbf{bold italics}} (e.g., \AG -AP). 
      }}
  {\scriptsize
  \begin{tabular}{|l|l|l|c|c|c|c|c|c|c|c|c|c|c|c|c|c|} 
    \hline 
\multirow{3}{*}{\textbf{Dataset}} &  \textbf{Type of} & \multicolumn{1}{l|}{\multirow{3}{*}{\shortstack{\textbf{Detection}\\\textbf{Setting}}}} & 
\multicolumn{10}{c|}{\textbf{Number of Resulting Clusters}}  & \multicolumn{2}{c|}{\multirow{2}{*}{\shortstack{\textbf{Backdoor} \\ \textbf{detected}}}} \\

& \textbf{Visible} &  & \multicolumn{9}{c|}{\textbf{Benign classes}} & \textbf{Poisoned class} & \multicolumn{2}{c|}{}  \\    
& \textbf{Backdoor} & & \textbf{0} & \textbf{1} &\textbf{ 2} & \textbf{3} & \textbf{4} & \textbf{5} & \textbf{6} & \textbf{8} & \textbf{9} & \textbf{7} & \textbf{Class} & \textbf{Model}   \\

    \hline
 \multirow{6}{*}{\textbf{CIFAR-10}} & \multirow{3}{*}{Localised}  & \AG -MS &  2 &  2 &  2 &  2 &  2 &  2 &  2 &  2  & 2 & \textbf{3} & \cmark & \cmark \\
 \cline{3-15}
 & & \AG -AP &  \textit{\textbf{19}} & \textit{\textbf{19}} & 	\textit{\textbf{34}}	 & \textit{\textbf{17}} & 	\textit{\textbf{196}} & \textit{\textbf{21}} &	\textit{\textbf{73}} & \textit{\textbf{23}} & \textit{\textbf{145}}  & 105 & \xmark & \xmark \\
 \cline{3-15}
 & & \AG -HDBSCAN  &  2 &  2 &  2 &  \textit{\textbf{3}} &  2 &  2 &  2 &  2  & 2 & \textbf{3} & \cmark & \cmark \\
     \cline{2-15}
    &  \multirow{3}{*}{Distributed}   & \AG -MS &  2 &  2 &  2 &  2 &  2 &  2 &  2 &  2  & 2 & \textbf{3} & \cmark   & \cmark \\
    \cline{3-15}
 & & \AG -AP & \textit{\textbf{20}} & 	\textit{\textbf{116}} & \textit{\textbf{104}}	& \textit{\textbf{104}} & \textit{\textbf{72}} &	\textit{\textbf{54}} &	\textit{\textbf{53}} & \textit{\textbf{20}} & \textit{\textbf{143}}  & 21 & \xmark & \xmark \\
 \cline{3-15}
      & &   \AG -HDBSCAN &  2 &  2 &  2 &  2 &  2 &  2 &  2 &  2  & 2 & \textbf{3} & \cmark & \cmark \\
     \hline
    \end{tabular}}
  \label{tab:design-choices-clustering}   
\end{center}
\vspace{-\baselineskip}
\end{table*}

\subsection{Discussions and Future Outlook}

\revise{
In this section, we discuss concerns about the application of \AG, in particular, how an adaptive attack can evade the detection of \AG, and how implicit assumptions of \AG (w.r.t. data distribution) can be applied to fool it or influence its performance. 
}

\smallskip\noindent
\revise{
\textbf{Counter-measures against \AG:} An attacker that is aware of \AG 's detection methodology can ensure that clean or backdoored models are trained in a manner that tricks \AG and reduces its effectiveness. For instance, instead of the typical backdoor data poisoning attack vector, a powerful attacker can train a backdoor-infected model such that the backdoor image mimics neuron output values (seen in clean models). This powerful attack may evade the detection of \AG, such that 
instead of simply causing a mis-classification of the backdoored image, it fools \AG to believe the backdoor neuron representation is similar to the neuron representation of clean images. Likewise, an adaptive attacker can deceptively train clean models to 
 reduce the accuracy of \AG. As an example, an attacker may fool \AG by ensuring that clean models (similar to backdoored models) also have more than one data distribution. Although, this attack does not affect the detection of backdoor models by \AG, it may cause false positives, where \AG also detects such deceptive clean models as backdoored models. In the future, we plan to investigate these more powerful attack vectors and explore potential defenses to protect against them beyond \AG and our current threat model. 
}

\smallskip\noindent
\revise{
\textbf{Data Distribution assumption:}
In this work, we have assumed that \textit{clean models have only one data distribution for each class label}, hence, \AG detects backdoors by examining if the backdoored model has more than one data distribution for a class label. Concretely, there is an implicit assumption in our method that the data corresponding to each label in the dataset contains data of only one distribution. Although, this assumption is valid within our threat model, it may not hold in other scenarios. 
As an example, consider a binary computer vision classifier 
which detects {\em dog} images, 
such that it has two classes or output labels, i.e, {\em dog}, and \textit{not dog}. Consider that this classifier is trained 
on a dataset containing 
multiple animal images (e.g., cat, horse, rat etc.), which may correspond to multiple distributions for the {\em not 
dog} class. As a result, our data distribution assumption may not hold in this scenario. 
Besides, this assumption may lead to wrong detection of clean models as backdoored model, if the clean models also have multiple distributions, e.g., because of the limitations of the training (e.g., local optima or incomplete training), or the complexity of the task/dataset (e.g., for multi-label or multi-output classification). For instance, consider a classifier trained on a (fashion) dataset, where a data point (e.g., an image of a person wearing a piece of clothing) can be classified into multiple labels (e.g., the gender of the person, the type/size of clothing and the color of the cloth). In this scenario, the data distribution assumption may not hold for each label. 
As an example, consider the distribution of the ``shirts'' clothing label, which may contain multiple data distributions representing different gender, sizes and colors of shirts.  
However, in our threat model the user has access to the training data, 
and can examine the data distribution before-hand (e.g., through methods such as t-SNE). 
Thus, the user can successfully analyze whether the number of distributions learned in the  trained model correspond to the actual data, and if not there may be a  backdoor distribution. 
In a different threat model where the user does not have access to the training data, they may have no means to verify such an assumption. However, 
note that \AG can still detect a backdoored model in the absence of this assumption, i.e., 
even when this data distribution assumption does not hold, we expect that the backdoored model still has multiple data distributions. 
}

\revise{
In a scenario where the user has no access to the training data for inspection, then the user may not be able to determine whether our data distribution assumptions holds or not. In this threat model, an attacker can further leverage this lack of data access to fool \AG, e.g., by ensuring that our data distribution assumption does not hold for clean models. 
As an example, an attacker can ensure that  
clean models are not properly trained, thus they converge to a local optima, hence, an \textit{adaptive attacker} can ensure that trained clean models have more than one data distribution. Alternatively, the attacker can ensure that backdoored models also converge to a local optima with one data distribution. These attack vectors and threat models can fool \AG and reduce its effectiveness. However, we expect that such an adaptive attack will reduce the performance of the model (e.g., in terms of accuracy), since it forces the trained model to be sub-optimally trained. Furthermore, this assumption does not influence the ability of \AG to detect backdoor models, although it may cause a false positive detection (i.e., wrongly detect a clean model as a backdoored model because the clean model has more than one data distribution). In the future, we plan to investigate these line of attacks concerning alternative threat models and multiple data distributions, 
in order to develop potential defenses to mitigate these attack vectors. 
}

\smallskip\noindent
\textbf{\revision{Robustness Check:}} \revision{
Since \AG is designed to detect backdoors in robust models, it assumes the analyzed model is robust, i.e., trained under robust optimization conditions. To ascertain an examined model is robust, there are automatic tests for adversarial robustness. For instance, Madry et al.\cite{pgd-madry} demonstrates a white-box approach that inspects the last layer of a model to check if a model is robust. Likewise, a brute-force black-box test is to check the performance of the model on adversarial examples within the expected perturbation bound. Both of these approaches are reliable and easy to automate. Thus, using the aforementioned methods, it is possible to check for model robustness to determine the applicability of \AG for the model-at-hand. 
Besides, we have demonstrated that \AG is not applicable to standard models: It does not produce false positives for standard models and it does not detect backdoors in backdoor-infected standard models (see RQ2). Thus, in the absence of a robustness check, \AG does not detect a backdoor in a standard model (as intended) and it does not classify a clean standard model as backdoor-infected. 
}

\smallskip\noindent
\textbf{\revision{Alternative Threat Models:}} \revision{
In the following we discuss changes in the threat model that may influence the performance of \AG in detecting backdoor in robust models. 
Firstly, this work assumes the 
\textit{attacker 
trains on the entire training dataset} (\textit{see} \autoref{sec:overview}). However, an \textit{adaptive attacker can train on a subset of the training data} such that the model still attains an acceptable performance. In this threat model, such an attack may make it more difficult for \AG to detect the backdoor-infected model or class. 
We expect that achieving an acceptable performance requires a substantial subset of the dataset that preserves the mixed data distribution hypothesis leveraged by \AG. However, in the future, we plan to investigate the potential of such adaptive attacks, 
how much they may influence (e..g, decrease) the performance of \AG, and how to (extend \AG to) appropriately mitigate them. 
}

\revision{
We also assume an attacker directly poisons the training examples during outsourced model training, but \textit{a sophisticated attacker may have access to the data
preparation pipeline such that he/she can directly poison the objects or scenes captured in the training images}. This attack is much stronger than our default setting because of the potential \textit{naturalness} of the resulting images. Given that the attacker only poisons for a targeted class and the mixed data distribution is preserved, we expect \AG to detect the backdoor in the resulting training images from such an attack. However, such a powerful attack may fool \AG because of the naturalness of the attack, especially if the attacker has sufficient resources to poison a substantial number of real-world objects/scenes (beyond the targeted class). Hence, such a ``natural'' attack may require new detection methods. We 
encourage researchers to investigate this ``natural'' backdoor attack vector and how to mitigate such attacks.
}
%

\revision{
Finally, \AG is designed to be a white-box, post-training backdoor detection method. Hence, it may not be directly amenable to other threat settings, e.g., black-box, pre-training and in-training scenarios. 
For instance, consider scenarios that require detecting backdoors in poisoned datasets before (e.g., fingerprinting the information content in poisoned examples) or during the training process, \AG is not applicable in these scenarios since it requires white-box access to an already trained model and a clean training dataset. 
In addition, \AG may not be directly applicable to scenarios involving an ensemble of models or multiple datasets (e.g. federated learning). These scenarios are beyond the scope of this work and may require fundamentally different backdoor detection methods than \AG, since these threat models do not fulfill the requirements/assumptions of \AG.
In the future, we plan to investigate these alternative threat models and how to mitigate backdoor in these settings. As an example, we plan to investigate the information content of backdoor examples to inform the automatic identification of backdoor in black-box, pre-training or in-training scenarios with no access to the model. In addition, we plan to investigate the impact of ensemble models and multiple data sources (e.g., federated learning settings) on the effectiveness of backdoor injection and detection. We also encourage researchers to develop practical approaches to defend and mitigate against backdoor in such alternative threat models. 
}

\revision{
In summary, similar to Athena~\cite{meng2020athena} -- a generic defense against adversarial attacks, we encourage researchers to investigate general defense mechanisms against backdoor that are applicable to several threat models. 
In the same vein, we plan to investigate generic defenses that are applicable to different threat models, e.g., with varying access levels including zero-knowledge, black-box, gray-box, and white-box.
}

\section{Threats to Validity}
\label{sec:threatsToValidity}
%
%
%


Our evaluation is limited by the following threats to validity:

\smallskip\noindent
\textbf{External validity:} \revision{
This refers to the generalisability of our approach and results. In particular, our findings may not generalize to other settings, different from the employed setting, specifically different (image classification) tasks, neural architectures, datasets and robust optimization methods. In the following, we discuss these threats/limitations in detail. 
}

\smallskip\noindent
\textit{\revision{Tasks, Datasets and Class labels}:} \revision{There is a threat that our findings and approach (\AG) may not generalize to other classification tasks, datasets or more complex, larger labels. We have mitigated this threat by evaluating the performance of our approach using three major image classification tasks with varying levels of complexity (CIFAR-10, MNIST and FashionMNIST). These tasks have thousands of training and test images, providing confidence that our approach will work on similarly complex tasks and models. Despite this mitigation, our findings are limited to these settings. Indeed, our findings may not be applicable to other object recognition datasets, other image classification tasks (e.g. image segmentation) and other classification tasks (e.g., image captioning). Besides, our experiments involve few ($<=$10) class labels, thus it may not generalize to models with a much larger number of labels or more complex labels (e.g., multi-label classification). 
In the future, we plan to investigate the applicability of \AG to different or more complex tasks. We also plan to develop generic approaches that are applicable across several tasks. 
}

\smallskip\noindent
\textit{\revision{Neural Architecture and Robust Optimization:}} 
\revision{
Our experiments were conducted using a specific neural architecture and robust optimization method, in particular, the ResNet architecture~\cite{he2016deep}, and adversarial training (AT)~\cite{pgd-madry}. Hence, there is a threat that our findings do not generalize to simpler or more complex neural architectures where the model has less or more capacity. Besides, our findings may not generalize to other robust optimization methods beyond adversarial training. In the future, we plan to examine backdoor injection and defense across different neural architectures and robust optimization methods (e.g., adversarial defense via diversity and ensemble models~\cite{abbasi2017robustness, li2021ensemble, pang2019improving, kariyappa2019improving}). For a general evaluation of backdoor in robust models, we encourage researchers to employ a standard and wide range of adversarial defenses under different threat models (e.g., AutoAttack\footnote{\url{https://github.com/fra31/auto-attack}}~\cite{croce2020reliable, croce2021mind} and RobustBench\footnote{\url{https://robustbench.github.io/}}~\cite{croce2020robustbench}). 
}

\smallskip\noindent
\textbf{Internal validity:} This concerns the correctness of our implementation of backdoor attacks and \AG ' defense. This includes 
whether we have performed adversarial training rightly, accurately defined (in)visible backdoor triggers, successfully injected backdoors, and correctly implemented \AG. We mitigate this threat by thoroughly testing our implementations on sample images to ensure our implementation works as expected. In addition, we provide our implementation, datasets and results for replication and scrutiny. 

\smallskip\noindent
\textbf{Construct validity:} It is possible that advanced backdoor triggers can be crafted to align to the input distribution of the training dataset. We mitigate this threat by ensuring that our backdoor triggers are similar to the ones described in the literature, as reported in previous related research. 
We emphasize that for robust models, the success and mitigation of backdoor attack variants such as blind backdoors~\cite{bagdasaryan2020blind}, trojaning~\cite{trojannn,guo2020trojannet,zou2018potrojan} and adaptive attacks~\cite{gao2019strip} are open research problems. These attacks have not been investigated for robust models. We consider the investigation of these advanced attacks against robust models as future work.


\revision{
The backdoor detection scenario employed in this work is a threat to the construct validity of \AG. In our attacker model, the attacker injects backdoors in robust models via a \textit{third-party platform} and the user has access to the the clean training data, clean testing data and the trained robust model. We assume an attacker introduces poisoned examples into the training data when the model training is outsourced to a third-party. However, Li et al.~\cite{li2020backdoor} has shown that there are alternative processes for injecting backdoors. Specifically, backdoors can also be injected via two alternative scenarios, i.e., (i) \textit{third-party datasets} -- where an attacker provides the poisoned dataset to users directly, or \ (ii) \textit{third-party models} -- where the attacker provides trained infected DNNs to the user~\cite{li2020backdoor}. We expect that \AG is  applicable in the \textit{third-party dataset} scenario since the user has the capabilities required by \AG, i.e., access to the clean training data, clean testing data and white-box access. However, our approach (\AG) may not be directly applicable in the \textit{third-party model} scenario since the user/defender may lack access to the training set and white-box access to the trained models. 
In the future, we shall investigate the injection and detection of backdoors in alternative attacker/defender scenarios.
}


\revision{
Lastly, our design choices pose a threat to construct validity. 
Specifically, our use of \textit{mean-shift clustering} and \textit{t-SNE for dimensionality reduction} 
may influence the effectiveness of \AG. To mitigate this threat, we have conducted an ablation study investigating the effectiveness of our choices in comparison to alternative design choices (\textit{see} RQ7). 
}

\section{Related Work}
\label{sec:background}

%
%
%



\textbf{
Robust Optimization:}
Adversarial attacks for Neural Networks (NNs) were first introduced in \cite{adversarial-init}. 
Researchers have introduced better adversarial attacks and built systems that are resilient to these attacks \cite{def-distillation, deep-learning-limit, blackbox-adversarial,  ensemble-defence}. A significant leap has been made by introducing robust optimisation to mitigate adversarial attacks \cite{pgd-madry, provable-defence, certified-defence,  distributional-robustness}. These defences aim to guarantee the performance of machine learning models against adversarial examples. In this paper, we study the susceptibility of the models trained using robust optimisation to backdoor attacks. Then, we leverage the inherent properties of robust models to detect backdoor attacks.

\revision{
In this work, we have studied backdoor detection/injection in robust models using adversarial training (AT) as the \textit{only} robust optimization method. Even though adversarial training is an effective and well-known defense against adversarial examples (AE), there are other robust optimization techniques beyond adversarial training which may not be susceptible to backdoor detection or amenable to backdoor detection by \AG. A number of researchers have demonstrated that adversarial robustness can be achieved via an ensemble of diverse models~\cite{li2021ensemble, pang2019improving, kariyappa2019improving}, or by detecting unrecognised, potentially adversarial examples~\cite{abbasi2017robustness}. Indeed, we do not know the susceptibility of these robust optimization methods (except AT) to backdoor attacks, and \AG may not be sufficient to defend against backdoor injection in these  settings. In the future, we plan to investigate the susceptibility of different robust optimization methods to backdoor attacks and how to effectively defend against them. 
}


\revision{In addition, our findings and observations about backdoor attacks and \AG 
is \textit{strictly empirical}. We provide no theoretical bound to the susceptibility of robust optimization to backdoor attacks or guarantees of \AG 's defense. 
Indeed, it is vital to understand how our empirical observations relate to the robust optimization theory, e.g., in terms of the susceptibility of standard and robust models to backdoor attacks (\textit{see} RQ1). In addition, it is interesting to know the lower and upper bound accuracy of  \AG on certain poisoning attacks (e.g., (in)visible distributed or localized backdoor triggers), and how this relates to formulating backdoor defense as studying spurious/non-robust features in robust models~\cite{ilyas2019adversarial}. In particular, we encourage the theoretical investigation of how backdoor injection and defenses relate to robust optimization theory to provide mathematical insights into our empirical observations. 
}

\smallskip
\noindent
\textbf{Backdoor Attacks:} Backdoor attacks were introduced in BadNets~\cite{BadNets}, where an attacker  poisons the training data by augmenting it. A pre-defined random shape (called trigger) is chosen for the attack. TrojanNN~\cite{trojannn} improves the attack  by engineering the trigger and reducing the number of examples needed  to insert the backdoor. Yao et al.~\cite{latent-backdoor} propose a transfer learning based backdoor. All of these attacks are visible to the human eye. Besides, other variants of backdoor attacks have also recently been developed such as blind 
backdoors~\cite{bagdasaryan2020blind},  trojaning~\cite{trojannn,guo2020trojannet,zou2018potrojan} and adaptive  attacks~\cite{gao2019strip}. In addition, Zhong et al. proposed a backdoor attacks where the trigger is hidden~\cite{zhong2020backdoor}. 

\revision{Li et al.~\cite{li2020backdoor} provides a systematic literature review of backdoor attack mechanisms. This work demonstrates varying attack/threat models for backdoor attacks, for instance in terms of the access level of the attacker (e..g, access to training set, training schedule, model and/or inference pipeline). The paper also provides a taxonomy of poisoning based attacks (e.g., trigger properties such as target level, visibility, selection, appearance, and type (digital versus physical).) In this work (\AG), we have focused on the injection and detection of both invisible and visible backdoor attacks in robust models.} The aforementioned attacks were demonstrated for standard models, not for robust training. To the best of our knowledge, we are the first to demonstrate the susceptibility of models trained under robust optimisation conditions~\cite{pgd-madry} to (both visible and invisible) backdoor attacks.

\smallskip
\noindent
\textbf{Backdoor Detection and Mitigation:}
 \revision{
Several approaches have been developed to detect and mitigate backdoor attacks on standard machine learning models.
Li et al.~\cite{li2020backdoor} provides a comprehensive analysis of 
different defense mechanisms under different threat models.\footnote{\revision{\url{https://github.com/THUYimingLi/backdoor-learning-resources}}} 
\Cref{tab:defense-comparison} compares an excerpt of the main characteristics of these approaches.
These approaches can be categorized into three main types, namely, backdoor detection via (1) outlier suppression, (2) input perturbation and (3) model anomalies~\cite{bagdasaryan2020blind}.
}

\emph{Outlier suppression} based defenses prevent backdoored inputs from being introduced into the model~\cite{du2019robust,hong2020effectiveness}. 
The main idea of these approaches is to employ differential privacy mechanism to ensure that backdoored inputs are under-represented in the training set. Unlike these approaches, our approach is not a training-time defense, rather the focus of our approach is to detect models that are already poisoned with backdoored inputs.

\emph{Input perturbation} methods detect backdoors by attempting to reverse engineer small input perturbations that trigger backdoor behavior in the model. Such approaches include Neural Cleanse (NC)~\cite{NeuralCleanse}, ABS~\cite{liu2019abs}, TABOR~\cite{guo2019tabor}, STRIP~\cite{gao2019strip}, NEO~\cite{neo}, DeepCleanse~\cite{doan2019deepcleanse},
AD~\cite{xiang2020detection} and MESA~\cite{qiao2019defending}. In this paper, we focus on comparison to Neural Cleanse (NC)~\cite{NeuralCleanse}, we used NC as the representative backdoor defense. We compare our approach to NC (see \RQ3), since NC is the state of the art and it has realistic defense assumptions (similar to \AG) (\textit{see \Cref{tab:defense-comparison}}). In particular, NC relies on finding a fixed perturbation that mis-classifies a large set of inputs, but since robust models are designed to be resilient to exactly such perturbations, we show that NC is inapplicable for robust models.

\emph{Model anomaly} defenses detect backdoors by identifying anomalies in the model behavior. Most of these techniques focus on identifying how the model behaves differently on benign and backdoored inputs, using model information such as logit layers, intermediate neuron values and spectral representations. These approaches include SentiNet~\cite{chou2018sentinet}, spectral signatures~\cite{spectral-signatures},
fine-pruning~\cite{finePruning}, NeuronInspect~\cite{huang2019neuroninspect}, activation clustering~\cite{activation-clustering}, SCAn~\cite{tang2019demon}, NNoculation~\cite{veldanda2020nnoculation} and MNTD~\cite{xu2019detecting}. 
However, unlike our approach, none of these techniques detect backdoors in robust models. Additionally, SCAn~\cite{tang2019demon}, SentiNet~\cite{chou2018sentinet}, activation clustering~\cite{activation-clustering} 
and spectral signatures~\cite{spectral-signatures} assume access to the poisoned dataset -- an impractical assumption 
for backdoor defense (\textit{see \Cref{tab:defense-comparison}}). Moreover, fine-pruning~\cite{finePruning} is shown to be ineffective in existing work~\cite{NeuralCleanse} 
and NNoculation~\cite{veldanda2020nnoculation} and MNTD~\cite{xu2019detecting} require training a shadow model for 
defense, leading to a computationally inefficient process. In contrast, \AG is computationally efficient, it does not 
require access to the poisoned dataset and it accurately detects backdoor-infected robust models. 
 
Unlike the aforementioned works, we rely on the \textit{clustering of feature representations in robust models} to  detect backdoor attacks. Like our approach, Chen et al.~\cite{activation-clustering} employs feature  clustering to detect backdoors in standard DNNs; 
it uses the feature representations of the 
training and poisoned data to detect the poisoned data. However, 
their approach relies on \textit{the strong  assumption that the user 
has access to the poisoned  dataset}. 
Our approach requires access to only the model and the clean training 
dataset.



\smallskip\noindent
\revision{
\textbf{Adversarial Training and Backdoor Robustness:} 
Several researchers have studied the relationship between adversarial inputs and poisoned models (including backdoor~\cite{pang2020tale, weng2020trade}). Remarkably, Pang et al.~\cite{pang2020tale} systematically studied the relationship between adversarial inputs and poisoned models in a unified manner by developing a new attack model that jointly optimizes both attacks. This work shows that there is a mutual reinforcement effect between the two attack vectors which can be easily exploited to optimize attacks with respect to multiple metrics. For instance, this work shows that leveraging one attack vector significantly amplifies the effectiveness of the other. Similar to our work (\AG), the paper encourages the need to study both attacks by designing countermeasures, albeit from multiple complementary perspectives (e.g., efficacy, fidelity and specificity) to account for the mutual reinforcement effects.
}

\revision{
Similarly, researchers have shown that there is a trade-off between adversarial robustness and backdoor attacks. 
Notably, Weng et al.~\cite{weng2020trade} demonstrated that adversarial robustness is at odds with backdoor robustness. 
The authors found that increasing adversarial robustness via adversarial training makes a model more vulnerable to backdoor attacks. Consequently, this trade-off can influence the strength of both attacks and defenses against backdoor attacks. %
Weng et al.~\cite{weng2020trade} 
shows that this trade-off can be leveraged to create more concealed backdoor attacks that evade existing backdoor defenses, and it can also be leveraged to further strengthen some defenses. Similarly, 
this work (\AG) shows that the inherent properties of adversarial training based robust optimization can aid the detection/defense against backdoor attacks in robust models. 
In the future, we plan to investigate the extent to which increasing or decreasing adversarial robustness may influence the success of backdoor attacks and the defense of \AG. In addition, we plan to investigate how to extend \AG to achieve joint defense against both adversarial and backdoor attacks. 
}

\section{Conclusion}
\label{sec:conclusion}

In this paper, we demonstrate a new attack vector for PGD-trained robust DNN models, namely backdoor attacks. We show that such robust models are susceptible to several variants of backdoor attacks, including visible and invisible backdoors. Then, we leverage the inherent properties of these robust ML models to detect this attack. Our proposed detection technique (i.e., AEGIS) is based on clustering the feature representation of PGD-trained robust models to find anomalous clusters. In our evaluation, AEGIS accurately detects backdoor-infected PGD-trained robust models and identifies the poisoned class, without any access to the poisoned data, for all visible backdoor triggers. We also found that invisible backdoor triggers are more stealthy and slightly more difficult to detect for AEGIS. Overall, AEGIS detects a backdoor-infected model with 91.6\% accuracy (i.e., 11 out of 12 backdoor-infected models), without any false positives. Furthermore, AEGIS detects the targeted class in the backdoor-infected model with a reasonably low (11.1\%) false positive rate. Our work reveals that inherent properties of PGD-based robust optimization method allows to expose backdoors. 
%
%
%
%
Our code and experimental data are available for replication: 
\begin{center}
        \textbf{\url{https://github.com/sakshiudeshi/Expose-Robust-Backdoors}}
\end{center}






\balance

\bibliographystyle{IEEEtran}
\bibliography{AEGIS}


\onecolumn
\appendix
%


\section{Additional Tables}
\label{sec:additional-tabs} 

 \begin{table}[h]
  \begin{center}
      \caption{
    Standard hyperparameters used for model training.}
  { 
  \begin{tabular}{|l|r|r|r|r|}
    \hline 
\multicolumn{1}{|c|}{\textbf{Dataset}} & \multicolumn{1}{c|}{\multirow{1}{*}{\textbf{Epochs}}} & \multicolumn{1}{c|}{\textbf{LR}} & \multicolumn{1}{c|}{\textbf{Batch Size}}  & \multicolumn{1}{c|}{\textbf{LR Schedule}} \\ 
    \hline
     CIFAR-10  & 110  &  0.1 & 128 & Drop by 10 at epochs  $\in [50, 100]$ \\  
    \hline	
    MNIST  & 100  & 0.1 & 128 & Drop by 10 at epochs  $\in [50, 100]$  \\ 
    \hline
    Fashion-MNIST & 100  & 0.1 & 128 & Drop by 10 at epochs  $\in [50, 100]$  \\  
     \hline     
    \end{tabular}}
  \label{tab:hyperparameters}   
\end{center}
\vspace{-\baselineskip}
\end{table}

\begin{table}[h]
  \begin{center}
    \caption{
  Backdoor Detection Parameters}
  { 
  \begin{tabular}{|rlr|r|r|} 
  \hline
  \multicolumn{1}{|c|}{\multirow{2}{*}{\shortstack{\textbf{Detection}\\\textbf{Parameters}}}} &  \multicolumn{3}{c|}{\textbf{All Models}} \\
  \multicolumn{1}{|c|}{} & \multicolumn{1}{c|}{\textbf{MNIST}} & \multicolumn{1}{c|}{\textbf{Fashion-MNIST}} &  \multicolumn{1}{c|}{\textbf{CIFAR-10}} \\
  \hline 
  \multicolumn{1}{|c|}{Epsilon ($\epsilon$)} & \multicolumn{1}{c|}{100} & \multicolumn{1}{c|}{100}  & \multicolumn{1}{c|}{500} \\
  \hline
  \multicolumn{1}{|c|}{t-SNE Perplexity} & \multicolumn{1}{c|}{30} &  \multicolumn{1}{c|}{30} & \multicolumn{1}{c|}{30} \\
  \hline
  \multicolumn{1}{|c|}{Mean shift Bandwidth} & \multicolumn{1}{c|}{35} & \multicolumn{1}{c|}{28} & \multicolumn{1}{c|}{21} \\      
  \hline
    \end{tabular}}
   \label{tab:detection-params}   
   \vspace*{-0.2in}
\end{center}
\end{table}


\begin{table*}[h]
  \begin{center}
    \caption{Detection Efficacy: Number of feature clusters for each class for clean model and visible trigger infected backdoor models}
  {\scriptsize
  \begin{tabular}{|lrlr|r|r|r|r|r|r|rr|r|} 
  \hline 
\multicolumn{1}{|c|}{\multirow{3}{*}{\shortstack[r]{\textbf{Class}\\\textbf{Type}}}} & \multicolumn{1}{c|}{\multirow{3}{*}{\shortstack[l]{\textbf{Class}\\\textbf{Labels}}}}  & \multicolumn{3}{c|}{\textbf{MNIST Models}} & \multicolumn{3}{c|}{\textbf{Fashion-MNIST Models}} &  \multicolumn{3}{c|}{\textbf{CIFAR-10 Models}}\\
\multicolumn{1}{|c|}{}  & \multicolumn{1}{c|}{} & \multicolumn{2}{c|}{\textbf{Backdoor-Infected}} & \multirow{2}{*}{\textbf{Clean}} &  \multicolumn{2}{c|}{\textbf{Backdoor-Infected}}  & \multirow{2}{*}{\textbf{Clean}} & \multicolumn{2}{c|}{\textbf{Backdoor-Infected}} & \multicolumn{1}{c|}{\multirow{2}{*}{\textbf{Clean}}} \\
\multicolumn{1}{|c|}{}  & \multicolumn{1}{c|}{} & \multicolumn{1}{c|}{\textbf{Local}} & \multicolumn{1}{c|}{\textbf{Distributed}} & & \textbf{Local} & \textbf{Distributed} &  & \textbf{Local} & \textbf{Distributed} & \multicolumn{1}{c|}{}  \\
      \hline
      \multicolumn{1}{|c|}{Targeted} & \multicolumn{1}{c|}{$\{7\}$} & \multicolumn{1}{c|}{\textbf{3}} & \textbf{3} & 2 & \textbf{4} & \textbf{3} & 2 & \textbf{3} & \textbf{4} & \multicolumn{1}{c|}{2}  \\
        \hline 
      \multicolumn{1}{|c|}{Untargeted} & \multicolumn{1}{c|}{$\{0-6,8,9\}$} & \multicolumn{1}{c|}{2} & 2 & 2 & 2 & 2 & 2 & 2 & 2 & \multicolumn{1}{c|}{2} \\
     \hline     
    \end{tabular}}

   \label{tab:visible-attack-detection-clusters}   
   \vspace*{-0.2in}
\end{center}
\end{table*}

\begin{table*}[h]
  \begin{center}
    \caption{Detection Efficacy: Number of feature clusters for each class for invisible backdoors}
  {\scriptsize
  \begin{tabular}{|lrlr|r|r|r|r|rr|} 
  \hline 
\multicolumn{1}{|c|}{\multirow{3}{*}{\shortstack[r]{\textbf{Class}\\\textbf{Type}}}} & \multicolumn{1}{c|}{\multirow{3}{*}{\shortstack[l]{\textbf{Class}\\\textbf{Labels}}}}  & \multicolumn{2}{c|}{\textbf{MNIST Models}} & \multicolumn{2}{c|}{\textbf{Fashion-MNIST Models}} &  \multicolumn{2}{c|}{\textbf{CIFAR-10 Models}}\\
\multicolumn{1}{|c|}{}  & \multicolumn{1}{c|}{} & \multicolumn{2}{c|}{\textbf{Backdoor-Infected}} &  \multicolumn{2}{c|}{\textbf{Backdoor-Infected}}  &  \multicolumn{2}{c|}{\textbf{Backdoor-Infected}}  \\

\multicolumn{1}{|c|}{}  & \multicolumn{1}{c|}{} & \multicolumn{1}{c|}{\textbf{Static}} & \multicolumn{1}{c|}{\textbf{Adversarial}}  & \textbf{Static} & \textbf{Adversarial}   & \textbf{Static} & \textbf{Adversarial} \\
      \hline
      \multicolumn{1}{|c|}{Targeted} & \multicolumn{1}{c|}{$\{7\}$} & \multicolumn{1}{c|}{\textbf{2}} & {\textbf{2}}  & \textbf{3} & \textbf{3}  & \textbf{3} & \textbf{4}  \\
        \hline 
      \multicolumn{1}{|c|}{Untargeted} & \multicolumn{1}{c|}{$\{0\}$} & \multicolumn{1}{c|}{1} & 2 & 3 & 2 & 2 & 2  \\ \cline {2-8} 
     \multicolumn{1}{|c|}{} & \multicolumn{1}{c|}{$\{1\}$} & \multicolumn{1}{c|}{2} & 2 & 3 & 2 & 2 & 3  \\ \cline {2-8}    
      \multicolumn{1}{|c|}{} & \multicolumn{1}{c|}{$\{2\}$} & \multicolumn{1}{c|}{2} & 2 & 2 & 2 & 2 & 2  \\ \cline {2-8}    
      \multicolumn{1}{|c|}{} & \multicolumn{1}{c|}{$\{3\}$} & \multicolumn{1}{c|}{2} & \textbf{3} & 2 & 2 & 2 & 2  \\ \cline {2-8}    
      \multicolumn{1}{|c|}{} & \multicolumn{1}{c|}{$\{4\}$} & \multicolumn{1}{c|}{2} & 2 & 2 & 3 & 2 & 2  \\ \cline {2-8}    
      \multicolumn{1}{|c|}{} & \multicolumn{1}{c|}{$\{5\}$} & \multicolumn{1}{c|}{2} & 2 & 2 & 2 & 2 & 2  \\ \cline {2-8}    
      \multicolumn{1}{|c|}{} & \multicolumn{1}{c|}{$\{6\}$} & \multicolumn{1}{c|}{2} & 2 & 2 & 2 & 2 & 2  \\ \cline {2-8}  
     \multicolumn{1}{|c|}{} & \multicolumn{1}{c|}{$\{8\}$} & \multicolumn{1}{c|}{2} & 2 & 3 & 2 & 2 & 2  \\ \cline {2-8}   
      \multicolumn{1}{|c|}{} & \multicolumn{1}{c|}{$\{9\}$} & \multicolumn{1}{c|}{2} & 2 & 2 & 2 & 2 & 2  \\ \cline {1-8}          
    \end{tabular}}

   \label{tab:invisible-attack-detection-clusters}   
   \vspace*{-0.2in}
\end{center}
\end{table*}

\begin{figure}[h]
\begin{center}
\begin{tabular}{cc}

\includegraphics[scale=0.35]{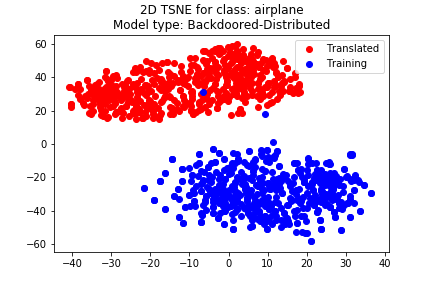}& 
\includegraphics [scale=0.35]{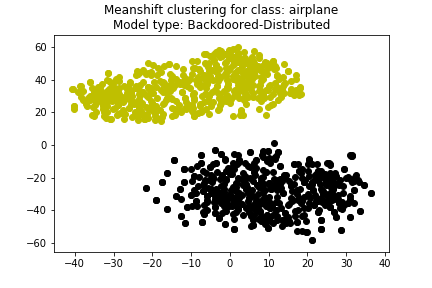} \\
{\textbf{Representative benign class}} & {\textbf{Predicted clusters for benign class}} \\
\includegraphics[scale=0.35]{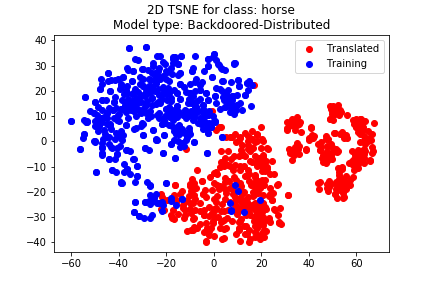}& 
\includegraphics [scale=0.35]{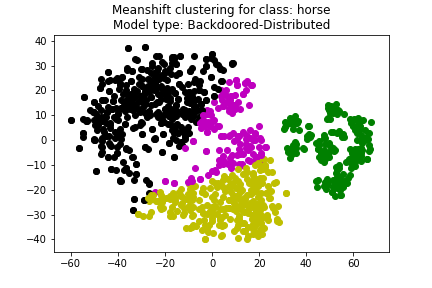} \\
{\textbf{Target class}} & {\textbf{Predicted clusters for backdoored class)}} \\

\end{tabular}
\end{center}
\vspace*{-0.15in}
\caption{
Feature representation clusters for backdoored CIFAR models (Distributed) with target class {\em Horse} (7). This figure shows class {\em 0} and {\em 7}. The left column shows the feature representations of the translated and the training images, whereas the right column shows the result of the Mean shift clustering on the corresponding points where different colours represent different classes.
}
\vspace*{-0.1in}
\label{fig:CIFAR-distributed-2}
\end{figure}

\begin{figure*}[h]
\begin{center}
\begin{tabular}{cc}

\includegraphics[scale=0.35]{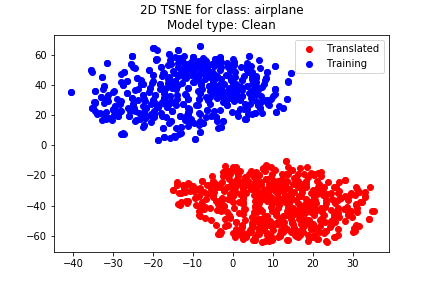}& 
\includegraphics [scale=0.35]{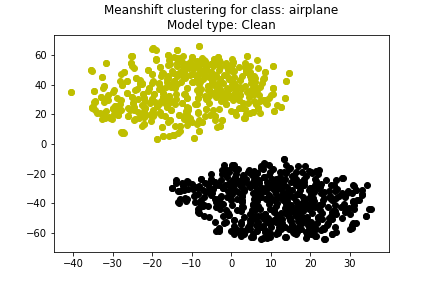} \\
\includegraphics[scale=0.35]{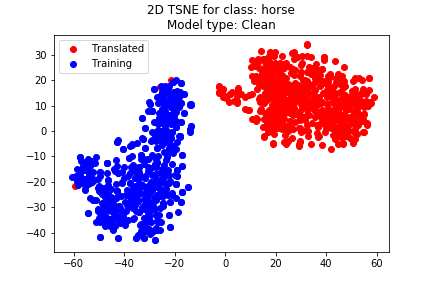}& 
\includegraphics [scale=0.35]{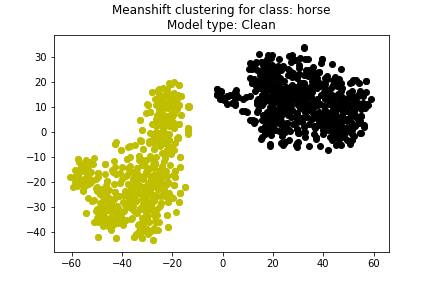} 

\end{tabular}
\end{center}
\vspace*{-0.15in}
\caption{
Feature representation clusters for clean CIFAR10 models. This figure shows class {\em 0} and {\em 7}. The left column shows the feature representations of the translated and the training images, whereas the right column shows the result of the Mean shift clustering on the corresponding points where different colours represent different classes.
}
\vspace*{-0.1in}
\label{fig:CIFAR-clean-1}
\end{figure*}

\begin{figure*}[h]
\begin{center}
\begin{tabular}{cc}

\includegraphics[scale=0.35]{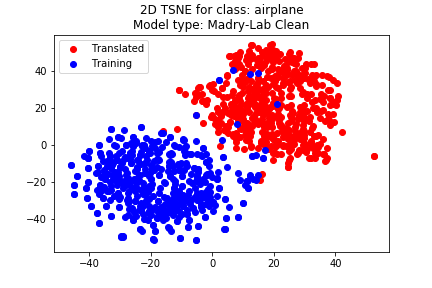}& 
\includegraphics [scale=0.35]{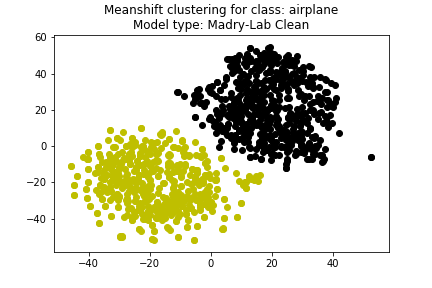} \\

\end{tabular}
\end{center}
\vspace*{-0.15in}
\caption{
Feature representation clusters for clean CIFAR10 models from 
Madry-Lab. This figure shows class {\em 0}. The left column shows the 
feature representations of the translated and the training images, whereas 
the right column shows the result of the Mean shift clustering on the 
corresponding points where different colours represent different classes.
It is important to note that the translated images and training set images 
form separate clusters.
}
\vspace*{-0.1in}
\label{fig:CIFAR-Madry-Lab-clusters}
\end{figure*}

\begin{figure*}[h]
\begin{center}
\begin{tabular}{cc}

\includegraphics[scale=0.35]{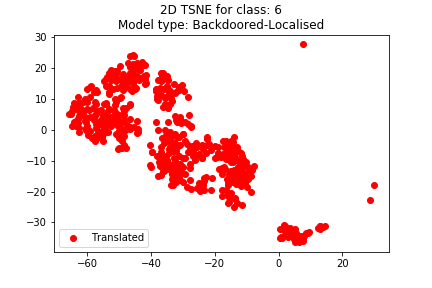}& 
\includegraphics [scale=0.35]{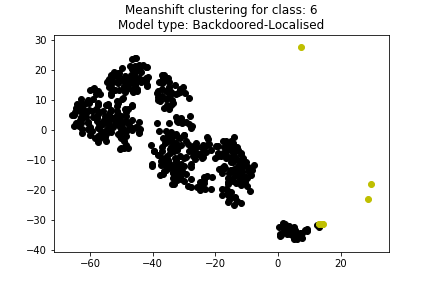} \\

\end{tabular}
\end{center}
\vspace*{-0.15in}
\caption{
Representative false positives. These kinds of false positives occur 
when \AG only considers the translated images in the detection for 
backdoors. This figure shows class {\em 6} of a robust MNIST model poisoned 
with a localised backdoor. The left column shows the 
feature representations of the translated and the training images, whereas 
the right column shows the result of the Mean shift clustering on the 
corresponding points where different colours represent different classes.
}
\vspace*{-0.1in}
\label{fig:false-positive-only-translated}
\end{figure*}

\begin{figure}[t]
\begin{center}
\begin{tabular}{cc}
\includegraphics[scale=0.5]{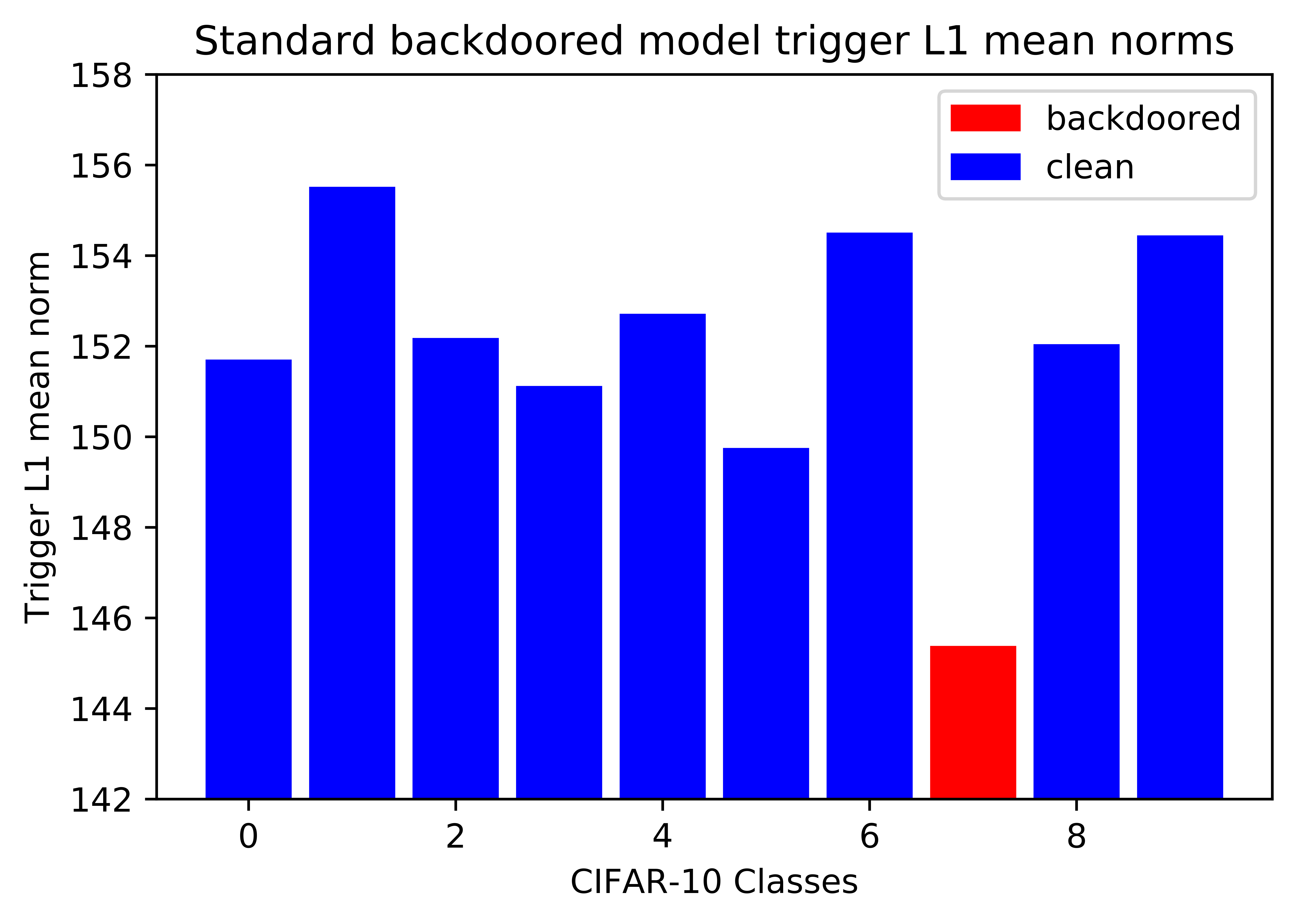} & 
\includegraphics[scale=0.5]{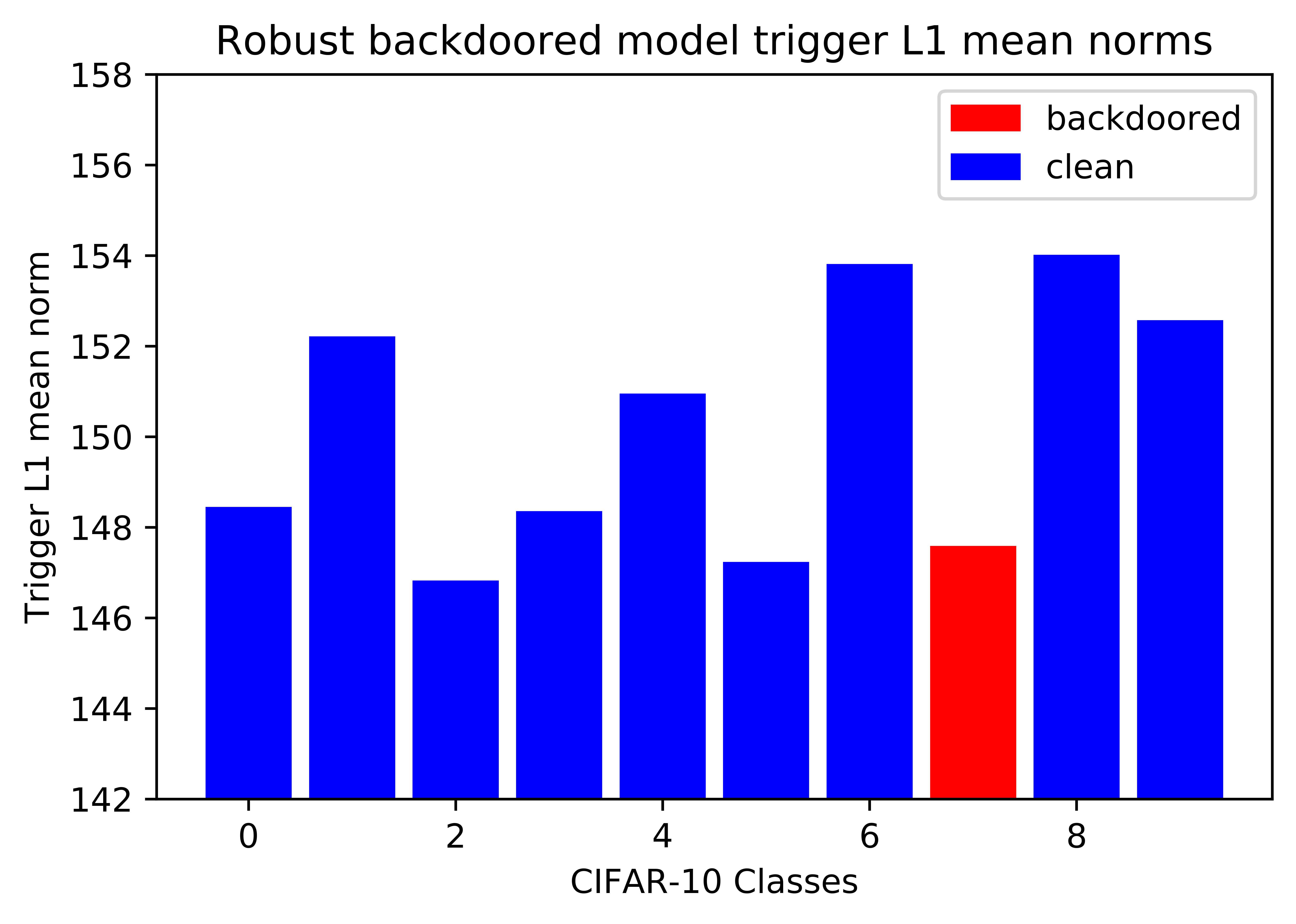} \\
{\bf(a) Standard model} & {\bf(b) Robust model}  \\
\end{tabular}
\end{center}
\vspace*{-0.15in}
\caption{
L1 norms (mean) of the reverse engineered triggers for backdoor-infected 
standard and robust models. \revise{The L1 norms for the reverse engineered 
triggers are in line with the sizes of the reverse engineered triggers seen 
in ~\cite{NeuralCleanse}.}}
\vspace*{-0.1in}
\label{fig:mean-norms}
\end{figure}




\end{document}